\documentclass{article} %
\usepackage{conference,times}

\usepackage{amsmath,amsfonts,bm}

\def\eqref#1{equation~\ref{#1}}

\def\1{\bm{1}}

\DeclareMathAlphabet{\mathsfit}{\encodingdefault}{\sfdefault}{m}{sl}
\SetMathAlphabet{\mathsfit}{bold}{\encodingdefault}{\sfdefault}{bx}{n}

\usepackage{hyperref}
\usepackage{url}

\usepackage{verbatim}

\usepackage{comment}

\usepackage[utf8]{inputenc} %
\usepackage[T1]{fontenc}    %

\usepackage{url}            %
\usepackage{booktabs}       %
\usepackage{amsfonts}       %
\usepackage{nicefrac}       %
\usepackage{microtype}      %
\usepackage{xcolor}         %

\usepackage{xcolor}

\usepackage{booktabs}

\usepackage{comment}
\usepackage{booktabs}
\usepackage{multirow}
\usepackage{adjustbox}
\usepackage[table]{xcolor}
\usepackage{colortbl}
\usepackage{url}
\usepackage{xspace}
\usepackage{xcolor}
\usepackage{colortbl}

\usepackage[ruled,vlined]{algorithm2e}

\usepackage{multirow}
\usepackage{balance}
\usepackage{subfigure}
\usepackage{graphicx}
\usepackage{mdframed}
\usepackage{mathtools}
\usepackage{enumitem}
\usepackage[noabbrev,capitalise]{cleveref}

\usepackage{booktabs}
\usepackage{caption}   %
\usepackage{float}    %

\crefname{appendix}{Appendix}{appendices}
\Crefname{appendix}{Appendix}{Appendices}
\crefname{equation}{Eq.}{Eqs.}

\Crefname{equation}{Eq.}{Eqs.}

\usepackage{amsthm} 
\usepackage{amsthm}

\usepackage{amsmath}
\usepackage{amsthm}
\usepackage{bm}
\usepackage{ragged2e}
\usepackage{colortbl}
\usepackage{booktabs}
\usepackage{colortbl}
\usepackage{xcolor}
\usepackage{amsmath, amssymb}
\usepackage{tcolorbox}
\usepackage{pifont}  %

\usepackage{xcolor} 
\usepackage{booktabs}
\usepackage{xcolor, booktabs, tcolorbox, amsmath, enumitem}
\tcbuselibrary{skins, breakable}

\usepackage{xcolor, booktabs, tcolorbox, amsmath, enumitem}
\tcbuselibrary{skins, breakable}

\usepackage{xcolor}
\usepackage{setspace}
\usepackage{listings}

\makeatletter
\let\MASOPDorigstartsection\@startsection
\def\@startsection#1#2#3#4#5#6{%
  \setlength{\parskip}{.5pc}%
  \MASOPDorigstartsection{#1}{#2}{#3}{#4}{#5}{#6}}
\makeatother

\newcommand{\ourmodel}{MAS-OPD\xspace}

\title{MAS-OPD: On-Policy Distillation for \\ Multi-Agent Systems}

\author{%
\begin{minipage}{\dimexpr\textwidth-2\tabcolsep\relax}
\centering
\textbf{Qiyong Zhong}$^{1,2}$\thanks{Equal contribution.} \quad
\textbf{Mao Zheng}$^{2}$\footnotemark[1] \quad
\textbf{Mingyang Song}$^{2}$\footnotemark[1] \quad
\textbf{Houcheng Jiang}$^{1}$ \\[0.35em]
\textbf{Jiajie Su}$^{3}$ \quad
\textbf{Huwei Ji}$^{3}$ \quad
\textbf{Li Zhang}$^{3}$ \quad
\textbf{Junfeng Fang}$^{4}$\thanks{Corresponding author.} \\
\vspace{0.8em}
{\normalfont
$^{1}$University of Science and Technology of China \quad
$^{2}$Foundation Model Department, Tencent \\
$^{3}$Zhejiang University \quad
$^{4}$National University of Singapore \\
}
\vspace{0.8em}
{\normalfont\small
\texttt{\{youngzhong365,zhanglizl80\}@gmail.com} ;
\texttt{\{sujiajie,jihuwei\}@zju.edu.cn} \\
\texttt{\{moonzheng,nickmysong\}@tencent.com} ;
\texttt{jianghc@mail.ustc.edu.cn} \\
\texttt{fangjf@nus.edu.sg}
}
\end{minipage}
}

\conffinalcopy

\begin{document}

\maketitle

\begin{abstract}

Multi-agent systems (MAS) split a task across specialized roles and are promising on complex tasks, yet a prevailing approach relies on inference-time orchestration alone.
General-purpose APIs are costly and hard to customize, while small models with role prompts rarely develop stable role competence or reliable collaboration, so post-training a MAS jointly is central.
Most attempts use reinforcement learning, whose team-level reward leaves undetermined which step of which agent brought about the outcome, while local rewards need redesigning per task.
On-policy distillation (OPD) gives token-level teacher supervision on trajectories the student samples, a denser signal needing no per-role reward, yet is underexplored for the interdependent agents of a MAS.
Two difficulties arise: building complementary specialization from a judgement of which role a behavior belongs to while preserving the knowledge all roles need, and turning cross-agent collaborative information into supervision OPD can exploit.
We present \mbox{\ourmodel}, where Role-Advantage Specialization defines the role advantage as the difference between the teacher signals under target and non-target role conditions, and Privileged Attribution for Coordination attributes an interaction conflict to its source and supplies it to the teacher alone as privileged information.
Extensive experiments on code and mathematics benchmarks show that \mbox{\ourmodel} attains the highest mean score at both student scales and leads the agents to develop clearer role specialization and more effective collaborative behavior.

\end{abstract}

\section{Introduction}\label{sec:introduction}

Multi-agent systems (MAS) built on large language models split a task across specialized roles collaborating over multiple turns~\citep{wu2023autogen,li2023camel}, and show considerable potential on complex tasks such as code generation, mathematical reasoning and long-horizon planning~\citep{qian2023chatdev,du2023llmdebate,Park2023GenerativeAgents}.
A prevailing approach relies on inference-time orchestration alone~\citep{guo2024llmma,pan2025whydomasfail}.
A system built on general-purpose APIs is costly and hard to customize for a domain or protocol~\citep{chen2024ioa,ye2025xmas}, whereas one built from small models and role prompts is cheaper but rarely develops stable role competence or reliable collaboration~\citep{wang2024moa,belcak2025slm_agentic}.
Jointly post-training a MAS for its target task is therefore central to efficient and customizable multi-agent systems.

Most work post-trains a MAS with reinforcement learning by having the agents act together and updating each policy from a team-level reward~\citep{magrpo,liao2025marft}.
A joint trajectory spanning many agents and turns contains many decisions while the environment returns only a single success signal at the end~\citep{shao2024deepseekmath,guo2025deepseekr1}.
Distributing this scalar over all agents and all their outputs leaves undetermined which step of which agent brought about the outcome.
Several works alleviate agent-level and turn-level credit assignment by designing local rewards for roles and turns~\citep{strongermas,feng2025group}, yet such feedback still operates at the granularity of a trajectory or a turn and depends on task-specific rules and reward shaping.
Whenever the task format, the role responsibilities or the protocol changes, these rewards must be redesigned, which limits how far reinforcement learning transfers across MAS settings.

On-policy distillation (OPD)~\citep{opd,gu2024minillm,lu2025onpolicydistillation} is an appealing alternative for MAS post-training.
Rather than optimizing against a scalar reward, OPD samples trajectories from the current student and queries a teacher for token-level supervision on them, so the student learns on the states it visits and receives a denser signal~\citep{yang2026learning,li2026rethinking}.
Extending this to a MAS would let the teacher score every agent's tokens at every turn and supervise each role directly, without a local reward per role and per turn.
Existing OPD research~\citep{opdsurvey,scope,srpo} nevertheless targets a single agent.
\textbf{How to use OPD to train the multiple interdependent agents of a MAS remains underexplored.}
This is not a matter of adding policies, and we highlight two challenges, illustrated in~\Cref{fig:motivation}:

\begin{figure*}[t]
\centering
\includegraphics[width=1.0\linewidth]{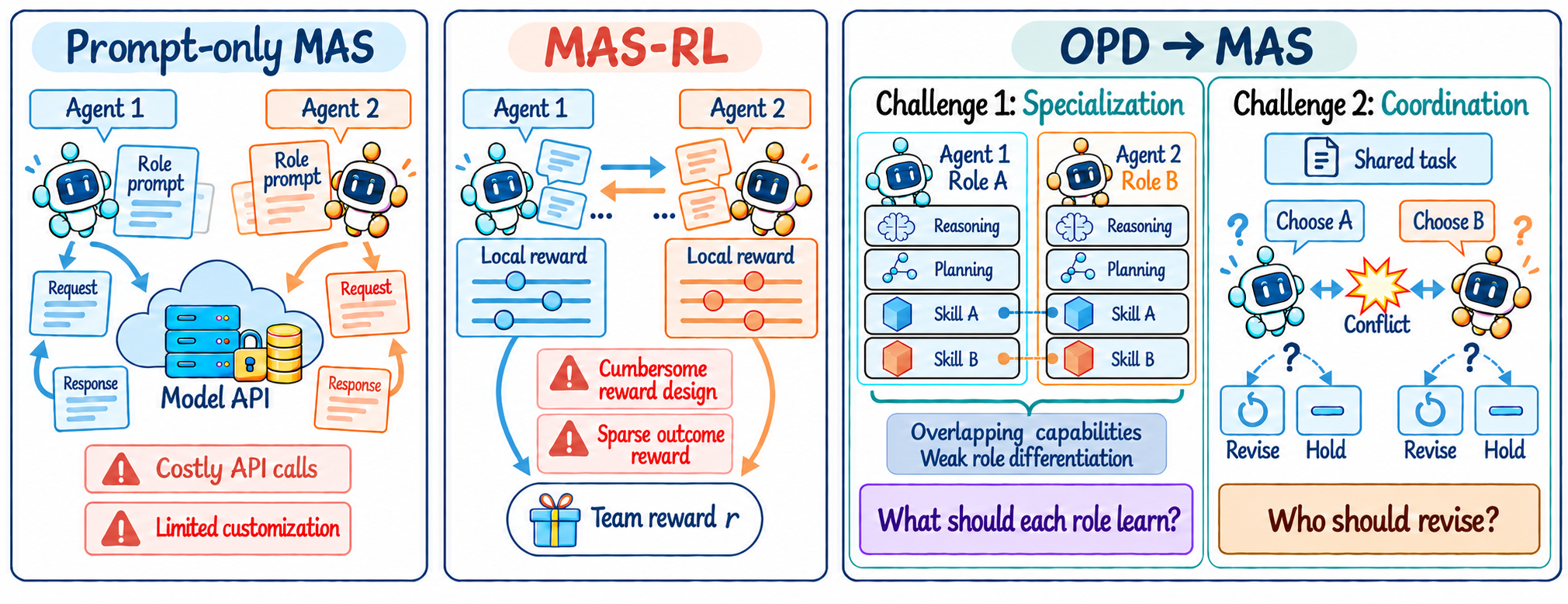}
\caption{\textbf{Motivation of \ourmodel.} Prompt-only MASs are costly and hard to customize, while MAS-RL suffers from sparse outcome rewards and cumbersome reward design. Extending OPD to MASs raises two challenges: building complementary role specialization while preserving shared knowledge, and turning cross-agent information into supervision for joint decisions and coordination.}
\label{fig:motivation}
\vspace{-0.25 in}
\end{figure*}

\textbf{C1: judging which role a high-quality behavior belongs to and building complementary specialization from that judgement while preserving the knowledge that all roles need.}
Single-agent OPD transfers the whole capability of the teacher into one student, whereas a MAS must decompose it across agents with distinct responsibilities.
A general-purpose teacher holds solving, verification and revision abilities at once, so the supervision it gives one role can cross role boundaries and lead several students to absorb the same generic knowledge.
For capacity-limited students this redundancy blurs the division of labour and consumes representational capacity that would otherwise serve role-specific ability~\citep{gudibande2024false,belcak2025slm_agentic}.

\textbf{C2: turning cross-agent collaborative information into supervision that OPD can exploit, so that distillation improves joint decision making and collaboration directly.}
The performance of a MAS depends not only on the individual ability of each agent but also on whether their behaviors stay coordinated across turns~\citep{pan2025whydomasfail,smit2024mad}.
When the outputs of different roles conflict, the agents have to identify the source of that conflict together and settle which of them should revise and which should hold, and improving the local behavior of each agent separately gives no guarantee that such decisions remain consistent at the system level.

We present \ourmodel, a multi-agent on-policy distillation framework for role specialization and collaborative learning, illustrated in~\Cref{fig:framework}.
For C1 we propose \textbf{Role-Advantage Specialization (RAS)}, which evaluates the same on-policy student token under the target role condition and under a non-target role condition and defines the difference between the two distillation signals as the role advantage.
Standard OPD judges whether a behavior is worth learning and the role advantage further judges which role it suits, so role-specific behavior is strengthened and cross-role behavior is suppressed while the ability shared by all roles is preserved.
For C2 we propose \textbf{Privileged Attribution for Coordination (PAC)}, which uses the reference solutions and verification results available during training to attribute an interaction conflict to its source and supplies that attribution to the teacher alone as privileged information.
The teacher accordingly gives mutually consistent token-level supervision for the on-policy outputs of all agents while every student still sees only the raw environment feedback available at deployment, so the collaborative decision is internalized into the policies themselves.
These two designs extend OPD from transferring capability into a single policy to jointly training role division and collaboration.
Extensive experiments on code generation and mathematical reasoning benchmarks show that \ourmodel improves task performance and leads the agents to develop clearer role specialization and more effective collaboration.

\section{Preliminaries}\label{sec:preliminaries}

\subsection{Multi-Agent Language Model Systems}\label{sec:mas}

We consider a multi-agent system of $N$ language model agents, in which agent $i$ takes a predefined role $r_i$ and carries an independent policy $\pi_{\theta_i}$ initialized from the same base model~\citep{ye2025xmas,strongermas}.

At interaction turn $t$, the task description $q$, the environment feedback $e^{t-1}$ and the interaction history $h^t$ form the role-neutral state shared by all agents,
\begin{equation}
s^t=(q,e^{t-1},h^t).
\label{eq:state}
\end{equation}
Agent $i$ instantiates it with its role into an input $x_i^t=P_i(s^t,r_i)$ through a prompt template $P_i$ and generates a response
\begin{equation}
y_i^t=(y_{i,t,1},\ldots,y_{i,t,L_i^t})\sim\pi_{\theta_i}(\cdot\mid x_i^t),
\label{eq:response}
\end{equation}
which is submitted to the environment as a single macro-action~\citep{liao2025marft,feng2025group}.
The environment returns feedback $e^t$ from the joint output $\mathbf y^t=(y_1^t,\ldots,y_N^t)$ and updates the interaction history, so a task execution yields
\begin{equation}
\tau=\left\{(x_i^t,y_i^t)\right\}_{t=0,i=1}^{T-1,N}\sim\prod_{i=1}^{N}\pi_{\theta_i},
\label{eq:trajectory}
\end{equation}
with $T$ the maximum number of turns.
Since the training context of each agent depends on all current policies, $\tau$ is a joint on-policy trajectory rather than a concatenation of independent ones.

We focus on a parallel multi-turn workflow of two complementary roles~\citep{cure,du2023llmdebate}: the agents produce their outputs independently and then revise from the disagreement the environment reports, terminating once the two agree or the turn limit is reached.

\subsection{On-Policy Distillation}\label{sec:opd}

On-policy distillation (OPD) transfers teacher knowledge on the states the current student visits~\citep{opd,lu2025onpolicydistillation}: the student policy $\pi_\theta$ samples a response $y$ in a context $x$, and a frozen teacher $\pi_T$ force-decodes it without generating a trajectory of its own.
Its objective is the reverse KL~\citep{gu2024minillm}, optimized through a surrogate weighted by the token-level advantage~\citep{yang2026learning,srpo}
\begin{equation}
A_k^{\mathrm{OPD}}=\log\pi_T(y_k\mid x,y_{<k})-\log\pi_\theta(y_k\mid x,y_{<k}),
\label{eq:opd_adv}
\end{equation}
namely
\begin{equation}
\mathcal{L}_{\mathrm{OPD}}=-\mathbb{E}_{y\sim\pi_\theta(\cdot\mid x)}\left[\frac{1}{|\mathcal I(y)|}\sum_{k\in\mathcal I(y)}\operatorname{sg}\!\left(A_k^{\mathrm{OPD}}\right)\log\pi_\theta(y_k\mid x,y_{<k})\right],
\label{eq:opd_loss}
\end{equation}
where $\mathcal I(y)$ is the set of token positions generated by the student and $\operatorname{sg}(\cdot)$ denotes the stop-gradient operator.
This paradigm targets a single student, whose context the teacher shares.

\section{Methodology}\label{sec:method}

\begin{figure*}[t]
\centering
\includegraphics[width=1.0\linewidth]{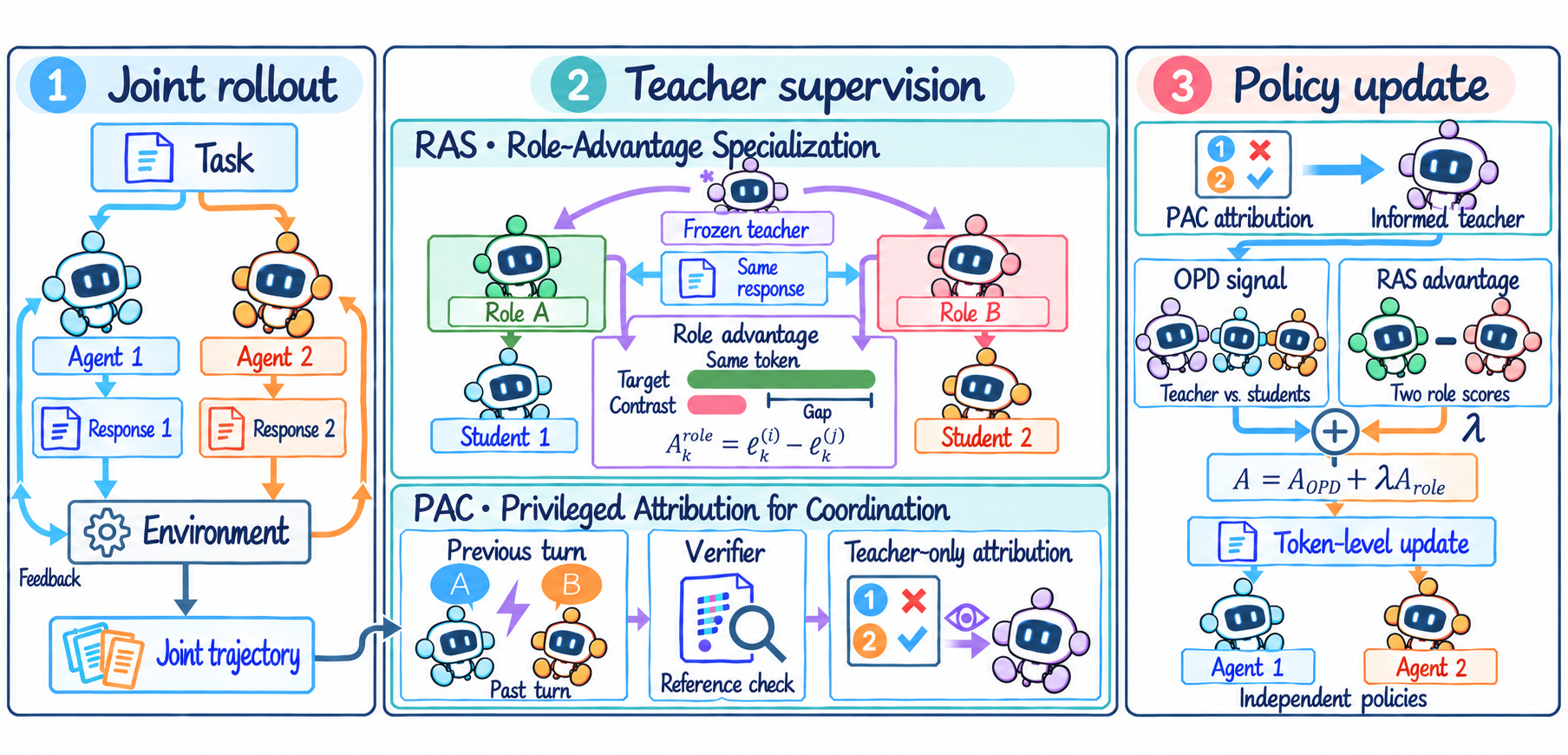}
\caption{
\textbf{Overview of \ourmodel.}
\ourmodel consists of three stages:
\emph{(1) joint rollout}, where agents interact with the environment to produce an on-policy joint trajectory;
\emph{(2) teacher supervision}, where RAS derives role-specific supervision by contrasting teacher scores across roles and PAC provides teacher-only privileged attribution for coordination;
and \emph{(3) policy update}, where the OPD and role-advantage signals are combined into token-level updates of the independent agent policies.
}
\label{fig:framework}
\vspace{-0.09 in}
\end{figure*}

\subsection{Role-Conditioned Multi-Agent Distillation}\label{sec:role_conditioned}

We first extend OPD to a multi-agent system, in the three stages drawn in~\Cref{fig:framework}.
The role policies produce the on-policy joint trajectory $\tau$ of~\Cref{sec:mas}, in which every response has a definite generator, so the teacher supervises them one by one and routes the gradient accordingly.

Since the agents carry different responsibilities, the teacher cannot evaluate all outputs under a single identity.
The advantage compares the teacher and the student on the same tokens, so the teacher is conditioned to produce the response rather than to judge it and its input is the context of the student itself with the role made explicit:
\begin{equation}
\text{teacher input}=u\oplus r_i,
\label{eq:teacher_input}
\end{equation}
where $u$ is the student input $x_i^t$ with the role condition removed, so it carries the state $s^t$ together with the response format supplied by $P_i$, while $r_i$ states only that role's boundary of responsibility.

To lighten the notation we fix the agent $i$ and the interaction turn when discussing a single response below, and keep only the token index $k$.
Under the state $u$ and the role $r_i$, the teacher scores the token the student actually generated as
\begin{equation}
\ell_k^{(i)}(u)=\log\pi_T(y_k\mid u,r_i,y_{<k}),
\label{eq:teacher_score}
\end{equation}
and the corresponding token-level advantage is
\begin{equation}
A_k^{\mathrm{OPD}}(u)=\ell_k^{(i)}(u)-\log\pi_{\theta_i}(y_k\mid x_i,y_{<k}).
\label{eq:mas_opd_adv}
\end{equation}
The teacher and the student are therefore conditioned on the same state $s^t$ in this form and differ only in that the role is supplied to the teacher as a separable condition, so nothing beyond what the student itself read enters the teacher.
The teacher thereby provides dense token-level supervision for every role at every turn, and the roles improve together within the joint interaction.

\subsection{Role-Advantage Specialization}\label{sec:ras}

The role condition acts on where the teacher places its emphasis and does not alter the composition of its abilities, so a general-purpose teacher passes on the abilities of every role in much the same way whichever role it is conditioned on, and the differentiation the policies acquire is limited to what a change of emphasis can produce.
Sharpening it requires the distillation signal itself to carry the role attribution of a behavior.

We observe that the decomposition of~\Cref{sec:role_conditioned} already lets the teacher evaluate the same output under different identities, so this attribution can be characterized by the internal difference between the evaluations of the teacher.
Specifically, on the same state, prefix and token, the teacher produces a second evaluation $\ell_k^{(j)}(u)$ under a non-target role $r_j$, the only difference between the two evaluations being the role the teacher takes.
Both carry the same student log-probability, so subtracting one from the other cancels the student term and yields the \textbf{role advantage}
\begin{equation}
A_k^{\mathrm{role}}(u)=\ell_k^{(i)}(u)-\ell_k^{(j)}(u).
\label{eq:role_adv}
\end{equation}

The cancellation of the student term makes this signal entirely determined by the frozen teacher and independent of the current ability of the student, so its value reflects nothing but the judgement of the teacher about role attribution.
A positive $A_k^{\mathrm{role}}$ means the behavior is favored more along the direction of the current role and should receive additional reinforcement; a value close to zero means it belongs to the knowledge every role needs and it is still transferred as usual; a negative value means it lies closer to the responsibility of the teammate.
Combining the two signals, the training signal for token $k$ is
\begin{equation}
A_k^{\mathrm{RAS}}(u)=A_k^{\mathrm{OPD}}(u)+\lambda A_k^{\mathrm{role}}(u),\qquad\lambda\ge0,
\label{eq:ras}
\end{equation}
where $\lambda$ controls the strength of the differentiation signal and $\lambda=0$ recovers the form of~\Cref{sec:role_conditioned}.
The contrasting role is the one whose responsibility the target role is told not to take over, determined in general by~\Cref{app:scaling_ras}.
Because the direction of differentiation is determined entirely by the role preference of the teacher, RAS produces an additional update only where the teacher expresses a definite role difference, and leaves tokens the teacher scores alike under both role conditions to the original distillation signal.
Where the signal falls within a response is shown token by token in~\Cref{app:case_ras}.

\subsection{Privileged Attribution for Coordination}\label{sec:pac}

A multi-agent system corrects itself across turns, and the decisive moment arrives when a conflict appears: the workflow reports the disagreement without indicating where it originates, and each agent has to judge for itself where the error lies and settle whether to revise or to hold its current output.
We turn this decision into distillable supervision by providing that criterion to the teacher during training.

We assume a task verifier $V$ is accessible during training, taking the form of any automatic mechanism able to decide the correctness of an output.
The verifier only has to locate the output at fault, and never has to be turned into a scalar reward for a role or a turn, a distinction~\Cref{app:reward_vs_attribution} sets against the reward design the baselines require.
PAC uses this verifier to attribute the joint behavior,
\begin{equation}
a^t=V(q,\mathbf y^t,e^t),
\label{eq:attribution}
\end{equation}
where $a^t$ records how the responsibility for the conflict falls and the evidence supporting that judgement.
We render $a^t$ into a structured privileged description $c^t$ stating verifiable facts alone, with no instruction addressed to any particular role, so that either role identity derives its own action from the same facts.

A turn is attributed only once it is complete, so the $c^{t'}$ obtained from $\mathbf y^{t'}$ becomes available for turn $t'+1$ and the teacher context is never contaminated by information determined by the response it is about to score.
The history the teacher reads therefore carries the attributions of all completed turns, written $c^{<t}=(c^{0},\ldots,c^{t-1})$.
The context of the teacher in~\Cref{sec:role_conditioned} is thereby reconstructed from the student-visible $u$ into
\begin{equation}
\widetilde u=u\oplus c^{<t},
\label{eq:priv_state}
\end{equation}
which retains the state and the response format that $u$ carries and adds the attributions of the completed turns, while the student still generates its response from the original input $x_i^t$ alone.
The set $c^{<t}$ is empty at the initial turn, where no turn has yet been judged, so PAC takes effect from the second turn onward without needing a separate case.
The teacher thus reads a history in which every past turn carries a verdict on how the responsibility fell while the student observes the same turns with the verdicts withheld, and the preceding signals are evaluated at $\widetilde u$, so PAC introduces no new loss term and instead changes the values of the same signals by reconstructing the teacher context.
Reducing the corresponding loss therefore requires the student to reproduce from that raw feedback alone the behavior the teacher produces with the attribution in hand, which is the pressure that drives the attribution judgement and the coordination policy it implies into the policy itself, and~\Cref{sec:rq3} measures how far the trained students reach that judgement once the verifier is gone.
The verifier and the privileged description are removed after training, and the roles remain able to reach mutually compatible decisions from their own observations, as traced turn by turn in~\Cref{app:casestudy}.

\subsection{Joint Training Objective}\label{sec:objective}

The teacher occupies the context $\widetilde u$ of~\Cref{eq:priv_state} at every turn, the initial one included, where the set of attributions it carries is still empty, and evaluates the same student response under the target role and under a non-target role to yield the token-level signal $A_{i,t,k}^{\mathrm{RAS}}(\widetilde u)$, which drives the update of the \mbox{student policies:}
\begin{equation}
\mathcal{L}_{\text{\ourmodel}}=-\frac{1}{N}\sum_{i=1}^{N}\frac{1}{|\mathcal T_i|}\sum_{t\in\mathcal T_i}\frac{1}{L_i^t}\sum_{k=1}^{L_i^t}\operatorname{sg}\!\left(A_{i,t,k}^{\mathrm{RAS}}(\widetilde u)\right)\log\pi_{\theta_i}\!\left(y_{i,t,k}\mid x_i^t,y_{i,t,<k}\right),
\label{eq:mas_opd_loss}
\end{equation}
where $\mathcal T_i$ is the set of effective interaction turns of role $i$.

\begin{table*}[t]
    \centering
    \caption{
    \textbf{Overall performance of \ourmodel on code generation and mathematical reasoning.}
    We report the mean performance and standard deviation over 5 independent
    runs, separately seeded training runs for every trained method and
    repeated evaluations for the three prompt-only baselines.
    Within each scale, the highest mean in each column is in \textbf{bold}
    and the second highest is \underline{underlined}.
    }
    \vspace{1mm}
    \label{tab:main_results}

    \begingroup
    \setlength{\tabcolsep}{4.0pt}
    \renewcommand{\arraystretch}{0.95}
    \normalsize

    \begin{adjustbox}{max width=0.98\textwidth}
    \begin{tabular}{lcccccc}
        \toprule

        & \multicolumn{3}{c}{\textbf{Code}}
        & \multicolumn{3}{c}{\textbf{Math}} \\
        \cmidrule(lr){2-4}
        \cmidrule(lr){5-7}

        {\footnotesize\textbf{Method}}
        & {\footnotesize\textbf{LiveCodeBench}}
        & {\footnotesize\textbf{APPS}}
        & {\footnotesize\textbf{CodeContests}}
        & {\footnotesize\textbf{AIME24}}
        & {\footnotesize\textbf{AIME25}}
        & {\footnotesize\textbf{OlympiadBench}} \\

        \midrule

        \rowcolor[HTML]{FEE090}
        \multicolumn{7}{c}{
            \textbf{Teacher Model from the Qwen3-14B Series}
        } \\

        SA
        & 26.86 {\scriptsize $\pm$ 1.51}
        & 36.28 {\scriptsize $\pm$ 1.17}
        & 16.73 {\scriptsize $\pm$ 1.69}
        & 29.33 {\scriptsize $\pm$ 3.65}
        & 23.33 {\scriptsize $\pm$ 4.08}
        & 59.11 {\scriptsize $\pm$ 0.67} \\

        \midrule
        \midrule

        \rowcolor[HTML]{E0F3F8}
        \multicolumn{7}{c}{
            \textbf{Student Models from the Qwen3-1.7B Series}
        } \\

        SA + ST
        & 16.69 {\scriptsize $\pm$ 0.85}
        & 16.44 {\scriptsize $\pm$ 1.31}
        & 3.76 {\scriptsize $\pm$ 1.31}
        & 7.33 {\scriptsize $\pm$ 3.65}
        & 1.33 {\scriptsize $\pm$ 2.98}
        & 17.33 {\scriptsize $\pm$ 0.72} \\

        SA + MT
        & 12.80 {\scriptsize $\pm$ 1.32}
        & 10.56 {\scriptsize $\pm$ 0.73}
        & 1.21 {\scriptsize $\pm$ 1.87}
        & 2.67 {\scriptsize $\pm$ 2.79}
        & 0.00 {\scriptsize $\pm$ 0.00}
        & 12.49 {\scriptsize $\pm$ 1.15} \\

        SA + ST + GRPO
        & 19.54 {\scriptsize $\pm$ 1.02}
        & 17.24 {\scriptsize $\pm$ 1.09}
        & 3.15 {\scriptsize $\pm$ 1.08}
        & 6.00 {\scriptsize $\pm$ 4.35}
        & 2.67 {\scriptsize $\pm$ 1.49}
        & 18.55 {\scriptsize $\pm$ 0.53} \\

        SA + MT + GRPO
        & 17.71 {\scriptsize $\pm$ 1.51}
        & 13.32 {\scriptsize $\pm$ 0.92}
        & 1.70 {\scriptsize $\pm$ 2.16}
        & 4.00 {\scriptsize $\pm$ 1.49}
        & 0.67 {\scriptsize $\pm$ 1.49}
        & 14.27 {\scriptsize $\pm$ 0.94} \\

        \cmidrule(lr){1-7}

        MAS
        & 20.69 {\scriptsize $\pm$ 0.75}
        & 20.04 {\scriptsize $\pm$ 0.55}
        & 6.55 {\scriptsize $\pm$ 1.57}
        & 14.00 {\scriptsize $\pm$ 2.79}
        & 8.00 {\scriptsize $\pm$ 3.80}
        & 23.18 {\scriptsize $\pm$ 0.42} \\

        MAGRPO
        & 21.26 {\scriptsize $\pm$ 1.78}
        & 19.56 {\scriptsize $\pm$ 1.23}
        & 6.06 {\scriptsize $\pm$ 0.86}
        & 16.67 {\scriptsize $\pm$ 4.08}
        & 9.33 {\scriptsize $\pm$ 4.94}
        & 22.67 {\scriptsize $\pm$ 0.34} \\

        MAS + GRPO
        & 22.63 {\scriptsize $\pm$ 1.32}
        & 21.16 {\scriptsize $\pm$ 0.68}
        & \underline{8.24} {\scriptsize $\pm$ 1.69}
        & 17.33 {\scriptsize $\pm$ 3.65}
        & \underline{13.33} {\scriptsize $\pm$ 2.36}
        & 25.40 {\scriptsize $\pm$ 0.81} \\

        CURE
        & 17.03 {\scriptsize $\pm$ 1.91}
        & 15.32 {\scriptsize $\pm$ 0.46}
        & 3.39 {\scriptsize $\pm$ 1.40}
        & --
        & --
        & -- \\

        MARFT
        & \underline{23.66} {\scriptsize $\pm$ 0.96}
        & \underline{22.32} {\scriptsize $\pm$ 0.82}
        & 8.00 {\scriptsize $\pm$ 1.98}
        & \underline{19.33} {\scriptsize $\pm$ 4.94}
        & 12.67 {\scriptsize $\pm$ 3.65}
        & \underline{27.09} {\scriptsize $\pm$ 0.62} \\

        \cmidrule(lr){1-7}

        \textbf{\ourmodel}
        & \textbf{27.77} {\scriptsize $\pm$ 0.96}
        & \textbf{26.00} {\scriptsize $\pm$ 0.75}
        & \textbf{11.52} {\scriptsize $\pm$ 1.13}
        & \textbf{21.33} {\scriptsize $\pm$ 1.83}
        & \textbf{14.00} {\scriptsize $\pm$ 1.49}
        & \textbf{30.42} {\scriptsize $\pm$ 0.55} \\

        \midrule
        \midrule

        \rowcolor[HTML]{E0F3F8}
        \multicolumn{7}{c}{
            \textbf{Student Models from the Qwen3-4B Series}
        } \\

        SA + ST
        & 20.11 {\scriptsize $\pm$ 1.10}
        & 28.96 {\scriptsize $\pm$ 0.74}
        & 11.15 {\scriptsize $\pm$ 1.33}
        & 19.33 {\scriptsize $\pm$ 2.79}
        & 18.00 {\scriptsize $\pm$ 5.06}
        & 47.21 {\scriptsize $\pm$ 0.68} \\

        SA + MT
        & 16.23 {\scriptsize $\pm$ 0.77}
        & 19.52 {\scriptsize $\pm$ 1.12}
        & 4.97 {\scriptsize $\pm$ 0.90}
        & 12.67 {\scriptsize $\pm$ 4.35}
        & 13.33 {\scriptsize $\pm$ 2.36}
        & 40.95 {\scriptsize $\pm$ 0.91} \\

        SA + ST + GRPO
        & 23.31 {\scriptsize $\pm$ 1.42}
        & 34.84 {\scriptsize $\pm$ 0.62}
        & 9.82 {\scriptsize $\pm$ 1.63}
        & 22.67 {\scriptsize $\pm$ 1.49}
        & 22.00 {\scriptsize $\pm$ 3.80}
        & 49.02 {\scriptsize $\pm$ 0.48} \\

        SA + MT + GRPO
        & 21.60 {\scriptsize $\pm$ 0.94}
        & 30.04 {\scriptsize $\pm$ 0.96}
        & 8.48 {\scriptsize $\pm$ 2.10}
        & 20.00 {\scriptsize $\pm$ 4.08}
        & 19.33 {\scriptsize $\pm$ 1.49}
        & 45.96 {\scriptsize $\pm$ 1.08} \\

        \cmidrule(lr){1-7}

        MAS
        & 24.57 {\scriptsize $\pm$ 1.67}
        & 33.56 {\scriptsize $\pm$ 1.27}
        & 13.94 {\scriptsize $\pm$ 0.74}
        & 25.33 {\scriptsize $\pm$ 2.98}
        & 24.67 {\scriptsize $\pm$ 4.47}
        & 53.06 {\scriptsize $\pm$ 0.59} \\

        MAGRPO
        & 25.37 {\scriptsize $\pm$ 1.25}
        & 34.96 {\scriptsize $\pm$ 0.52}
        & 14.67 {\scriptsize $\pm$ 1.17}
        & 31.33 {\scriptsize $\pm$ 3.80}
        & 27.33 {\scriptsize $\pm$ 2.79}
        & 54.36 {\scriptsize $\pm$ 0.77} \\

        MAS + GRPO
        & \underline{28.23} {\scriptsize $\pm$ 1.96}
        & 37.88 {\scriptsize $\pm$ 0.88}
        & \underline{17.82} {\scriptsize $\pm$ 1.85}
        & 30.67 {\scriptsize $\pm$ 4.94}
        & \underline{34.00} {\scriptsize $\pm$ 3.65}
        & 57.30 {\scriptsize $\pm$ 0.35} \\

        CURE
        & 27.09 {\scriptsize $\pm$ 0.87}
        & 37.00 {\scriptsize $\pm$ 1.36}
        & 16.36 {\scriptsize $\pm$ 1.29}
        & --
        & --
        & -- \\

        MARFT
        & 27.77 {\scriptsize $\pm$ 1.54}
        & \underline{39.04} {\scriptsize $\pm$ 1.04}
        & 17.09 {\scriptsize $\pm$ 1.00}
        & \underline{32.67} {\scriptsize $\pm$ 3.65}
        & 32.00 {\scriptsize $\pm$ 1.83}
        & \underline{58.52} {\scriptsize $\pm$ 1.13} \\

        \cmidrule(lr){1-7}

        \textbf{\ourmodel}
        & \textbf{32.11} {\scriptsize $\pm$ 0.94}
        & \textbf{41.72} {\scriptsize $\pm$ 0.69}
        & \textbf{20.00} {\scriptsize $\pm$ 1.29}
        & \textbf{33.33} {\scriptsize $\pm$ 2.36}
        & \textbf{36.67} {\scriptsize $\pm$ 2.36}
        & \textbf{60.24} {\scriptsize $\pm$ 0.50} \\

        \bottomrule
    \end{tabular}
    \end{adjustbox}

    \endgroup

    \vspace{-10pt}
\end{table*}

We take an unweighted average over tokens within a response, over turns within a role and over roles within the system, so that longer responses or trajectories do not dominate it.
The gradient of role $i$ updates $\theta_i$ only, and the policies update synchronously once a batch of joint rollouts completes.

RAS and PAC are accordingly not two separate loss terms but act on the two sides of the interface established in~\Cref{sec:role_conditioned}: PAC reconstructs the context $\widetilde u$ occupied by the teacher through $c^{<t}$, and RAS varies the role condition of the teacher in that context and adds $\lambda A_{i,t,k}^{\mathrm{role}}$, with both feeding into the same token-level advantage and the whole framework introducing a single core hyperparameter $\lambda$.
Relative to~\Cref{sec:role_conditioned}, the additional cost is one teacher force decoding under the non-target role per student response, and~\Cref{app:algorithm} sets out the order of a training step.

\section{Experiment}\label{sec:experiment}

\subsection{Experimental Setup}
\label{sec:setup}

\paragraph{Datasets \& Benchmarks.}
The code domain trains on CodeContests~\citep{li2022competition} and the mathematics domain on Polaris-Dataset-53K~\citep{polaris2025}, both described in~\Cref{app:training_data}, and we evaluate on six benchmarks described one by one in~\Cref{app:benchmarks}.
\emph{Code generation}: (1) LiveCodeBench-v6~\citep{jain2025livecodebench}, (2) APPS~\citep{hendrycks2021apps} and (3) CodeContests~\citep{li2022competition}, evaluated on the test split held out from the training corpus of the same name.
\emph{Mathematical reasoning}: (4) AIME 2024, (5) AIME 2025 and (6) OlympiadBench~\citep{he2024olympiadbench}.

\paragraph{Experiment \& Evaluation Setups.}
All models are from the Qwen3 series~\citep{qwen2025qwen3}: the teacher is Qwen3-14B and the students are Qwen3-1.7B and Qwen3-4B, and an experiment reports the 1.7B student unless stated otherwise.
We report Pass@1 for code generation and accuracy for mathematical reasoning, and~\Cref{app:models_eval} gives the use of the teacher, the inference mode, the decoding settings and the interaction horizon $T$ of~\Cref{sec:mas}.

\paragraph{Baselines \& Implementation Details.}
\emph{Single-agent baselines}: (1) SA + ST, (2) SA + MT, (3) SA + ST + GRPO and (4) SA + MT + GRPO.
\emph{Multi-agent baselines}: (5) MAS, (6) MAGRPO~\citep{magrpo}, (7) MAS + GRPO, (8) CURE~\citep{cure} and (9) MARFT~\citep{liao2025marft}, of which CURE is reported on the code domain alone.
\Cref{app:baselines} describes the nine one by one and~\Cref{app:implementation} gives the implementation details.

\subsection{Overall Performance}\label{sec:rq1}

\Cref{tab:main_results} reports code generation and mathematical reasoning results for two student scales against single-agent and multi-agent baselines, where SA + ST denotes a single agent answering in one turn and SA + MT a single agent revising its own output across turns.

\begin{itemize}[leftmargin=*]

\item \textbf{Obs 1: \ourmodel attains the highest mean score on every benchmark at both student scales.}
It improves on the strongest baseline of each column by more than two points on average, and remains the top-performing method from 1.7B to 4B, over which the gain on the untrained MAS falls from more than forty percent to close to thirty in relative terms while growing in absolute ones, the smaller student having the most to gain from supervision that separates role-specific behavior from shared ability.
The 4B system moreover exceeds the 14B teacher on all six benchmarks, which is possible because the teacher is queried as a single agent whereas the students are trained as a collaborating pair, so the distilled system exploits an interaction the teacher never performs.

\item \textbf{Obs 2: the gain tracks cross-agent interaction rather than additional turns.}
Letting one agent revise its own output is consistently harmful, since SA + MT falls below SA + ST on every benchmark at both scales and the ordering survives GRPO training.
In the two workflows we run, extra turns help only when shared between two roles, as the prompt-only MAS improves on the prompt-only single agent on every benchmark without any training.

\item \textbf{Obs 3: outcome-reward training improves a multi-agent system unevenly and with a wide run-to-run spread.}
MAGRPO falls below the untrained MAS on three of the six benchmarks at 1.7B while improving on the remaining three, and the reinforcement learning baselines reach standard deviations near five points on the AIME benchmarks, whose thirty problems make each one worth $3.3$ points.
\ourmodel improves on the untrained MAS in all twelve cells and gives the smallest spread among the multi-agent methods on AIME24 at both scales, which is consistent with replacing that scalar signal with dense per-token targets.

\end{itemize}

\subsection{Role Specialization}\label{sec:rq2}

\begin{table}[t]
    \centering
    \caption{
    \textbf{Contribution of the two modules.}
    Accuracy of the 1.7B student when each module is removed from \ourmodel, with every other setting as in~\Cref{app:implementation_common}.
    Removing the role advantage sets $\lambda=0$; removing privileged attribution restores the student-visible context.
    }
    \vspace{1mm}
    \label{tab:ablation}

    \begingroup
    \setlength{\tabcolsep}{4.0pt}
    \renewcommand{\arraystretch}{1.08}
    \normalsize

    \begin{adjustbox}{max width=0.90\columnwidth}
    \begin{tabular}{lcccccc}
        \toprule

        & \multicolumn{3}{c}{\textbf{Code}}
        & \multicolumn{3}{c}{\textbf{Math}} \\
        \cmidrule(lr){2-4}
        \cmidrule(lr){5-7}

        {\footnotesize\textbf{Method}}
        & {\footnotesize\textbf{LiveCodeBench}}
        & {\footnotesize\textbf{APPS}}
        & {\footnotesize\textbf{CodeContests}}
        & {\footnotesize\textbf{AIME24}}
        & {\footnotesize\textbf{AIME25}}
        & {\footnotesize\textbf{OlympiadBench}} \\

        \midrule

        \textbf{\ourmodel}
        & \textbf{27.77} {\scriptsize $\pm$ 0.96}
        & \textbf{26.00} {\scriptsize $\pm$ 0.75}
        & \textbf{11.52} {\scriptsize $\pm$ 1.13}
        & \textbf{21.33} {\scriptsize $\pm$ 1.83}
        & \textbf{14.00} {\scriptsize $\pm$ 1.49}
        & \textbf{30.42} {\scriptsize $\pm$ 0.55} \\

        \cmidrule(lr){1-7}

        \quad w/o RAS
        & 26.51 {\scriptsize $\pm$ 1.18}
        & 23.92 {\scriptsize $\pm$ 0.61}
        & 9.33 {\scriptsize $\pm$ 0.92}
        & 20.00 {\scriptsize $\pm$ 2.36}
        & 13.33 {\scriptsize $\pm$ 2.36}
        & 29.67 {\scriptsize $\pm$ 0.41} \\

        \quad w/o PAC
        & 25.94 {\scriptsize $\pm$ 1.04}
        & 25.00 {\scriptsize $\pm$ 0.63}
        & 10.55 {\scriptsize $\pm$ 0.69}
        & 18.67 {\scriptsize $\pm$ 1.83}
        & 12.00 {\scriptsize $\pm$ 1.83}
        & 28.69 {\scriptsize $\pm$ 0.45} \\

        \bottomrule
    \end{tabular}
    \end{adjustbox}

    \endgroup

    \vspace{-6pt}
\end{table}

\begin{figure*}[t]
    \centering
    \includegraphics[width=\textwidth]{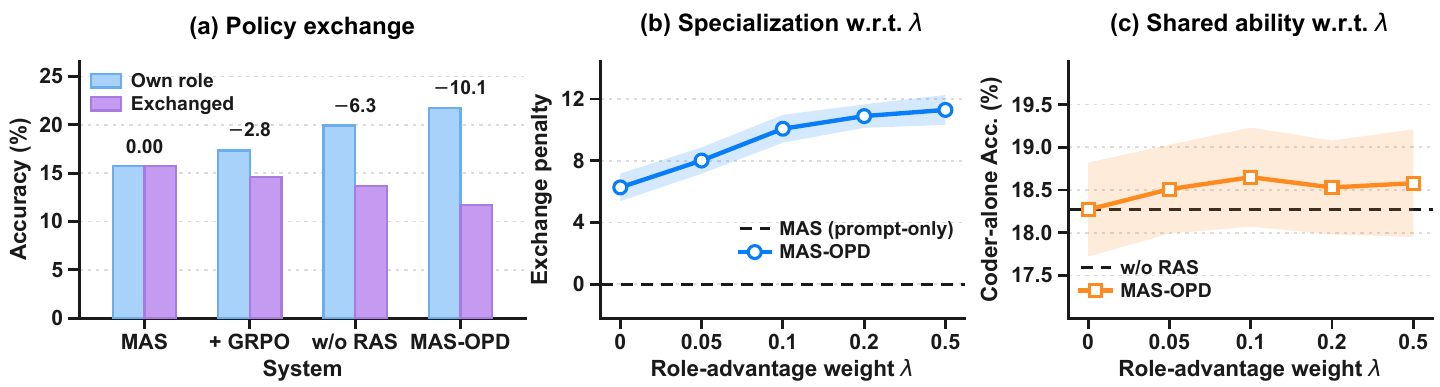}
    \caption{
    \textbf{Role specialization of \ourmodel and its dependence on the role-advantage weight $\lambda$.}
    All three report the 1.7B student on the three code benchmarks, with the Coder and the Tester as the two roles.
    The figure has three parts:
    \emph{(a) policy exchange}, giving the accuracy of four systems under their own and exchanged assignments, annotated with the loss;
    \emph{(b) specialization w.r.t.\ $\lambda$}, tracking that loss as $\lambda$ varies;
    and \emph{(c) Coder-alone accuracy}, under the single-agent single-turn protocol of~\Cref{tab:main_results}.
    Here w/o RAS is \ourmodel with $\lambda=0$, and shaded bands give the standard deviation over five runs.
    }
    \label{fig:role_spec}
    \vspace{-6pt}
\end{figure*}

\begin{itemize}[leftmargin=*]

\item \textbf{Obs 4: both modules contribute and neither substitutes for the other.}
Removing either module costs accuracy on all six benchmarks in~\Cref{tab:ablation}, so the two accumulate rather than stand in for one another, as expected of modules acting on the two sides of the single interface of~\Cref{sec:role_conditioned}.
The load is divided rather than shared: the role advantage accounts for the larger part of the gap on code and privileged attribution for the larger part on mathematics, as two modules addressing different failures would produce, and~\Cref{sec:rq3} reads the same table for its own purposes.

\item \textbf{Obs 5: the trained policies are not interchangeable, and they are already so before the role-advantage term is applied.}
Every trained system in panel (a) of~\Cref{fig:role_spec} loses accuracy when the two policies are exchanged, while the prompt-only system loses nothing, which rules out the reading that the penalty is an artifact of each role receiving a prompt written for the other.
Removing the role advantage leaves a system that still gives up much of its accuracy under exchange, its role conditioning already giving the teacher a role-dependent target, and \ourmodel gives up more still while reaching the highest own-role accuracy of the four systems.

\item \textbf{Obs 6: specialization grows with the weight on the role advantage, saturates beyond the reported setting, and leaves the Coder's own accuracy unchanged.}
Panel (b) shows the exchange penalty growing with $\lambda$ from the non-zero value already present at $\lambda=0$, and the growth is concentrated in the lower part of the range: the penalty gains $3.80$ points up to the reported setting and a further $1.22$ over the fourfold increase beyond it.
Over that same range panel (c) keeps the Coder's own accuracy within the run-to-run standard deviation of the $\lambda=0$ reference.
Since $\lambda$ multiplies nothing but the difference between the two teacher evaluations of the same response, that difference drives the separation rather than a property accompanying training.
The role advantage vanishes wherever the teacher evaluates a token alike under both role conditions, leaving the tokens carrying shared competence as the module would leave them (\Cref{sec:ras}).

\end{itemize}

\subsection{Privileged Attribution}\label{sec:rq3}

\suppressfloats[t]

\begin{figure*}[t]
    \centering
    \includegraphics[width=\textwidth]{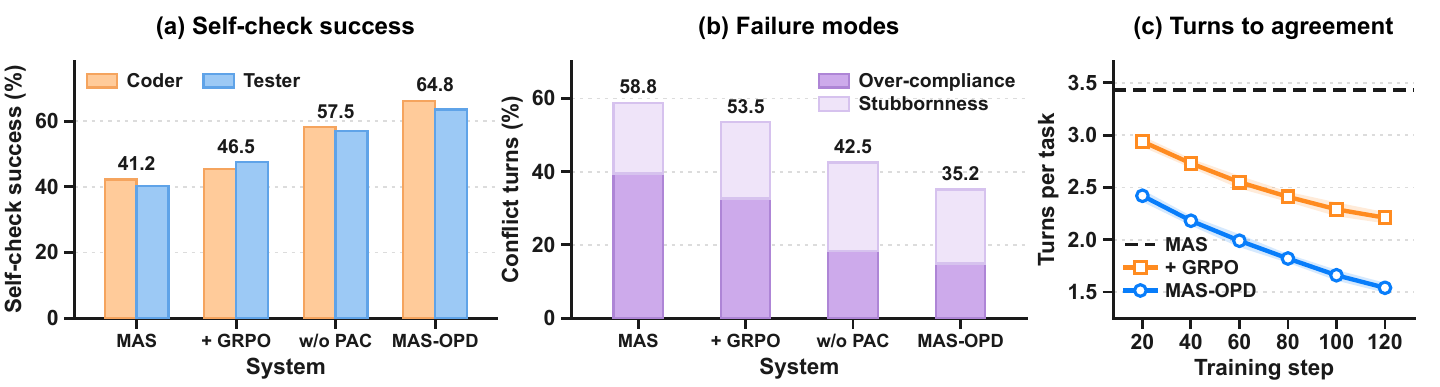}
    \caption{
    \textbf{Internalization of the attribution judgement and its effect on how quickly the roles agree.}
    All three report the 1.7B student on the three code benchmarks.
    The first two consider the turns in which the two roles disagree, \emph{(a) self-check success} giving the share in which a role revises exactly when its own output is wrong and \emph{(b) failure modes} the remainder, while \emph{(c) turns to agreement} tracks the turns a task consumes.
    No model sees the attribution at evaluation, w/o PAC restores the student-visible context of~\Cref{sec:role_conditioned}, and bands give the standard deviation over five runs.
    }
    \label{fig:pac_analysis}
    \vspace{-6pt}
\end{figure*}

\begin{itemize}[leftmargin=*]

\item \textbf{Obs 7: the students resolve conflicts in the direction their own correctness warrants.}
Panel (a) of~\Cref{fig:pac_analysis} shows \ourmodel settling a larger share of conflicts correctly than the same method without privileged attribution, and both roles improve alike.
The two systems differ only in whether the teacher reads the privileged context, and neither is evaluated with the verifier, so the judgement the teacher had during training is one the students reach on their own afterwards.

\item \textbf{Obs 8: what training leaves behind is the opposite failure, and privileged attribution is what removes it.}
Panel (b) splits the conflicts each system mishandles into revising an output that was already correct and keeping one that was not.
Deference dominates before distillation and falls to less than half of its prompt-only level in the distilled systems, but the residue that survives runs the other way, and w/o PAC keeps an incorrect output more often than any other system in the figure.
Deciding whether to hold or to give way requires knowing which side erred, and \ourmodel lowers that residue by the larger of its two reductions while lowering the other as well, so the module does not make the roles more compliant or more persistent but improves when they should be either.
This underlies~\Cref{tab:ablation}: the role advantage and privileged attribution address different failures, which is why removing one is not compensated by the other.

\item \textbf{Obs 9: the roles reach agreement in fewer turns as training proceeds.}
Panel (c) shows the number of turns a task consumes falling over training for both trained systems and falling further for \ourmodel at every checkpoint, ending well short of the four turns the workflow allows, with the reductions growing smaller as training continues.
Since a task ends once the two roles agree, this is the trajectory-level counterpart of what panels (a) and (b) report turn by turn, and those two are what establish that the agreements reached are also better ones rather than merely earlier: roles that judge the source of a disagreement correctly spend fewer exchanges settling it.

\end{itemize}

\paragraph{Further experiments.}
Three further studies are reported in the appendix.
Replicating both role types symmetrically to build systems of up to eight agents raises the average accuracy of both domains at every step, with no sign of saturation, and the construction that lets both modules operate unchanged as agents are added is given in~\Cref{app:scaling}.
Sweeping the role-advantage weight $\lambda$ over a factor of four leaves accuracy stable, while removing the role advantage or raising the weight far above the reported setting both cost accuracy, as~\Cref{app:lambda_sweep} shows.
\Cref{app:cost} reports the measured per-step wall-clock cost of training, separated into the joint rollout and the update.

\section{Related Work}\label{sec:related}

We outline the two lines of work this paper builds on and give the full discussion, with the complete set of references, in~\Cref{app:related}.

\paragraph{RL Training for MAS.}
Multi-agent systems of language models are assembled at inference time~\citep{wu2023autogen}, and surveys report the effort stays there rather than in training~\citep{pan2025whydomasfail}.
Reinforcement learning is the standard alternative~\citep{shao2024deepseekmath}, and its multi-agent extensions restrict the interaction or port single-agent algorithms~\citep{cure,magrpo}, rewarding a trajectory or a turn and leaving the decisions inside without a target.

\paragraph{On-Policy Distillation.}
On-policy distillation supervises the trajectories a student itself produces, removing the mismatch between a fixed corpus and the states it visits~\citep{gu2024minillm,opd}, and later work reads its objective as reinforcement learning with teacher-supplied per-token rewards~\citep{li2026rethinking}.
It has been extended to teacher-free settings and used in place of a scalar reward~\citep{opsd,zhong2026sod}, yet always for one student, so neither attributing behavior to a role nor supervising a decision that depends on a teammate arises.

\section{Conclusion}\label{sec:conclusion}
We studied how to post-train a multi-agent system with on-policy distillation, which had been formulated for a single agent.
\ourmodel adds the role advantage, which keeps only what distinguishes a target from a non-target role condition, and privileged attribution, which resolves an interaction conflict with training-only information given to the teacher alone.
It attains the highest mean score on every benchmark at both student scales, and its agents specialize more sharply and collaborate more effectively than a prompt-only system.

\bibliography{conference}
\bibliographystyle{conference}

\newpage
\appendix

\crefalias{section}{appendix}
\crefalias{subsection}{appendix}
\crefalias{subsubsection}{appendix}

\section*{Appendix}
\addcontentsline{toc}{section}{Appendix}

\section{Limitations}\label{app:limitations}

Two boundaries of the study are worth making explicit.
The first is that all of the models involved come from a single open-weight family, chosen because it is released at many sizes with stable checkpoints and is used widely enough that the figures reported here can be set against published ones, which is what allows the comparison against nine baselines in~\Cref{tab:main_results} to rest on the method rather than on the backbone; whether the same conclusions hold across families is left to future work.
The second is that the teacher is held at one scale throughout the main experiments while the student varies, which is what makes each row of that table differ from the others in the student alone and keeps the effect attributable to it; how the size of the teacher interacts with what we report is likewise left to future work.

\section{Extended Related Work}\label{app:related}

This section expands the two lines of work summarized in~\Cref{sec:related}.

\paragraph{RL Training for MAS.}
Multi-agent systems built on large language models have largely been assembled at inference time, with frameworks such as AutoGen~\citep{wu2023autogen} and MetaGPT~\citep{hong2024metagpt} eliciting role-specific behavior from one frozen policy through prompt augmentation alone.
Since the competence of a single model varies considerably across the abilities a workflow calls for~\citep{chen2024ioa,wang2024moa,belcak2025slm_agentic}, a line of work instead assigns a separate and more suitable model to each role~\citep{ye2025xmas,belcak2025slm_agentic}, yet surveys of the area report that the design effort remains concentrated on inference-time orchestration rather than on training the policies themselves~\citep{pan2025whydomasfail,guo2024llmma}.
Reinforcement learning with group-relative and rule-based rewards has meanwhile become the standard route for post-training a language agent, and it improves reasoning, long-horizon planning, game playing and tool use in the single-agent setting~\citep{shao2024deepseekmath,feng2025group,wang2025ragen,Qian2025ToolRL,hu2025lmgame}.
Extending it to several agents has so far been done under restricted interaction patterns or fixed role structures: CURE~\citep{cure} co-evolves a coder and a unit tester under one shared policy for code generation, SPIRAL~\citep{liu2025spiral} trains a single model through self-play on zero-sum games, MHGPO~\citep{chen2025mhgpo} targets retrieval-augmented generation, and MAPoRL~\citep{park2025maporl} together with CoRY~\citep{ma2024cory} optimize homogeneous-role debate workflows.
More recent systems relax the workflow but not the supervision: MARFT~\citep{liao2025marft} confines the agents to a single sequential exchange, MAGRPO~\citep{magrpo} and MARTI~\citep{zhang2025marti} carry single-agent policy-gradient algorithms over to the multi-agent setting, and all of them optimize a return defined on a trajectory or a turn, which leaves the individual decisions inside a joint rollout without a target of their own.

\paragraph{On-Policy Distillation.}
On-policy distillation supervises the trajectories a student actually produces and therefore removes the mismatch between the states a fixed corpus covers and the states the student visits~\citep{opdsurvey,opcd,gad,hybridOPD,scope,tip,chen2026soda}.
\cite{gu2024minillm} casts the problem as reverse KL minimization under the student distribution, and \cite{opd} places on-policy and off-policy distillation in one family indexed by the divergence being minimized.
Two later analyses account for why the dense signal helps: \cite{yang2026learning} reads the objective as KL-regularized reinforcement learning whose per-token rewards are supplied implicitly by the teacher, while \cite{li2026rethinking} finds that the procedure aligns the student with its teacher locally on the states the student reaches and therefore depends on how compatible their reasoning patterns already are.
The teacher can also be dispensed with, and a body of work distills a policy from its own higher-quality samples~\citep{opsd,opsd2,sdpo,sdft,dqchen}, while a further line applies the same token-level supervision in place of a scalar reward in order to escape the credit-assignment and stability problems of reinforcement learning~\citep{srpo,sdrlvr,bousselham2025vold,wang2026openclaw,skillsd,zhong2026sod}.
Debate has also been placed on the supervising side, where \cite{wang2026mad} has several teachers argue a problem out so that their conclusion forms a stronger target for distillation, which makes the arrangement a multi-teacher one rather than a general multi-agent system whose own agents are being trained.
Each of these formulations supervises one student on trajectories it generated by itself, so neither the question of which role a piece of high-quality behavior should be attributed to nor the question of how a teacher can supervise a decision that depends on another agent arises in them.

\section{Algorithm}
\label{app:algorithm}

\Cref{sec:role_conditioned,sec:ras,sec:pac} each define part of the training signal, and this section states the order in which they are applied over a batch of rollouts.
The students first run the workflow of~\Cref{sec:mas} untouched, keeping the interaction on-policy; the verifier and the teacher are then called on the completed trajectory, the only phase in which privileged information exists; and the policies are updated from the advantage of~\Cref{eq:ras} before every privileged quantity is discarded.

\begin{algorithm}[H]
\caption{One training step of \ourmodel on a batch of joint rollouts}
\label{alg:masopd}
\SetAlgoLined
\DontPrintSemicolon
\KwIn{role policies $\{\pi_{\theta_i}\}_{i=1}^{N}$, frozen teacher $\pi_T$, verifier $V$, role conditions $\{r_i\}$, contrast map $\kappa$ on role conditions, weight $\lambda$}
\KwOut{updated role policies $\{\pi_{\theta_i}\}$}
\BlankLine

\textbf{Phase 1: joint rollout, students only.}\;
\For{$t=0,1,\ldots,T-1$}{
  \For{$i=1,\ldots,N$}{
    form $x_i^t=P_i(s^t,r_i)$ and sample $y_i^t\sim\pi_{\theta_i}(\cdot\mid x_i^t)$, retaining $\log\pi_{\theta_i}(y_{i,t,k}\mid x_i^t,y_{i,t,<k})$\;
  }
  obtain $e^t$ from the joint output $\mathbf y^t$, append the turn to the history\;
  \lIf{the roles agree}{terminate the episode}
}
\BlankLine

\textbf{Phase 2: attribution and teacher scoring, neither visible to a student.}\;
\For{$t=0,1,\ldots$ over the completed turns}{
  $a^t\leftarrow V(q,\mathbf y^t,e^t)$ and render $a^t$ into the privileged description $c^t$\;
}
\For{$t$ and $i$ over every response of $\tau$}{
  $\widetilde u\leftarrow u\oplus c^{<t}$ \tcp*{empty at $t=0$; never contains $c^{t'}$ for $t'\ge t$}
  $r_j\leftarrow\kappa(r_i)$\;
  $\ell_k^{(i)}\leftarrow$ force-decode $y_i^t$ under $\pi_T(\cdot\mid\widetilde u,r_i)$ \tcp*{target role}
  $\ell_k^{(j)}\leftarrow$ force-decode $y_i^t$ under $\pi_T(\cdot\mid\widetilde u,r_j)$ \tcp*{contrasting role}
  $A_{i,t,k}^{\mathrm{OPD}}\leftarrow\ell_k^{(i)}-\log\pi_{\theta_i}(y_{i,t,k}\mid x_i^t,y_{i,t,<k})$\;
  $A_{i,t,k}^{\mathrm{role}}\leftarrow\ell_k^{(i)}-\ell_k^{(j)}$\;
  $A_{i,t,k}^{\mathrm{RAS}}\leftarrow A_{i,t,k}^{\mathrm{OPD}}+\lambda A_{i,t,k}^{\mathrm{role}}$\;
}
\BlankLine

\textbf{Phase 3: update.}\;
form $\mathcal{L}_{\text{\ourmodel}}$ as in~\Cref{eq:mas_opd_loss} from $\operatorname{sg}(A_{i,t,k}^{\mathrm{RAS}})$, averaged without weights over tokens, then turns, then roles\;
update each $\theta_i$ from the gradient of its own responses alone, all policies synchronously\;
discard $\{c^t\}$ and every teacher score\;
\end{algorithm}

No student is ever conditioned on what the teacher receives, since Phase 1 completes before the verifier is called and $x_i^t$ is formed from the student-visible $s^t$ alone.
The attributions precede the scoring loop because scoring turn $t$ requires $c^{<t}$, so an attribution exists before the teacher reaches the turn it guides while never describing that turn itself.
Both teacher passes are inference only and the student log-probabilities are those retained from the rollout, so a step costs one sampling pass and two teacher forward passes per response.

\section{Experimental Setup}
\label{app:setup}

\subsection{Training Datasets}
\label{app:training_data}

Each domain is trained on a single corpus, the code domain on CodeContests and the mathematics domain on Polaris-Dataset-53K, both described below.

\paragraph{CodeContests~\citep{li2022competition}.}
CodeContests is a large-scale competitive-programming dataset released in conjunction with AlphaCode and designed to support training and evaluation of competition-level program synthesis.
The dataset aggregates programming problems from several major online judging platforms, including Aizu, AtCoder, CodeChef, Codeforces, and HackerEarth.
Its official release is organized into training, validation, and test splits, with the training split containing more than thirteen thousand programming problems.
Each instance provides a natural-language problem specification together with executable input--output tests.
The dataset distinguishes public tests, which are typically visible to contestants as examples in the problem statement, from private tests used for evaluation, and additionally includes automatically generated tests obtained by modifying existing inputs and validating the resulting outputs with known correct solutions.

Beyond problem statements and test cases, CodeContests provides both correct and incorrect human submissions in multiple programming languages, including C++, Python, and Java.
It also preserves rich problem-level metadata, such as the original source, execution time and memory limits, and, when available for Codeforces problems, contest identifiers, ratings, points, and algorithmic tags.
The problems cover a wide range of competitive-programming topics and difficulty levels, requiring models to understand lengthy specifications, derive appropriate algorithms, and produce complete executable programs.
Importantly, correctness can be determined directly by executing a generated program against the associated test cases, providing an objective functional-correctness signal without relying on a learned evaluator or language-model judge.

\paragraph{Polaris-Dataset-53K~\citep{polaris2025}.}
Polaris-Dataset-53K is an open-source mathematical reasoning corpus released as part of the POLARIS reinforcement-learning recipe.
The dataset contains approximately 53K mathematical reasoning problems curated from two larger open-source resources, DeepScaleR-Dataset-40K~\citep{deepscaler2025} and AReaL-boba-Data~\citep{fu2025areal}.
Its construction is motivated by the observation that the effectiveness of reinforcement-learning data depends strongly on problem difficulty relative to the current policy: examples that are solved almost universally provide little useful learning signal, whereas examples that are overwhelmingly unsolvable can lead to excessively sparse positive rewards.
POLARIS therefore explicitly characterizes its mathematical problems according to model-relative difficulty rather than treating all collected examples as equally informative.

To estimate problem difficulty during dataset construction, the POLARIS authors use DeepSeek-R1-Distill-Qwen-7B~\citep{guo2025deepseekr1} to generate eight candidate solutions for each problem and measure the corresponding pass rate.
Problems solved correctly in all eight rollouts are removed from the released corpus in order to reduce the proportion of trivially easy examples.
The resulting dataset contains approximately 26K problems originating from DeepScaleR and 27K from AReaL.
Each released instance contains a mathematical \texttt{problem}, its reference \texttt{answer}, and a \texttt{difficulty} annotation derived from the observed rollout pass rate.
Since perfectly solved examples are excluded during construction, the released difficulty annotations range from \texttt{0/8} to \texttt{7/8}.
This combination of answer-verifiable mathematical problems and explicit model-based difficulty information makes Polaris-Dataset-53K particularly suitable for reinforcement-learning-based post-training of mathematical reasoning models.

\subsection{Evaluation Benchmarks}
\label{app:benchmarks}

We evaluate mathematical reasoning and code generation on six established benchmarks.
We describe the origin, task characteristics, and evaluation structure of each benchmark below.

\subsubsection{Code Generation}
\label{app:benchmarks_code}

\paragraph{(1) LiveCodeBench-v6~\citep{jain2025livecodebench}.}
LiveCodeBench is a continuously updated benchmark designed to evaluate the coding capabilities of large language models while mitigating the data contamination issues associated with static code benchmarks.
It continuously collects newly released competitive-programming problems from LeetCode, AtCoder, and Codeforces, with temporal metadata that enables evaluations on problems released during different periods.
Beyond conventional code generation, the full benchmark also covers several complementary capabilities, including self-repair, code execution, and test-output prediction.
For code generation, each instance provides a natural-language problem specification together with input/output examples and executable test cases, and a generated program is evaluated according to whether it passes the associated tests.
LiveCodeBench maintains temporally separated releases and fine-grained version configurations, with v6 corresponding to one of these temporally defined problem sets.
This continuously updated construction makes LiveCodeBench particularly suitable for assessing coding ability on recent, previously unseen competitive-programming problems.

\paragraph{(2) APPS~\citep{hendrycks2021apps}.}
APPS (Automated Programming Progress Standard) is a large-scale code-generation benchmark designed to evaluate whether language models can synthesize complete programs from natural-language problem specifications.
It contains 10,000 English programming problems, divided evenly into 5,000 training and 5,000 test instances, and was manually curated from open-access programming platforms including Codewars, AtCoder, Kattis, and Codeforces.
The problems are organized into three difficulty levels named Introductory, Interview, and Competition, and they cover a broad spectrum from relatively elementary programming exercises to challenging algorithmic problems.
Each instance contains a natural-language problem description, executable input/output test cases, Python solutions when available, and metadata such as its difficulty and source.
The released dataset contains more than 130,000 test cases in total.
Evaluation is execution based: a generated program is judged by whether it produces the expected outputs on the corresponding tests, thereby assessing end-to-end program synthesis rather than surface-form similarity to a reference solution.

\paragraph{(3) CodeContests~\citep{li2022competition}.}
CodeContests is a competitive-programming dataset released in conjunction with AlphaCode and constructed to support training and evaluation of competition-level program synthesis.
It aggregates programming problems from multiple online competition platforms, including Aizu, AtCoder, CodeChef, Codeforces, and HackerEarth.
The released dataset contains 13,328 training problems, 117 validation problems, and 165 test problems.
Each problem is represented by a natural-language specification and can additionally contain public tests, private tests, automatically generated tests, correct and incorrect human submissions in multiple programming languages, and competition-specific metadata such as time and memory limits.
For Codeforces problems, metadata can further include ratings, points, and algorithmic tags.
The availability of executable tests and human submissions makes CodeContests substantially richer than benchmarks containing only a problem statement and a single reference implementation.
Candidate solutions are evaluated through program execution against input/output tests, emphasizing algorithmic reasoning and functional correctness under competitive-programming constraints.
The code domain trains on the training split of this dataset, as described in~\Cref{app:training_data}, and is evaluated on the held-out test split, so the problems reported here are disjoint from those seen during training.

\subsubsection{Mathematical Reasoning}
\label{app:benchmarks_math}

\paragraph{(4) AIME 2024.}
The American Invitational Mathematics Examination (AIME) is a challenging high-school mathematics competition administered by the Mathematical Association of America.
Unlike multiple-choice mathematics benchmarks, AIME uses a free-response format and requires each problem to be solved to a specific numerical answer.
The public AIME 2024 benchmark combines the 2024 AIME I and AIME II examinations, each containing 15 problems, for a total of 30 problems.
The questions cover major areas of competition mathematics, including algebra, geometry, number theory, and combinatorics, and generally require multi-step reasoning and nontrivial problem-solving insight rather than direct application of standard formulas.
Every problem has an integer answer between 0 and 999, which makes correctness unambiguous after the final answer has been extracted from a model response.
Public dataset releases provide the original problem statements and numerical answers, together with detailed reference solutions, making AIME 2024 a compact but demanding benchmark for evaluating advanced mathematical reasoning.

\paragraph{(5) AIME 2025.}
AIME 2025 is the subsequent annual edition of the American Invitational Mathematics Examination and follows the same free-response competition format as earlier AIME benchmarks.
The public benchmark contains 30 problems, consisting of 15 problems from AIME I and 15 from AIME II.
As in other AIME editions, every problem has a single integer answer in the range from 0 to 999, permitting deterministic verification of the final prediction.
The problems span the principal areas of olympiad-style high-school mathematics, including algebra, geometry, combinatorics, and number theory, and typically require several interconnected reasoning steps, careful case analysis, or problem-specific mathematical observations.
Public releases provide problem statements, gold answers, and reference solutions.
As a newly released annual competition set, AIME 2025 complements earlier AIME evaluations with a temporally distinct collection of challenging mathematical problems while retaining the standardized answer format that makes AIME convenient for automatic evaluation.

\paragraph{(6) OlympiadBench~\citep{he2024olympiadbench}.}
OlympiadBench is a challenging benchmark for advanced mathematical and scientific reasoning constructed from olympiad-level mathematics and physics problems.
The complete benchmark contains 8,476 problems collected from international and Chinese academic competitions, including the Chinese college entrance examination, with expert-level step-by-step solution annotations.
In contrast to AIME, OlympiadBench is deliberately heterogeneous: it covers both mathematics and physics, English and Chinese, text-only and multimodal problems, as well as open-ended questions and theorem-proving tasks.
Accordingly, the public release is organized into fine-grained subsets according to question type, modality, subject, language, and source category, including a text-only English open-ended competition-mathematics subset.
Individual instances provide structured information such as the problem, reference solution, final answer, answer type, mathematical subfield, units when applicable, and modality.
Its competition-level difficulty and diversity of problem and answer formats make OlympiadBench a broader test of advanced reasoning than mathematical benchmarks restricted to short integer-valued answers.

\subsection{Baselines}
\label{app:baselines}

Based on the construction of Stronger-MAS~\citep{strongermas}, we include a series of single-agent and multi-agent variants to disentangle the effects of additional interaction turns, reinforcement learning, and role-based multi-agent collaboration.
Unless otherwise specified, these baselines use the same backbone initialization, task instances, verifier resources, generation constraints, and evaluation protocol as our main experiments, while each method keeps the reward construction its own design prescribes.

\subsubsection{Single-Agent Baselines}
\label{app:baselines_sa}

\paragraph{(1) SA + ST.}
The \textbf{Single-Agent + Single-Turn (SA + ST)} baseline represents the standard inference-only setting in which a frozen language model solves each task end-to-end with a single response.
Following the task-specific single-agent construction in Stronger-MAS~\citep{strongermas}, the Code domain uses the Coder role to directly synthesize a solution program from the problem statement, while the Math domain uses the Reasoner role to directly derive the final answer.
No additional agent is instantiated, no cross-agent feedback is introduced during generation, and the model parameters remain unchanged.
This baseline therefore measures the task capability of the underlying model without either reinforcement-learning post-training or multi-agent collaboration, and serves as the basic reference for the remaining single-agent and multi-agent variants.

\paragraph{(2) SA + MT.}
The \textbf{Single-Agent + Multi-Turn (SA + MT)} baseline extends SA + ST by allowing the same agent to repeatedly reconsider and revise its own previous output over multiple turns~\citep{strongermas}.
Importantly, the additional turns do not introduce a second role: the Coder iteratively refines its own program in the Code domain, while the Reasoner iteratively revises its own solution in the Math domain.
Thus, unlike the multi-agent workflows described below, successive turns contain no feedback from a complementary agent and do not introduce the role-structured interaction between Coder and Tester or between Reasoner and Tool-User.
The agent revises until its output is self-consistent across turns or the same interaction horizon as the corresponding multi-turn setting is exhausted, so it is given a stopping rule of its own rather than being made to run out the turn budget.
This baseline controls for the effect of simply allocating additional generation and refinement turns, allowing us to distinguish gains from genuine cross-role collaboration from gains that could arise from additional single-agent computation alone.

\paragraph{(3) SA + ST + GRPO.}
The \textbf{Single-Agent + Single-Turn + GRPO (SA + ST + GRPO)} baseline retains the same single-turn inference structure as SA + ST but post-trains the underlying policy with Group Relative Policy Optimization (GRPO)~\citep{shao2024deepseekmath}.
For each training problem, multiple candidate responses are sampled from the current single-agent policy and evaluated using the corresponding task verifier.
GRPO then constructs a group-relative learning signal by centering and normalizing the rewards within the response group and optimizes the policy with its clipped policy-gradient objective.
Since all candidates in a standard single-agent group are generated from the same problem prompt, their rewards are directly comparable under the conventional GRPO formulation.
The Code setting trains the single-agent Coder, while the Math setting trains the single-agent Reasoner.
This baseline isolates the benefit of conventional reinforcement-learning post-training without introducing additional interaction turns or multi-agent coordination.

\paragraph{(4) SA + MT + GRPO.}
The \textbf{Single-Agent + Multi-Turn + GRPO (SA + MT + GRPO)} baseline is the reinforcement-learning counterpart of SA + MT.
It retains the same multi-turn self-refinement process, in which a single Coder or Reasoner repeatedly revises its own preceding output, while optimizing the underlying single-agent policy with GRPO~\citep{shao2024deepseekmath,strongermas}.
As in SA + MT, no complementary role participates in the trajectory: later turns are conditioned on the agent's own preceding generations rather than on feedback produced by a Tester or Tool-User.
The same task-level verifier and reward construction are used for policy optimization, and no multi-agent-specific credit-assignment mechanism is introduced.
Comparing this baseline with SA + ST + GRPO controls for whether a longer self-refinement horizon improves a GRPO-trained single-agent policy, while comparison with the multi-agent baselines separates multi-turn computation from role-based collaborative interaction.

\subsubsection{Multi-Agent System Baselines}
\label{app:baselines_mas}

\paragraph{(5) MAS.}
The \textbf{Multi-Agent System (MAS)} baseline evaluates role-based collaboration without reinforcement-learning updates.
Following Stronger-MAS~\citep{strongermas}, all roles are instantiated from the same frozen backbone and are differentiated through role-specific prompts.
In the Code domain, a Coder and a Tester interact iteratively: the Coder proposes or refines the candidate program, while the Tester constructs test cases and provides complementary feedback on the current solution, with the two roles continuing the interaction until the environment finds their outputs in agreement or the turn limit of~\Cref{sec:mas} is reached.
In the Math domain, a Reasoner derives the answer by mathematical reasoning while a Tool-User computes it by writing and executing a program, with the two roles likewise interacting until the environment finds their two answers in agreement or the same turn limit is reached.
Because the backbone parameters remain fixed throughout, any improvement over the single-agent baselines is attributable to the inference-time interaction structure and complementary role specialization rather than to parameter updates.
MAS therefore provides the direct prompt-only reference for assessing the additional benefit brought by reinforcement learning on top of the same collaborative workflow.

\paragraph{(6) MAGRPO.}
We adapt Multi-Agent Group Relative Policy Optimization~\citep{magrpo} as a multi-agent RL baseline.
MAGRPO formulates LLM collaboration as a cooperative MARL problem and optimizes decentralized agent policies using a shared team reward.
For each task, it samples a group of complete joint trajectories and computes, at each turn, the return-to-go of each trajectory; the centralized group-relative advantage is then obtained by centering these returns within the trajectory group and is shared by all participating agents to update their respective actions.
To evaluate MAGRPO under our setting, we retain exactly the same MAS workflows as our main experiments: the Coder and Tester iteratively interact for code generation, while the Reasoner and Tool-User collaborate over multiple turns for mathematical reasoning.
The role policies are initialized from the same backbone and optimized independently, as in the other trainable multi-agent methods.
We keep the backbone models, role prompts, environment interactions, termination conditions, training/evaluation data, and verifiers identical across methods.
The same environment-level team reward is used for MAGRPO, while role-specific local rewards are removed to preserve its original shared-reward formulation.
We further follow the original MAGRPO design by grouping complete multi-turn joint trajectories, computing Monte-Carlo return-to-go followed by mean-centered group-relative advantages, and applying the same centralized advantage to all agents within a trajectory.
Consistent with the original method, we do not introduce a learned critic or agent-specific credit assignment, and retain its unclipped policy-gradient objective without KL regularization.
These choices isolate the effect of MAGRPO's trajectory-level cooperative optimization while ensuring that differences in performance are not attributable to changes in the underlying MAS workflow or evaluation environment.

\paragraph{(7) MAS + GRPO.}
The \textbf{MAS + GRPO} baseline combines the same multi-turn MAS workflows with conventional GRPO training~\citep{shao2024deepseekmath,strongermas}.
We retain the Coder--Tester interaction for Code and the Reasoner--Tool-User interaction for Math, while directly applying the standard group-relative optimization procedure to experience collected from the multi-agent rollouts.
The two role policies are initialized from the same backbone but maintained and optimized independently, and each is updated from the responses its own role produced.
In contrast to Stronger-MAS, this baseline does not introduce agent- and turn-wise grouping or tree-structured sampling to construct comparison groups with identical role-specific interaction histories.
Consequently, as multi-turn trajectories evolve, different rollout branches may encounter different prompts because their preceding cross-agent interactions have diverged, while conventional GRPO still computes relative learning signals without explicitly accounting for these heterogeneous interaction states.
MAS + GRPO therefore tests whether standard single-agent group-relative optimization can be transferred directly to a multi-agent workflow without modifying its grouping and credit-assignment mechanism.
All underlying MAS components, including the agent roles, prompts, interaction protocol, environment, termination conditions, rewards, and evaluation procedure, are otherwise kept aligned with the MAS setting.

\paragraph{(8) CURE.}
CURE~\citep{cure} jointly trains a shared language model to act as both a coder and a unit-test generator, with the key idea that the tester should not only produce valid tests but also generate tests that effectively distinguish correct programs from incorrect ones.
In particular, the coder is optimized using execution-based correctness rewards, while the tester is rewarded for accepting correct candidate programs and rejecting incorrect ones, thereby enabling the two capabilities to co-evolve through reinforcement learning.
To adapt CURE to our multi-agent code-generation setting, we retain the Coder--Tester interaction protocol provided by our common experimental framework and apply the original CURE reward construction to the resulting code and test candidates.
Specifically, the final candidate programs are evaluated using the reference test suite to determine their correctness, while generated tests are scored according to their ability to preserve correct programs and discriminate against incorrect ones.
Following the original CURE design, the Coder and Tester share a single policy and are instantiated through role-specific prompts, and their rewards are normalized within the corresponding rollout groups before policy optimization.
For a fair comparison, CURE uses the same backbone model, task instances, Coder--Tester interaction protocol, rollout and inference budgets, and final evaluation procedure as the other methods, while retaining CURE-specific components such as its shared-policy formulation and discriminative tester reward.
Since CURE is specifically designed and evaluated for code generation and unit-test synthesis, we implement and report this baseline only on the Code domain.
Its shared policy is the one departure from the independent role policies used by the other trainable multi-agent baselines, since learning the two capabilities within a single model is integral to the method.

\paragraph{(9) MARFT.}
Multi-Agent Reinforcement Fine-Tuning (MARFT)~\citep{liao2025marft} is a reinforcement-learning framework for jointly optimizing collaborating LLM agents.
Its core idea is to formulate an LLM-based multi-agent system as a sequential joint decision process, where each agent acts conditioned on the current environment state and preceding inter-agent interactions.
To address credit assignment across interdependent agents, MARFT adopts centralized training with decentralized execution: a centralized critic estimates the value of joint interaction states, generalized advantage estimation (GAE) provides learning signals over multi-turn trajectories, and the agent policies are optimized using a PPO-style objective.

We implement MARFT on top of the Stronger-MAS codebase~\citep{strongermas} and adapt it to the same multi-agent collaboration settings used therein.
Specifically, for code tasks, we retain the multi-turn Coder--Tester interaction protocol, while for mathematical reasoning tasks, we retain the multi-turn Reasoner--Tool-User interaction protocol.
The original interaction workflow and environment are kept unchanged, and only the reinforcement-learning optimization procedure is replaced with MARFT.
Trajectories generated from these multi-turn interactions are used to train MARFT's centralized critic and subsequently update the agent policies with GAE-based PPO optimization.
This adaptation allows MARFT to be evaluated directly under the same multi-turn MAS setting, rather than under its original experimental configuration.

For a fair comparison, we keep the key non-algorithmic settings consistent with the Stronger-MAS experimental setup, including the base model and initialization, agent roles and prompts, training and evaluation data, reward design, interaction protocol and horizon, environment dynamics, generation constraints, policy-sharing configuration, decoding strategy, and overall training and rollout budgets.
We do not incorporate Stronger-MAS-specific agent-and-turn-wise grouping or branch-selection mechanisms into MARFT.
Therefore, MARFT retains its original centralized-critic-based credit assignment and PPO optimization while being evaluated under an otherwise aligned multi-agent environment and experimental protocol.

\subsection{Reward Design}
\label{app:reward}

\ourmodel itself requires no reward, since~\Cref{eq:mas_opd_loss} is a distillation objective and the verifier enters it only through the attribution of~\Cref{app:pac_attribution}, which labels an output rather than scoring it.
The reinforcement-learning baselines do require one, so we follow the reward design of the multi-agent training setting of Stronger-MAS~\citep{strongermas} on the two domains this paper runs, and describe it here for reproducibility.
Each domain supplies a \emph{team reward} shared by the whole system and a \emph{local reward} per role, both taking values in $[0,1]$.
Methods that optimize a single shared team signal, MAGRPO among them, read the team reward alone; the remaining reinforcement-learning baselines read both, combining them for role $i$ at turn $t$ as
\begin{equation}
r_{t,i}=\alpha\,r^{\mathrm{team}}_{t}+r^{\mathrm{loc}}_{t,i},
\qquad \alpha=1,
\label{eq:reward_combination}
\end{equation}
which is the mixed credit assignment of Stronger-MAS with the weight it reports.

A local reward is a masked convex combination of verifiable component scores.
For a role $i$ at turn $t$,
\begin{equation}
r^{\mathrm{loc}}_{t,i}=b_{t,i}\sum_{\ell}w^{i}_{\ell}\,g^{i}_{\ell,t},
\qquad \sum_{\ell}w^{i}_{\ell}=1,
\qquad g^{i}_{\ell,t}\in[0,1],
\label{eq:local_reward}
\end{equation}
where $w^{i}_{\ell}$ are fixed coefficients, $g^{i}_{\ell,t}$ are the component scores of that role and $b_{t,i}\in\{0,1\}$ is an availability mask that is zero whenever the evidence a component needs cannot be obtained at that turn.
Every component below is decided by execution or by comparison against a reference, so no score anywhere in this subsection comes from a model judging an output.

\subsubsection{Mathematical Reasoning}
\label{app:reward_math}

Answers are parsed and normalised with \textsc{Math-Verify}\footnote{\textsc{Math-Verify} (Hugging Face), \href{https://github.com/huggingface/Math-Verify}{github.com/huggingface/Math-Verify}. We use it as a parsing and normalisation front end and apply the comparator below to its output.} and then compared numerically with a tolerance $\varepsilon=10^{-6}$, two values counting as equal when
\begin{equation}
\textsc{NumEq}(a,b)=\mathbf{1}\!\left\{|a-b|\le\varepsilon\ \ \text{or}\ \ \frac{|a-b|}{\max(1,|b|)}\le\varepsilon\right\}.
\label{eq:numeq}
\end{equation}

\paragraph{Team reward.}
The team signal is sparse and is decided at termination by numerical equality against the reference answer $y^\star$, then broadcast unchanged to every turn of the episode:
\begin{equation}
r^{\mathrm{team}}_t=\mathbf{1}\{\textsc{NumEq}(\hat y,y^\star)\}\in\{0,1\},\qquad\forall t,
\label{eq:team_reward_math}
\end{equation}
with $\hat y$ the final answer the system submits.

\paragraph{Reasoner.}
The Reasoner is scored on output format and on the correctness of its answer, with coefficients $w^{\mathrm{Reasoner}}_{\mathrm{fmt}}=0.20$ and $w^{\mathrm{Reasoner}}_{\mathrm{ans}}=0.80$.
The format score $g^{\mathrm{Reasoner}}_{\mathrm{fmt},t}$ indicates whether the response matches the schema its template prescribes, and the answer score is
\begin{equation}
g^{\mathrm{Reasoner}}_{\mathrm{ans},t}=
\begin{cases}
\textsc{NumEq}(\hat y_t,y^\star), & \text{if \textsc{Math-Verify} extracts a numeric } \hat y_t,\\
0, & \text{otherwise},
\end{cases}
\label{eq:reasoner_ans}
\end{equation}
so a response from which no answer can be parsed scores zero rather than being left unscored.
The mask $b_{t,\mathrm{Reasoner}}$ is one whenever the reference answer is available at that turn.

\paragraph{Tool-User.}
The Tool-User is scored on whether its program runs, on whether an answer can be recovered from what the program printed, and on the correctness of that answer, with coefficients $w^{\mathrm{Tool\text{-}User}}_{\mathrm{run}}=0.10$, $w^{\mathrm{Tool\text{-}User}}_{\mathrm{parse}}=0.10$ and $w^{\mathrm{Tool\text{-}User}}_{\mathrm{ans}}=0.80$.
The run score indicates whether the program terminates within the sandbox limits of~\Cref{app:implementation_common} without an uncaught exception or a timeout, the parse score whether \textsc{Math-Verify} extracts a numeric value $\tilde y_t$ from the captured output, and the answer score is
\begin{equation}
g^{\mathrm{Tool\text{-}User}}_{\mathrm{ans},t}=
\begin{cases}
\textsc{NumEq}(\tilde y_t,y^\star), & \text{if a numeric } \tilde y_t \text{ is recovered},\\
0, & \text{otherwise}.
\end{cases}
\label{eq:tooluser_ans}
\end{equation}
The mask $b_{t,\mathrm{Tool\text{-}User}}$ is one whenever the execution result and the reference answer are both available at that turn.
The two roles are therefore scored on the same quantity, the answer each of them arrives at, through the evidence its own medium produces.

\subsubsection{Code Generation}
\label{app:reward_code}

Let $\mathcal{T}^{\mathrm{gold}}$ be the fixed set of golden unit tests of a problem and write $\textsc{Run}(u,\cdot)$ for the outcome of executing a program on the test $u$ in the sandbox of~\Cref{app:implementation_common}.

\paragraph{Team reward.}
The team signal is dense and is the fraction of golden tests the submitted program passes, again broadcast to every turn:
\begin{equation}
r^{\mathrm{team}}_t=\frac{1}{|\mathcal{T}^{\mathrm{gold}}|}\sum_{u\in\mathcal{T}^{\mathrm{gold}}}\mathbf{1}\{\textsc{Run}(u,\mathrm{code})=\textsf{pass}\}\in[0,1],\qquad\forall t.
\label{eq:team_reward_code}
\end{equation}

\paragraph{Coder.}
The Coder is scored on two sanity checks and on the fraction of golden tests its program passes, with coefficients $w^{\mathrm{Coder}}_{\mathrm{build}}=0.10$, $w^{\mathrm{Coder}}_{\mathrm{run}}=0.10$ and $w^{\mathrm{Coder}}_{\mathrm{pass}}=0.80$.
The build score indicates whether the candidate program compiles and imports without syntax errors, the run score whether a smoke subset of $\mathcal{T}^{\mathrm{gold}}$ executes without an uncaught exception or a timeout, and the pass score is
\begin{equation}
g^{\mathrm{Coder}}_{\mathrm{pass},t}=\frac{1}{|\mathcal{T}^{\mathrm{gold}}|}\sum_{u\in\mathcal{T}^{\mathrm{gold}}}\mathbf{1}\{\textsc{Run}(u,\mathrm{code}_t)=\textsf{pass}\}.
\label{eq:coder_pass}
\end{equation}
The mask $b_{t,\mathrm{Coder}}$ is one whenever the build and run logs and the golden-test results are available at that turn.
The weight therefore sits on functional correctness while the two sanity checks keep a program that fails to run from being indistinguishable from one that runs and is wrong.

\paragraph{Tester.}
The Tester is scored on whether the test case it authors is well formed and on whether that test case is consistent with the problem specification, with coefficients $w^{\mathrm{Tester}}_{\mathrm{valid}}=0.20$ and $w^{\mathrm{Tester}}_{\mathrm{spec}}=0.80$.
The validity score $g^{\mathrm{Tester}}_{\mathrm{valid},t}$ indicates whether the test is executable and deterministic and respects the input and output format the problem states.
The specification score runs the reference solution $\mathrm{code}^\star$ on the test inputs the Tester authored, collected in $\mathcal{U}_t$, and credits those whose declared expected output the reference reproduces,
\begin{equation}
g^{\mathrm{Tester}}_{\mathrm{spec},t}=\frac{1}{|\mathcal{U}_t|}\sum_{u\in\mathcal{U}_t}\mathbf{1}\{\textsc{Run}(u,\mathrm{code}^\star)=\textsf{pass}\},
\label{eq:tester_spec}
\end{equation}
so a test whose expected output the specification does not support scores zero however well formed it is.
The mask $b_{t,\mathrm{Tester}}$ is one whenever the test runner and the reference solution are both available at that turn.
Scoring the test case against the reference rather than against any candidate program is what keeps the Tester from being rewarded for the work of the Coder, and it is the same check that supplies \texttt{test\_correct} in~\Cref{app:pac_attribution}.

\subsubsection{Reward Design Against Attribution Design}
\label{app:reward_vs_attribution}

\ourmodel does require design of its own, namely the training-time verifier of~\Cref{sec:pac} and, on the rule-based route, the attribution procedure of~\Cref{app:pac_attribution}.
The question is what kind of judgement each approach asks a designer to supply, and the two subsubsections above make the comparison concrete.

A local reward asks for a number.
Building the ones used here meant choosing which components of a response to score, fixing a coefficient for each, and deciding how partial credit is expressed on a scale: ten coefficients over four roles, with the format, build, run and parse checks held at $0.10$ or $0.20$ and functional correctness at $0.80$ in all four cases.
None of those choices is verifiable.
There is no experiment that establishes $0.10$ rather than $0.15$ for a build check, and a different split would train a different system, so the weights are a judgement about how much a designer believes each signal matters.
They are also specific to the domain and to the roles: the components of~\Cref{eq:coder_pass} refer to golden tests, those of~\Cref{eq:reasoner_ans} to a parsed numeric answer, and neither set transfers to the other domain, let alone to a workflow with different responsibilities.
A team reward adds a second such choice, sparse at termination on mathematics and dense over test outcomes on code, and methods that read both then inherit the balance between them fixed by~\Cref{eq:reward_combination}.

Attribution asks for a fact.
Given the outputs of one interaction, the procedure has to determine which of them is wrong, and unlike a weight that judgement has a correct answer, which is why it can be obtained by executing tests or comparing against a reference rather than being tuned.
It is also a single question rather than one question per component: the verifier that already decides whether a program passes its tests, or whether an answer matches a reference, is by itself enough to answer it, which is how the rules of~\Cref{app:pac_attribution} are built from checks the reward design needed anyway.
Nothing has to be settled about granularity, since the attribution is attached to the output it concerns, and nothing about scale, since the teacher consumes it as context rather than as a number to be traded off.
The design effort accordingly does not grow with the number of signals a designer might want to express, which is what~\Cref{app:scaling_pac} exploits when it replaces the rules with a single model call in the wider systems.

The distinction matters for what the supervision can then do.
A local reward has to compress the outcome of an interaction into one scalar per role and per turn before the learner sees it, whereas an attribution names the responsible output and enters the teacher's context through~\Cref{eq:priv_state}, leaving the teacher to express the consequence over the tokens of that output, which is the token-level signal~\Cref{eq:mas_opd_loss} distils and the reason the students can be taught which side should yield in a conflict rather than merely that the episode went badly.
That supervision is what~\Cref{sec:rq3} finds the students internalise, in that they continue to settle conflicts in the direction their own correctness warrants once the verifier is removed.
Since a correctness judgement is a quantity any verifiable domain already produces, the supervision PAC requires can be obtained wherever such a judgement exists, without the calibration a new reward design would need.

\subsection{Models and Evaluation Protocol}
\label{app:models_eval}

\paragraph{Models.}
All experiments use models from the Qwen3 series~\citep{qwen2025qwen3}.
The teacher is Qwen3-14B and the students are Qwen3-1.7B and Qwen3-4B, and every agent of a multi-agent system is instantiated from the student model of the configuration being reported.
Unless an experiment states otherwise, it reports the 1.7B student, which is the size used in~\Cref{sec:rq2,sec:rq3,sec:rq_scaling}.

\paragraph{Use of the teacher.}
The teacher is frozen throughout training and is never asked to generate a trajectory of its own.
It is force-decoded on responses the students have already produced, in the sense of~\Cref{sec:opd}, so it contributes token-level scores and no tokens of its own to the interaction.
It takes no part in evaluation, where only the trained student policies are run.

\paragraph{Inference and interaction.}
Every model is run in the no-thinking mode of Qwen3 and decodes at a temperature of $0.6$, with the remaining decoding parameters and generation limits given in~\Cref{app:implementation_common}.
Multi-agent interactions are capped at a horizon of $T=4$ turns in the sense of~\Cref{sec:mas}, and an interaction that reaches agreement earlier terminates at that point.

\paragraph{Metrics.}
We report Pass@1 on the code benchmarks and accuracy on the mathematics benchmarks, with final outputs judged by the task-specific verifier of each domain.
Those verifiers are described in~\Cref{app:reward} and the per-benchmark protocols in~\Cref{app:benchmarks}.

\subsection{Implementation Details}
\label{app:implementation}

\paragraph{Computing infrastructure.}
All training and evaluation runs are carried out on a single node with eight NVIDIA H20 GPUs of 96GB memory each.

\subsubsection{Settings Shared by All Methods}
\label{app:implementation_common}

\paragraph{Framework and alignment across methods.}
We implement all baselines within the same experimental framework and align their non-algorithmic settings whenever compatible with the original methods.
In particular, the methods use the same backbone initialization, training and evaluation data, task environments, role prompts, interaction horizon, generation limits, and number of optimization steps.
The same task-specific verifier resources are used whenever the corresponding algorithm does not prescribe a distinct reward construction.
The reward signals the reinforcement-learning baselines are trained with are set out in~\Cref{app:reward}, which \ourmodel does not draw on, having no reward term.
For trainable multi-agent methods, the role policies are initialized from the same backbone but optimized independently, matching the role-specialized policy organization of~\ourmodel and the system of~\Cref{sec:mas}.
CURE is the only exception: we retain its original shared-policy formulation because jointly learning code generation and unit-test generation within a single policy is integral to its design.
Method-specific reward constructions, credit-assignment mechanisms, and optimization objectives are otherwise preserved.

\paragraph{Common training configuration.}
Following the experimental protocol of Stronger-MAS~\citep{strongermas}, we use Qwen3~\citep{qwen2025qwen3} models in the no-thinking mode.
For both Code and Math, the maximum prompt length is 8,192 tokens and the maximum response length is 4,096 tokens.
All trainable baselines are run for 150 optimization steps with a global batch size of 128 and, where applicable, an optimization mini-batch size of 64.
Unless a method requires an algorithm-specific setting, we use Adam with a policy learning rate of $1\times10^{-6}$, weight decay of $0.01$, and gradient clipping at $1.0$.
During training, responses are sampled with temperature $1.0$, $\mathrm{top}\text{-}p=1.0$, and $\mathrm{top}\text{-}k=-1$.
For methods based on grouped rollouts, the rollout group size is set to four.
Multi-turn methods use a maximum interaction horizon of $T=4$.
Evaluation decodes with temperature $0.6$, $\mathrm{top}\text{-}p=0.95$, and $\mathrm{top}\text{-}k=20$.
Every method that is trained is trained five times from independent random seeds and each resulting system is evaluated once, so the standard deviations reported in~\Cref{tab:main_results} are taken over independent training runs and include both the variability of optimization and that of decoding.
The three prompt-only baselines train nothing, and their five runs are independent evaluations of the same model under this decoding.
Algorithm-specific hyperparameters that differ from these common settings are stated below.

\paragraph{Sandboxed code execution.}
Both domains are grounded in \emph{verifiable execution}: whenever an agent emits a program, the environment runs it, and the outcomes the protocol surfaces are fed back into the interaction, so that subsequent decisions rest on an observed execution result rather than on a model's self-assessment.
Each call is stateless and mutually isolated.
The program is materialised in a fresh working directory and launched as a separate operating-system process in its own process group; the test input is supplied on \texttt{stdin} and \texttt{stdout} is captured as the outcome.
Independent test cases are dispatched concurrently, as they share no state.
Every call is bounded by a wall-clock timeout of $30$\,s for code and $20$\,s for math, after which the entire process group is terminated.
Crucially, execution never propagates a failure into the rollout loop: a crash, a syntax error, or a timeout is converted into an error string that is recorded as the outcome and, where the protocol calls for it, surfaced verbatim to the agents on the following turn.
This keeps trajectories robust to arbitrary model-generated code while retaining the diagnostic content of the failure, which is often what enables the next revision.

\paragraph{Which roles invoke execution.}
The two domains differ in why they execute, and correspondingly in which roles are involved, as~\Cref{tab:execution_roles} sets out.
In the code domain, execution is the means of checking a candidate program, since nothing can be judged about it without running it.
The program executed is always the Coder's, as the Tester authors test cases and never a program of its own, and that one program is run twice per turn over two different test inputs.
It is run on the golden unit tests that ship with the dataset, which is what decides whether the task has been solved, and on the test case the Tester authored, whose outcome is the disagreement the environment reports and the condition on which the interaction terminates.
The two executions differ in who reads them: only the outcome on the Tester-authored case reaches the agents, while the result on the golden tests is read by the verifier alone and enters no agent's context at any turn, so no agent is told whether the task has been solved, and termination follows from the two roles agreeing or the turn budget running out rather than from that result.
In the mathematics domain, by contrast, execution is a means of solving rather than of checking.
Only the Tool-User writes code, a short program that computes and prints a candidate answer, which is then parsed from the captured output.
The Reasoner never executes anything and states its answer in natural language after the marker~\texttt{\#\#\#\#} that its template prescribes, and the two candidate answers are compared by a symbolic verifier independently of execution.
Consequently a mathematics episode still yields an answer from the Reasoner even when the program of the Tool-User fails to run, whereas in the code domain a program that cannot be executed leaves nothing to inspect.

\begin{table}[t]
\centering
\caption{\textbf{What is executed on behalf of each role.} A row is one execution the environment performs, not a program the role wrote: the only program run on the code domain is the Coder's, and the Tester contributes the test input it is run on, so the two code rows are the same program on two different inputs. The Reasoner writes no program at all and states its answer in natural language, which is compared against the Tool-User's printed answer by a symbolic verifier.}
\label{tab:execution_roles}
\vspace{1mm}
\begingroup
\setlength{\tabcolsep}{6.0pt}
\renewcommand{\arraystretch}{1.15}
\small
\begin{tabular}{llll}
\toprule
\textbf{Domain} & \textbf{Role} & \textbf{Program run} & \textbf{Test input} \\
\midrule
\multirow{2}{*}{Code}
 & Coder     & the Coder's     & the dataset's golden unit tests \\
 & Tester    & the Coder's     & the Tester's own test case \\
\midrule
\multirow{2}{*}{Math}
 & Tool-User & the Tool-User's & --- \\
 & Reasoner  & ---             & --- \\
\bottomrule
\end{tabular}
\endgroup
\end{table}

\subsubsection{Baselines}
\label{app:implementation_baselines}

\paragraph{Single-agent baselines (1)--(4).}
SA + ST and SA + MT use the Coder for Code and the Reasoner for Math, with model parameters kept frozen.
SA + ST generates a single response, whereas SA + MT revises its own output until that output is self-consistent across turns or the horizon $T=4$ is reached, which mirrors the agreement-or-horizon rule of the multi-agent workflows.
Their trainable counterparts, SA + ST + GRPO and SA + MT + GRPO, preserve the corresponding interaction structures and optimize the policy using standard GRPO~\citep{shao2024deepseekmath}.
We sample four rollouts per problem and otherwise use the common GRPO training configuration above.

\paragraph{(5) Prompt-only MAS.}
The MAS baseline uses the same Coder--Tester and Reasoner--Tool-User interaction workflows as the trainable multi-agent methods, with the backbone parameters kept frozen.
The roles are differentiated through their corresponding role prompts and interact until their outputs agree or the maximum horizon $T=4$ is reached.
Since MAS performs no parameter updates, the training-specific optimization settings above do not apply.

\paragraph{(6) MAGRPO.}
We adapt MAGRPO~\citep{magrpo} to the same multi-turn interaction workflows with a generation group size of $G=4$ and a maximum horizon of $T=4$.
The role policies are initialized from the same backbone and optimized independently.
Following the original MAGRPO formulation, training uses the shared environment-level team reward without additional role-specific local rewards, and retains its trajectory-level Monte-Carlo group-relative optimization.
We use the original policy-gradient formulation without a learned critic, PPO-style clipping, or KL regularization.
The policy learning rate, batch sizes, sampling configuration, and total number of optimization steps otherwise follow the common settings above.

\paragraph{(7) MAS + GRPO.}
MAS + GRPO applies conventional GRPO~\citep{shao2024deepseekmath} directly to the common multi-turn Coder--Tester and Reasoner--Tool-User workflows.
The two role policies are initialized from the same backbone but maintained and optimized independently.
For each task, we generate four multi-agent rollouts and update each role policy from the responses produced by that role using the common GRPO configuration above.
We do not introduce the agent- and turn-wise regrouping or tree-structured branch selection of Stronger-MAS~\citep{strongermas}; the conventional group-relative optimization procedure is instead applied directly to the collected multi-agent rollouts.

\paragraph{(8) CURE.}
We implement CURE~\citep{cure} on the Code domain only and retain its original shared-policy formulation, in which a single language model serves as both the Coder and the Tester under role-specific prompts.
The code-solution and unit-test rollout group sizes are both set to four, and we retain CURE's original cross-execution procedure and separate normalization of the coder and tester reward groups.
We use its clipped policy objective with reference-policy KL regularization coefficient $\beta=0.01$ and a policy learning rate of $1\times10^{-6}$.
The remaining compatible training and generation settings follow the common configuration above.
The response-length reward transformation introduced for CURE's long-CoT variant is not used because all models in our experiments operate in the no-thinking mode.

\paragraph{(9) MARFT.}
We adapt the action-level training procedure of MARFT~\citep{liao2025marft} to the same multi-agent framework used by the other methods, while retaining our Coder--Tester and Reasoner--Tool-User interaction environments.
The role-specific actor policies are initialized from the same backbone and optimized independently, while MARFT's centralized critic is shared across roles and estimates values from the joint interaction state.
Following the action-level MARFT implementation, the critic uses a frozen language-model encoder followed by a trainable MLP value head with hidden size $64$.
We use an actor learning rate of $1\times10^{-6}$, a critic learning rate of $1\times10^{-5}$, a discount factor of $0.99$, a GAE coefficient of $0.95$, a PPO clipping coefficient of $0.2$, a critic loss coefficient of $1.0$, and one PPO epoch per rollout batch.
The critic is optimized with Huber loss, response-level action log-probabilities are normalized by the number of generated tokens following MARFT's action-normalization procedure, and the maximum gradient norm is set to $0.5$.
The remaining generation, interaction-horizon, and training-step settings follow the common configuration above.
No Stronger-MAS-specific tree-structured sampling or agent- and turn-wise grouping is introduced into MARFT.

\subsubsection{\ourmodel}
\label{app:implementation_ours}

\ourmodel is trained in the same multi-agent workflows, on the same corpora and under the same common configuration as the trainable multi-agent baselines, with the two role policies initialized from the same backbone and optimized independently.
What distinguishes it is a third model that takes part in training alone.

\emph{Teacher.}
The teacher is Qwen3-14B, is used in the same no-thinking mode as the students and is never updated, so it carries no optimizer state.
It produces no trajectory of its own: it force-decodes a response the student has already sampled and returns the per-token log-probabilities of~\Cref{eq:teacher_score}, so the sampling parameters of the common configuration do not apply to it.
Its context is the privileged context $\widetilde u$ of~\Cref{eq:priv_state}, which carries the attribution of every completed turn in addition to what the student reads, and its prompt limit is therefore raised to 12,288 tokens from the 8,192 allowed for the students.
The teacher and the verifier are both required during training alone and are discarded once it ends, which leaves the deployed system at the size of its students.

\emph{Role-advantage specialization.}
Each student response is scored twice by the teacher, once under the role that produced it and once under the contrasting role, and the two passes differ in the role condition alone, with every other part of the context held identical.
The contrasting role is the one the role condition names as the responsibility not to take over, which pairs the Coder with the Tester and the Reasoner with the Tool-User, as~\Cref{app:scaling_ras} sets out.
The role advantage of~\Cref{eq:role_adv} enters the update with the weight $\lambda$ of~\Cref{eq:ras} set to $0.1$, and is treated as a token-level advantage rather than as a differentiable loss, so no gradient flows through either teacher pass.
A second force decoding per response is the entire additional cost of the module.

\emph{Privileged attribution for coordination.}
Attributions are produced by the verification rules of~\Cref{app:pac_attribution} rather than by an attribution model, which is the route every experiment outside~\Cref{sec:rq_scaling} uses.
On the code domain the two checks are whether the student program passes the golden unit tests and whether the golden reference solution reproduces the expected output the student declared on the student test input; on the mathematics domain they are whether the derived answer and the printed result each match the reference answer, decided by numerical tolerance and by symbolic simplification in turn.
The outcome is rendered into the two-line template of~\Cref{box:pac_template}, which states the candidate values the agents produced and the side to trust and never contains a reference value.
An attribution is attached to the record of the turn it judges and is visible to the teacher alone, so the set is empty at the first turn and the students never read one at any turn.

\emph{Optimization.}
Training follows~\Cref{eq:mas_opd_loss}, whose average is unweighted over tokens within a response, over turns within a role and over roles within the system, so that a longer response or trajectory does not dominate the update.
The gradient of a role updates that role alone, and the two policies are updated synchronously once a batch of joint rollouts is complete.
The optimizer, learning rate, weight decay, gradient clipping, batch sizes, number of optimization steps, rollout sampling parameters and interaction horizon are the common ones of~\Cref{app:implementation_common}.
Being an on-policy distillation objective rather than a group-relative or PPO-style one, it involves no rollout group, no learned critic, no clipping and no KL term, and the single hyperparameter it introduces is $\lambda$.

\subsubsection{Training Cost}
\label{app:cost}

\Cref{tab:cost} reports the wall-clock cost of training \ourmodel on the single node of eight NVIDIA H20 GPUs with 96GB of memory each specified in~\Cref{app:implementation}, separating the two phases that make up one optimization step: the joint rollout, in which the agents interact to produce an on-policy trajectory, and the update, in which the two policies are optimized.
Each entry is the mean over the optimization steps of a run, in seconds, and the number of steps is the one given in~\Cref{app:implementation_common}.
The cost of adding further agents is a separate question and is treated in~\Cref{app:scaling_cost}.

\begin{table}[t]
    \centering
    \caption{
    \textbf{Per-step training cost of \ourmodel.}
    Mean wall-clock seconds per optimization step for each student and domain, measured on a single node of eight NVIDIA H20 GPUs with 96GB of memory each, with the rollout and the update reported separately.
    All remaining settings are those of~\Cref{app:implementation_common}.
    }
    \vspace{1mm}
    \label{tab:cost}

    \begingroup
    \setlength{\tabcolsep}{6.0pt}
    \renewcommand{\arraystretch}{1.08}
    \normalsize

    \begin{adjustbox}{max width=\columnwidth}
    \begin{tabular}{lcccc}
        \toprule

        & \multicolumn{2}{c}{\textbf{Code}}
        & \multicolumn{2}{c}{\textbf{Math}} \\
        \cmidrule(lr){2-3}
        \cmidrule(lr){4-5}

        {\footnotesize\textbf{Student}}
        & {\footnotesize\textbf{Rollout (s)}}
        & {\footnotesize\textbf{Update (s)}}
        & {\footnotesize\textbf{Rollout (s)}}
        & {\footnotesize\textbf{Update (s)}} \\

        \midrule

        Qwen3-1.7B
        & 512.12
        & 34.44
        & 258.81
        & 55.68 \\

        Qwen3-4B
        & 485.89
        & 50.83
        & 478.90
        & 80.07 \\

        \bottomrule
    \end{tabular}
    \end{adjustbox}

    \endgroup

    \vspace{-6pt}
\end{table}

\section{Additional Experiments}\label{app:additional}

\subsection{Scaling to More Agents}\label{app:scaling}

A multi-agent system is organized around a fixed workflow, so adding agents means adding them inside that workflow rather than assembling an arbitrary collection of them.
The systems of~\Cref{sec:rq_scaling} grow in exactly this way, by replicating agents of a role that is already present, and the code domain accordingly holds the two role types of Coder and Tester however many agents it has, with the multi-turn interaction between them unchanged.
Comparing the outputs and reporting the disagreement is the task of the environment in~\Cref{sec:mas} rather than of any agent, so adding agents calls for no additional role.
Let $\mathcal R$ denote the set of role types and $\rho(m)\in\mathcal R$ the role of agent $m$.
Widening the system enlarges the set of agents while leaving $\mathcal R$ exactly as it is, and both modules are defined over $\mathcal R$ rather than over individual agents.
\ourmodel therefore carries over to a larger system without a new design decision, a new hyperparameter or a new prompt, and at a cost that stays proportional to the number of responses the system produces rather than growing with the number of agents on top of that.
The two parts below establish this for each module in turn and~\Cref{app:scaling_cost} collects the cost.

\subsubsection{Accuracy under More Agents}\label{sec:rq_scaling}

\begin{figure}[t]
    \centering
    \includegraphics[width=\textwidth]{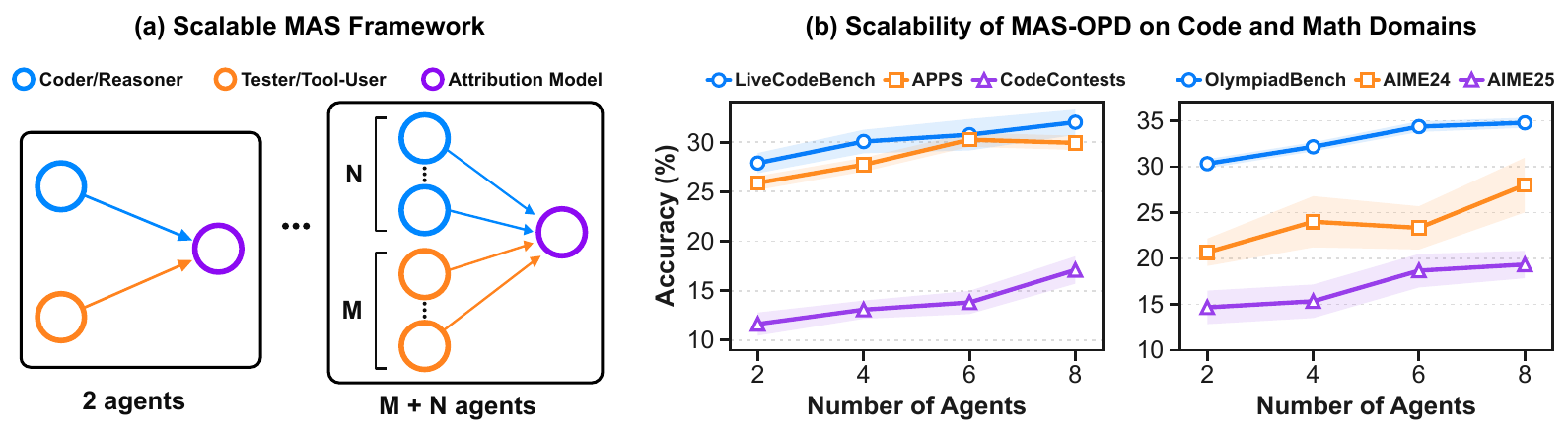}
    \caption{
    \textbf{Scaling \ourmodel to more agents.}
    The figure has two parts:
    \emph{(a) scalable MAS framework}, showing how the system grows, a role being replicated so that a system of $2n$ agents holds $n$ instances of each of the two roles;
    and \emph{(b) accuracy w.r.t.\ agent count}, reporting the 1.7B student on the six benchmarks as the number of agents grows from two to eight, with shaded bands giving the standard deviation over five training runs.
    The attribution model runs during training alone and is not counted.
    }
    \vspace{1mm}
    \label{fig:scaling}
\end{figure}

We add agents to the workflow of~\Cref{sec:mas} by replicating agents of the same role, leaving the roles themselves and the multi-turn interaction between them unchanged, and train the 1.7B student under systems of two, four, six and eight agents, as~\Cref{fig:scaling} shows.
All four systems obtain their attribution from the attribution model rather than the rules used in the main results, which keeps the attribution route fixed across the comparison, and~\Cref{app:scaling_setup} gives the full setup.

\begin{itemize}[leftmargin=*]

\item \textbf{Obs 10: adding agents improves both domains and has not saturated at eight agents.}
The average over the three benchmarks of a domain rises from roughly 21.8 points at two agents to 26.3 on code and 27.4 on mathematics, and all six benchmarks end above where they start.
The increment from each further pair of agents moreover stays of the same order rather than tailing off, so the curves give no indication that eight agents is where the benefit stops.

\item \textbf{Obs 11: the trend is carried by the domain average rather than by every individual curve.}
Two of the six benchmarks give back a fraction of a point at one point in the range, in both cases by less than the standard deviation there, while the domain average increases at every step on both domains.
Adding agents therefore acts on general competence rather than on any one benchmark, and the isolated dips are consistent with run-to-run variation instead of an approaching ceiling.

\item \textbf{Obs 12: the benchmarks the system finds hardest are the ones that gain most.}
CodeContests, AIME24 and AIME25 each improve by more than thirty percent over their two-agent accuracy, with CodeContests gaining close to a half, whereas the three it already scores highest on gain around fifteen percent.
Extra agents of an existing role thus broaden the search on the problems it was previously failing, rather than consolidating the ones it could already solve.

\end{itemize}

\subsubsection{Role-Advantage Specialization}\label{app:scaling_ras}

RAS is defined between role types.
Every role condition of~\Cref{app:role_conditions} names one role whose responsibility it must not take over, which gives a map $\kappa$ on role conditions with $\kappa(r_{\text{Coder}})=r_{\text{Tester}}$ and $\kappa(r_{\text{Reasoner}})=r_{\text{Tool-User}}$, and the role advantage of~\Cref{eq:role_adv} for a response of agent $m$ reads
\begin{equation}
A_{m,k}^{\mathrm{role}}=\ell_k\!\left(u_m,\,r_{\rho(m)}\right)-\ell_k\!\left(u_m,\,\kappa(r_{\rho(m)})\right),
\label{eq:scaling_ras}
\end{equation}
where $u_m$ is the context of the response being scored and $r$ is a role condition.
The agent index enters the right-hand side only through $u_m$ and through the role $\rho(m)$, and the contrasting term is the role condition of another role rather than the output of another agent.

\paragraph{Which teammate is contrasted against is therefore not a choice.}
Both evaluations in~\Cref{eq:scaling_ras} are taken on the same context $u_m$, which is what lets the student term cancel, and a role condition names a role rather than an agent, as the displays of~\Cref{app:role_conditions} show.
Three Coder agents thus contribute the same role condition and the same contrasting score, and scoring a response of any of them draws that contrast from the Tester role alone.
The module never has to pick an instance and the paper never has to say which one it picked.

\paragraph{The overhead is a constant multiple.}
A teacher score depends on the trajectory and the response as well as on the role condition, so every response is scored on its own and the number of forward passes grows with the number of responses, which is what scoring each response at all demands of any token-level distillation.
What RAS adds is one contrasting pass per response, a multiple set by $\kappa$ rather than by the number of agents, so the overhead stays at roughly twice that of the role-conditioned form of~\Cref{sec:role_conditioned} however wide the system becomes.

RAS therefore scales with the number of agents in the sense that matters for a method: it is defined between role types and so acquires nothing new as agents are added, it never needs to be told which teammate to compare against, and the cost it adds over ordinary distillation is a constant factor rather than one that grows with the size of the system.

\subsubsection{Privileged Attribution for Coordination}\label{app:scaling_pac}

PAC keeps the paradigm it has with two agents.
An attribution is obtained from the execution outcome, filled into a template and handed to the teacher as context, and adding agents changes neither of the last two steps.
The attribution itself is a verdict per agent rather than a joint label, since each check reads the output of one agent against a training-time reference and never the output of another, so with $N$ agents
\begin{equation}
a^t=\left(a_1^t,\ldots,a_N^t\right),\qquad c^t=\Pi\!\left(a^t\right),
\label{eq:scaling_pac}
\end{equation}
where $\Pi$ is the template of~\Cref{app:pac_attribution}.
The four named outcomes given there are the readable names of the values $a^t$ takes when $N=2$.
A larger system lengthens $c^t$ by one line per agent, and because a $c^t$ states facts and addresses no role, one attribution still serves the contexts of every agent at a turn, so PAC adds no teacher forward pass however many agents there are.
No reference value is disclosed either, since a verdict labels a value an agent produced rather than introducing one of its own.

\paragraph{What adding agents does change is how the attribution is obtained.}
With two agents the rules that produce $a^t$ are simple to write and quick to run.
With more agents they remain quick, and the per-agent checks remain independent, but the procedure takes more design work: the wording that reports the conflict has to accommodate several values rather than a pair, two agents of one role may disagree with each other, which the two-role template has no phrasing for, and both have to be settled again for every new workflow.
What has to be established is unchanged throughout, namely which agent is at fault, and once that is settled the template produces the privileged information as before.

\paragraph{A second route reaches the same attribution with a model call.}
An attribution model is a model called during training to locate which outputs disagree with the reference, and it returns an $a^t$ of the same form the rules would produce, after which~\Cref{eq:scaling_pac} proceeds unchanged.
The two routes therefore differ only in how $a^t$ is obtained and trade hand-written attribution rules against additional model calls during training, and~\Cref{sec:rq_scaling} takes the second route at every size it reports.
The model locates and does not decide correctness, which continues to come from the reference, so the attribution remains a verified statement rather than the opinion of another model and the model falls under the mechanism that $V$ already admits in~\Cref{eq:attribution}.
It runs during training only and takes no part in the workflow, its output enters the teacher context alone and no student receives it at any turn, and it is removed together with the verifier once training ends.
One call per turn suffices because a single attribution serves every agent, so this route does not grow more expensive as agents are added and it leaves inference untouched.

A larger system also gives the privileged information more to do.
Two agents leave a single pair of outputs to reconcile, whereas $n$ agents per role leave every one of them facing the outputs of $n$ counterparts, so the question of whether to revise or to hold arises against more conflicting evidence rather than less.
The attribution says which of those outputs agree with the reference, the teacher writes targets that express the resulting decision, and the students face the same conflicts with the verdicts withheld, so the only way open to them for lowering the distillation loss is learning to make the decision from the outputs alone.

PAC therefore scales in the same sense as RAS.
Its paradigm is unchanged as agents are added, since everything a larger system asks of it is confined to obtaining $a^t$ and the template that turns an attribution into privileged information stays as it is.
Its cost stays controlled as well, because a single attribution serves every agent at a turn, so PAC adds no teacher forward pass however many agents there are and the second route spends one model call per turn irrespective of that number.
That second route is also what keeps the design itself simple once the system grows: rather than extending the hand-written rules to every new configuration, one call locates which outputs disagree with the reference and the rest of the procedure proceeds unchanged, which is why the scaling experiment of~\Cref{sec:rq_scaling} adopts it throughout.
PAC does therefore ask for design of its own, namely a training-time verifier and, on the first route, the rules that turn its verdicts into an attribution, and what it asks is bounded in a way a local reward is not: both routes only have to establish which output is at fault, a question every verifier already answers and a model call answers directly, whereas a local reward has to convert that same judgement into a scalar per role and per turn and then be retuned whenever the task format, the role responsibilities or the protocol changes.
\Cref{app:reward_vs_attribution} develops the comparison against the reward design the baselines use.

\subsubsection{The Cost of Adding Agents}\label{app:scaling_cost}

The two modules place their cost in different places, and collecting them gives the total for a turn in which $N$ agents each produce one response.
Scoring those responses at all takes $N$ teacher passes, which is what any token-level distillation spends and which the role-conditioned form of~\Cref{sec:role_conditioned} already spends.
RAS adds one contrasting pass per response and PAC adds none, so
\begin{equation}
C_{\text{\ourmodel}}(N)=\underbrace{N}_{\text{OPD}}+\underbrace{N}_{\text{RAS}}+\underbrace{0}_{\text{PAC}}=2N,
\qquad
\frac{C_{\text{\ourmodel}}(N)}{C_{\mathrm{OPD}}(N)}=2 ,
\label{eq:scaling_cost}
\end{equation}
counted in teacher forward passes.
The ratio is what matters here.
The absolute count rises with $N$ because there are more responses to score, which is a property of the system rather than of the method, while the factor the method contributes on top is independent of $N$.
Alongside these passes the second attribution route spends one model call per turn, independent of $N$, since a single attribution serves every agent.

Nothing else grows.
The configuration holds one role condition and one verification check per role type, so it is unchanged as long as $\mathcal R$ is, and~\Cref{eq:scaling_ras,eq:scaling_pac} introduce no quantity that has to be chosen per agent.
The framework keeps the single hyperparameter $\lambda$ however many agents there are, and the verifier and the attribution model are both removed once training ends, so neither module leaves any component behind at deployment and a system of $N$ agents is deployed as those $N$ agents alone, its inference cost growing with the number of agents and interaction turns as that of any multi-agent system does.

\subsubsection{Setup of the Scaling Experiment}\label{app:scaling_setup}

\paragraph{How the attribution is obtained.}
The main results of~\Cref{sec:rq1}, and every other experiment we report, attribute a conflict with the rules of~\Cref{app:pac_attribution}, which suffice while each role is held by a single agent.
The scaling experiment of~\Cref{sec:rq_scaling} instead uses the attribution model at all four sizes, the two-agent one included, so that the attribution route is held fixed across the systems being compared and the differences between them are due to the number of agents alone.
Its two-agent column is consequently a separate run from the corresponding row of~\Cref{tab:main_results} and lands close to it without coinciding.
The two routes yield an attribution of the same form and the template that turns it into privileged information is the same in both, as~\Cref{app:scaling_pac} sets out, and the attribution model is removed together with the verifier once training ends, so no configuration carries it at inference.

\paragraph{Remaining details.}
The experiment trains the 1.7B student under systems of two, four, six and eight agents, which is the only quantity that varies between the four configurations of~\Cref{fig:scaling}.
A system of $2n$ agents holds $n$ agents of each of the two role types of~\Cref{sec:mas}, the Coder and the Tester on the code domain and the Reasoner and the Tool-User on mathematics, so the number of role types and the multi-turn interaction between them are the same at every size.
Each domain trains on the dataset it uses throughout the paper, the code domain on CodeContests and mathematics on Polaris-Dataset-53K, and every configuration is evaluated on the same six benchmarks as~\Cref{tab:main_results} under the protocol of~\Cref{app:implementation_common}, from which the four configurations differ in the number of agents alone.
Each point of~\Cref{fig:scaling} is the mean over five independent training runs and the band around it is their standard deviation.
Panel (b) plots the six benchmarks separately rather than the two domain averages, since~\Cref{sec:rq_scaling} reads both the averages and the individual curves.

\subsection{Sensitivity to the Role-Advantage Weight}\label{app:lambda_sweep}

\begin{figure*}[t]
\centering
\includegraphics[width=1.0\linewidth]{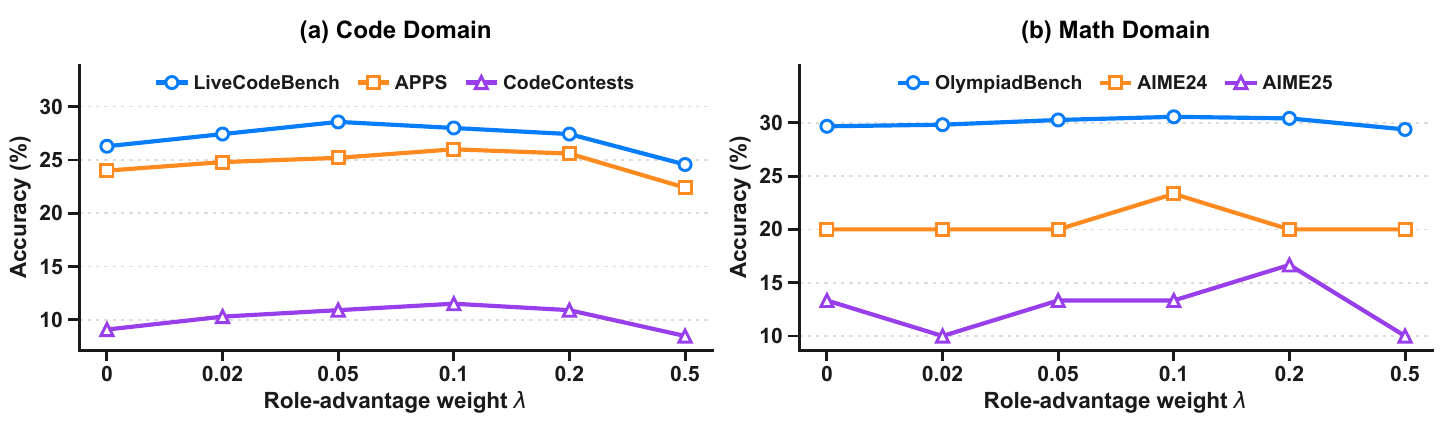}
\caption{
\textbf{Accuracy of \ourmodel as a function of the role-advantage weight $\lambda$.}
Each panel reports the three benchmarks of one domain for the 1.7B student, trained once per setting rather than repeated over seeds, so the settings $\lambda=0$ and $\lambda=0.1$ are single-run instances of the w/o RAS and \ourmodel systems of~\Cref{tab:ablation} rather than the five-run averages given there.
The two AIME sets contain thirty problems each, where one problem is worth $3.3$ points, which is why those curves move in larger steps than the other four.
The two panels share a common vertical span so that the dependence on $\lambda$ can be compared between the domains.
}
\label{fig:lambda_sweep}
\vspace{-0.09 in}
\end{figure*}

\ourmodel introduces a single hyperparameter, the weight $\lambda$ that~\Cref{eq:ras} places on the role advantage, and this subsection reports how the accuracy of the trained system depends on it.
$\lambda$ is held at the value given in~\Cref{app:implementation_ours} in every experiment reported elsewhere in this paper, and the sweep below varies it around that value to characterize the dependence.
We train one system at each of six values of $\lambda$ and evaluate all of them on the six benchmarks under the protocol of~\Cref{app:implementation_common}, so the configurations differ in $\lambda$ alone.
\Cref{fig:lambda_sweep} reports each benchmark separately, three curves per domain.
The leftmost setting $\lambda=0$ removes the role advantage while leaving the role condition of~\Cref{sec:role_conditioned} in place, and is therefore the w/o RAS system of~\Cref{tab:ablation}.

\begin{itemize}[leftmargin=*]

\item \textbf{Obs 13: accuracy is insensitive to the weight over a range spanning a factor of four.}
No benchmark anywhere between $\lambda=0.05$ and $\lambda=0.2$ falls below the accuracy it reaches with the role advantage removed, and within that range each curve varies by under $1.2$ points on the four larger benchmarks and by one problem on the two AIME sets.
The weight we report lies inside that range, so the role advantage does not have to be tuned for the conclusions of~\Cref{sec:rq2} to hold and any setting in the range supports them equally.

\item \textbf{Obs 14: removing the role advantage costs accuracy, which locates the gain in the contrast between the role conditions.}
Accuracy at $\lambda=0$ falls on most benchmarks and on the average of either domain, and that setting keeps the role condition of~\Cref{sec:role_conditioned} in place while removing only the difference between the two conditions.
What the module contributes is therefore the contrast between them rather than the role conditioning both settings share.

\item \textbf{Obs 15: too large a weight is worse than omitting the module altogether.}
Raising the weight to $\lambda=0.5$ puts most benchmarks and both domain averages below the system trained without the module at all, which is what~\Cref{eq:ras} leads one to expect: $\lambda$ scales the role advantage against the on-policy distillation term of~\Cref{eq:opd_adv}, and a large enough weight leaves the update tracking the difference between two role conditions rather than the token distribution of the teacher, so the students separate from one another without either learning to solve the task.

\end{itemize}

\definecolor{pcoral}{HTML}{E76F51}
\definecolor{psand}{HTML}{F4A261}
\definecolor{pgold}{HTML}{E9C46A}
\definecolor{pteal}{HTML}{2A9D8F}
\definecolor{pdeepteal}{HTML}{1D7874}
\definecolor{pslate}{HTML}{264653}
\definecolor{pplum}{HTML}{6D597A}
\definecolor{pinkdark}{HTML}{3A2E22}

\newcommand{\pacc}{pslate}

\newtcolorbox{promptbox}[3]{
  breakable,
  colback=#2!7,
  colframe=#2,
  colbacktitle=#2,
  coltitle=#3,
  fonttitle=\bfseries\small,
  title={#1},
  arc=1.6mm,
  boxrule=0.9pt,
  left=5pt, right=5pt, top=5pt, bottom=5pt,
  before upper={\renewcommand{\pacc}{#2}},
}

\newenvironment{prompttext}
  {\par\smallskip\begingroup\small\setlength{\parindent}{0pt}%
   \setlength{\parskip}{1.5pt}\raggedright\obeylines}
  {\endgroup\par\smallskip}

\newenvironment{promptflow}
  {\par\smallskip\begingroup\small\setlength{\parindent}{0pt}%
   \setlength{\parskip}{2pt}\raggedright}
  {\endgroup\par\smallskip}

\newcommand{\pfield}[1]{{\normalsize\bfseries\color{\pacc!78!black}#1}}
\newcommand{\popen}[1]{{\itshape\color{black!78}#1}}
\newcommand{\pcode}[1]{\texttt{#1}}

\newtcolorbox{ctxregion}[3]{
  enhanced, breakable=false,
  colback=#2!5, colframe=#2!45, boxrule=0.4pt, arc=1pt,
  borderline west={2.4pt}{0pt}{#2},
  left=6pt, right=5pt, top=3pt, bottom=3pt,
  fonttitle=\bfseries\scriptsize, coltitle=#2!72!black, colbacktitle=#2!16,
  title={#1\hfill\normalfont\scriptsize\itshape #3},
}

\newcommand{\elide}[1]{{\normalfont\itshape\color{pslate!62}$\langle$\,#1\,$\rangle$}}

\newtcolorbox{ctxloop}[1]{
  enhanced, breakable=false,
  colback=white, colframe=white, boxrule=0pt, arc=0pt,
  borderline={0.7pt}{1pt}{pslate!60, dashed},
  left=4pt, right=4pt, top=2pt, bottom=4pt,
  fonttitle=\bfseries\scriptsize, coltitle=pslate!85!black, colbacktitle=white,
  title={#1},
}

\newcommand{\prule}{\tcbline}
\newcommand{\ph}[1]{\textcolor{pplum}{\textbf{\texttt{\{#1\}}}}}
\newcommand{\lit}[1]{\textcolor{pslate}{\texttt{#1}}}
\newcommand{\cbq}{\texttt{\char96\char96\char96}}
\newcommand{\seg}[2]{\textcolor{#1}{\textbf{#2}}}
\newcommand{\swatch}[2]{\raisebox{-0.5pt}{\fcolorbox{#1}{#1!85}{\rule{0pt}{5pt}\rule{5pt}{0pt}}}\,\small #2}
\newcommand{\yes}{\textcolor{pteal}{\ding{51}}}
\newcommand{\no}{\textcolor{pcoral}{\ding{55}}}

\section{Prompt Templates}
\label{app:prompts}

This appendix collects every prompt \ourmodel uses on the two domains, reproduced as the models receive it.
The workflow templates of~\Cref{app:student_code,app:student_math} are shared by all methods we evaluate within our framework, and neither module alters them, so a student receives the same text during training and at deployment.
A teacher context is then assembled from three kinds of text, and the bar of every display below says which kind it holds.

\begin{center}
\setlength{\fboxsep}{0pt}
\begin{tabular}{@{}ll@{\hspace{2.2em}}l@{}}
\swatch{pcoral}{Coder} & \swatch{psand}{Tester} & \swatch{pplum}{Role condition, substituted by RAS} \\
\swatch{pteal}{Reasoner} & \swatch{pdeepteal}{Tool-User} & \swatch{pgold}{Attribution, inserted by PAC} \\
\end{tabular}
\end{center}

Warm bars mark the code domain and teal bars the mathematics domain.
A \seg{pplum}{role condition} replaces a passage the student also receives and asserts nothing the student does not already hold, so it carries no privilege, whereas an \seg{pgold!55!black}{attribution} is the one segment no student ever sees.
Within a display, \ph{placeholders} are filled in at run time.

\subsection{Student Prompts on the Code Domain}
\label{app:student_code}

The Coder writes a Python program and the Tester writes a unit test case made of a test input and an expected output, and the environment runs the program on that input and compares the two outputs.
In the later turns both roles are asked in the same terms to locate the inconsistency and then to revise or to keep their own artefact, so neither of them is predisposed to concede; were one side told to suspect itself first, the conflict outcomes we report in~\Cref{sec:experiment} could be read off the prompt rather than off training.
The opening line of each template is the passage a role condition replaces when a teacher context is formed.

\begin{promptbox}{Coder, first turn}{pcoral}{white}
\textbf{Input.} The problem description \ph{problem}.
\prule
\textbf{Prompt.}
\begin{prompttext}
\popen{You are a helpful assistant that writes Python to solve the problem.}
\popen{Think step by step, then output code.}

\pfield{Important:}
- Read all inputs via \pcode{input()}.
- Print all results with \pcode{print()}.
- Do not hardcode inputs.

\pfield{Problem:}
\ph{problem}

First, decide on the number and types of inputs required (e.g.,
\pcode{x = int(input())}, \pcode{b = int(input())}), then implement the solution
and print the result.

\pfield{Please answer in the following format:}
\lit{Code:} \cbq \pcode{python (your code here)} \cbq
\end{prompttext}
\prule
\textbf{Output.} A program.
\end{promptbox}

\begin{promptbox}{Tester, first turn}{psand}{pinkdark}
\textbf{Input.} The problem description \ph{problem}.
\prule
\textbf{Prompt.}
\begin{prompttext}
\popen{You are a helpful assistant that creates unit test cases (input +}
\popen{expected output) for a coding task.}

\pfield{Important:}
- The test input must follow exactly the input format stated in the problem.
- The expected output must be derived from the problem specification.

\pfield{Problem:}
\ph{problem}

First, identify the input format required by the problem, then construct
a valid test input and derive its expected output from the specification.

\pfield{Please answer in the following format:}
\lit{Test Input:} \cbq \pcode{(your test input here)} \cbq
\lit{Test Output:} \cbq \pcode{(the expected output here)} \cbq
\end{prompttext}
\prule
\textbf{Output.} A test case.
\end{promptbox}

\begin{promptbox}{Coder, later turns}{pcoral}{white}
\textbf{Input.} The problem description \ph{problem} and the mismatch history \ph{mismatch\_history}, a record of previous programs, test inputs, expected outputs and actual execution outputs.
\prule
\textbf{Prompt.}
\begin{prompttext}
\popen{You are a helpful assistant that corrects and refines code.}

\pfield{Important:}
- Read inputs via \pcode{input()}; output with \pcode{print()}.
- Do not hardcode inputs.

\pfield{Problem:}
\ph{problem}

\pfield{Use the history below to guide your decision:}
\ph{mismatch\_history}

If the previous program crashed, first fix the bug.

If execution succeeded but the produced output did not match the expected
output, decide where the inconsistency comes from.
- If it comes from the program, refine the program so that it satisfies
\hspace*{1em}the problem specification.
- If it comes from the test case, keep the program and briefly state why
\hspace*{1em}it already satisfies the specification.

\pfield{Provide the final program. Respond in the format:}
\lit{Code:} \cbq \pcode{python (your code here)} \cbq
\end{prompttext}
\prule
\textbf{Output.} A program.
\end{promptbox}

\begin{promptbox}{Tester, later turns}{psand}{pinkdark}
\textbf{Input.} The problem description \ph{problem} and the mismatch history \ph{mismatch\_history}, showing the test case and the differing execution output of the program.
\prule
\textbf{Prompt.}
\begin{prompttext}
\popen{You are a helpful assistant that corrects and refines unit test cases.}

\pfield{Important:}
- The test input must follow exactly the input format stated in the problem.
- The expected output must be derived from the problem specification.

\pfield{Problem:}
\ph{problem}

\pfield{Use the history below to guide your decision:}
\ph{mismatch\_history}

If the produced output did not match the expected output, decide where the
inconsistency comes from.
- If it comes from the test case, correct the test input or the expected
\hspace*{1em}output so that both follow the problem specification.
- If it comes from the program, keep the test case and briefly state why
\hspace*{1em}it already follows the specification.

\pfield{Provide the final test case. Respond in the format:}
\lit{Test Input:} \cbq \pcode{(your test input here)} \cbq
\lit{Test Output:} \cbq \pcode{(the expected output here)} \cbq
\end{prompttext}
\prule
\textbf{Output.} A test case.
\end{promptbox}

\subsection{Student Prompts on the Mathematics Domain}
\label{app:student_math}

The Reasoner derives the answer by mathematical reasoning and the Tool-User computes it by writing and executing a Python program, and the environment compares the two answers.
The later turns are again symmetric across the two roles.

\begin{promptbox}{Reasoner, first turn}{pteal}{white}
\textbf{Input.} The problem \ph{problem}.
\prule
\textbf{Prompt.}
\begin{prompttext}
\popen{You are a helpful assistant that solves math problems via careful reasoning.}

\pfield{Problem:}
\ph{problem}

Work through the problem step by step, then state the final answer.

\pfield{Please answer in the following format:}
\lit{Reasoning Steps:} \pcode{(your derivation here)}
\lit{\#\#\#\#} \pcode{(the final answer here)}
\end{prompttext}
\prule
\textbf{Output.} A derivation and a final answer.
\end{promptbox}

\begin{promptbox}{Tool-User, first turn}{pdeepteal}{white}
\textbf{Input.} The same problem \ph{problem}.
\prule
\textbf{Prompt.}
\begin{prompttext}
\popen{You are a helpful programming assistant that writes Python to solve the}
\popen{math problem.}

\pfield{Important:}
- Prefer exact arithmetic (fractions, integers, rational simplification)
\hspace*{1em}when possible.

\pfield{Problem:}
\ph{problem}

Print only the final answer.

\pfield{Please answer in the following format:}
\lit{Code:} \cbq \pcode{python (your code here)} \cbq
\end{prompttext}
\prule
\textbf{Output.} A program printing the answer.
\end{promptbox}

\begin{promptbox}{Reasoner, later turns}{pteal}{white}
\textbf{Input.} The problem \ph{problem} and the mismatch history \ph{mismatch\_history}, holding the prior derivation and its final answer together with the program and its printed output.
\prule
\textbf{Prompt.}
\begin{prompttext}
\popen{You are a helpful assistant that refines mathematical solutions through}
\popen{reasoning.}

\pfield{Problem:}
\ph{problem}

\pfield{History (previous attempts and outputs):}
\ph{mismatch\_history}

The answer obtained by reasoning and the printed result of the program
disagree. Decide where the inconsistency comes from.
- If it comes from the derivation, correct the faulty step and adopt the
\hspace*{1em}corrected value.
- If it comes from the program, keep the derived answer and briefly state
\hspace*{1em}why the derivation is sound.

\pfield{Provide the final answer. Respond in the format:}
\lit{Reasoning Steps:} \pcode{(your derivation here)}
\lit{\#\#\#\#} \pcode{(the final answer here)}
\end{prompttext}
\prule
\textbf{Output.} A derivation and a final answer.
\end{promptbox}

\begin{promptbox}{Tool-User, later turns}{pdeepteal}{white}
\textbf{Input.} The problem \ph{problem} and the mismatch history \ph{mismatch\_history}, holding the prior program and its printed output together with the derivation and its final answer.
\prule
\textbf{Prompt.}
\begin{prompttext}
\popen{You are a helpful programming assistant that refines Python solutions for}
\popen{math problems.}

\pfield{Important:}
- Prefer exact arithmetic (fractions, integers, rational simplification)
\hspace*{1em}when possible.

\pfield{Problem:}
\ph{problem}

\pfield{History (previous attempts and outputs):}
\ph{mismatch\_history}

The printed result of the program and the answer obtained by reasoning
disagree. Decide where the inconsistency comes from.
- If it comes from the program, fix the defect (numerical stability, edge
\hspace*{1em}cases, exact versus floating point) and recompute.
- If it comes from the derivation, keep the program and briefly state why
\hspace*{1em}the computation is reliable.

\pfield{Provide the final program. Respond in the format:}
\lit{Code:} \cbq \pcode{python (your code here)} \cbq
\end{prompttext}
\prule
\textbf{Output.} A program printing the answer.
\end{promptbox}

\subsection{Role Conditions}
\label{app:role_conditions}

A role condition takes the place of the opening line of a student template and is the only part of a teacher context that differs between the two evaluations of one response, as~\Cref{eq:teacher_input} sets out.
The four conditions share one structure, a first paragraph naming the role together with its collaborator and its primary responsibility and a second paragraph naming what the role centres on together with the part it must not take over.
Holding the four isomorphic and of comparable length is what keeps the role advantage of~\Cref{eq:role_adv} from measuring differences in wording rather than in responsibility.

\begin{promptbox}{Role condition, Coder}{pplum}{white}
\label{box:role_coder}
\small
You are the \textbf{Coder} in a collaborative programming system, working together with a \textbf{Tester} to solve the given programming problem through iterative interaction. Your primary responsibility is to analyze the task, develop an appropriate algorithmic solution, and produce a correct implementation. You should use the task specification and any feedback provided during the interaction to refine your reasoning and code when necessary.
\prule
Your role is centered on \textbf{solution design and program implementation}. You may inspect and reason about testing feedback when it is provided, but you should not take over the Tester's primary responsibility of constructing test inputs or independently deriving their expected outputs. Those activities belong to the \textbf{Tester}.
\end{promptbox}

\begin{promptbox}{Role condition, Tester}{pplum}{white}
\label{box:role_tester}
\small
You are the \textbf{Tester} in a collaborative programming system, working together with a \textbf{Coder} to verify and improve the solution to the given programming problem through iterative interaction. Your primary responsibility is to evaluate the proposed solution by constructing informative test inputs, deriving their correct expected outputs, and identifying cases that may expose logical errors, corner cases, or incorrect assumptions. You should use the task specification and any feedback provided during the interaction to refine your test cases when necessary.
\prule
Your role is centered on \textbf{solution validation and error discovery}. You may analyze the Coder's reasoning or implementation when necessary to design effective tests and provide useful feedback, but you should not take over the Coder's primary responsibility of designing the algorithm or implementing the final program. Those activities belong to the \textbf{Coder}.
\end{promptbox}

\begin{promptbox}{Role condition, Reasoner}{pplum}{white}
\label{box:role_reasoner}
\small
You are the \textbf{Reasoner} in a collaborative mathematical problem-solving system, working together with a \textbf{Tool-User} to solve the given problem through iterative interaction. Your primary responsibility is to analyze the problem, carry out a sound mathematical derivation, and arrive at the final answer through reasoning. You should use the problem statement and any feedback provided during the interaction to refine your derivation when necessary.
\prule
Your role is centered on \textbf{analytic derivation and mathematical justification}. You may inspect and reason about the program and its printed result when they are provided, but you should not take over the Tool-User's primary responsibility of writing or debugging the computational program. Those activities belong to the \textbf{Tool-User}.
\end{promptbox}

\begin{promptbox}{Role condition, Tool-User}{pplum}{white}
\label{box:role_tooluser}
\small
You are the \textbf{Tool-User} in a collaborative mathematical problem-solving system, working together with a \textbf{Reasoner} to solve the given problem through iterative interaction. Your primary responsibility is to compute the answer by writing a correct and numerically reliable program, executing it, and handling precision and edge cases properly. You should use the problem statement and any feedback provided during the interaction to refine your program when necessary.
\prule
Your role is centered on \textbf{computation and numerical verification}. You may analyze the Reasoner's derivation when necessary to decide what the program should compute, but you should not take over the Reasoner's primary responsibility of producing the analytic derivation. Those activities belong to the \textbf{Reasoner}.
\end{promptbox}

\subsection{Composition of a Teacher Context}
\label{app:teacher_context}

Ordinary on-policy distillation hands the teacher the student trajectory itself: to force decode a response the teacher is given the trajectory as that agent saw it, ending exactly where the response begins, which is the input $x_i^t$ of~\Cref{sec:mas} and carries the teammate's outputs through the interaction history.
\ourmodel builds its teacher context out of that same trajectory by two insertions and nothing else, so~\Cref{fig:compose_code,fig:compose_math} read as the trajectory with the two places our modules act on marked out.
An edge in \seg{pslate}{slate} is trajectory text carried over unchanged, an edge in \seg{pplum}{plum} is where RAS substitutes the role, and an edge in \seg{pgold!55!black}{gold} is where PAC inserts the attribution.
Every region states on its right whether the student holds it too, and the asymmetry of the construction is visible in that column alone: the two inserted regions read \emph{teacher only} and no student receives either of them at any turn.
The displays read as a timeline, and the second insertion recurs along it: a turn is verified once it is complete, and its verdict is placed ahead of the turn that follows, which is the turn it guides.
Turn 0 therefore stands alone with nothing before it, and from the second turn onward the teacher enters each turn knowing where the previous one went wrong while the students enter it holding the same records with every verdict withheld.

\begin{tcolorbox}[enhanced, breakable, colback=white, colframe=pcoral,
  boxrule=1pt, arc=1.6mm, left=4pt, right=4pt, top=4pt, bottom=4pt,
  fonttitle=\bfseries\small, colbacktitle=pcoral, coltitle=white,
  title={Teacher context on the code domain, scoring a Coder response at turn $t\ge1$}]

\begin{ctxregion}{\ding{182}\; RAS substitutes the role here}{pplum}{\bfseries teacher only}
\small
The trajectory opens with \popen{You are a helpful assistant that corrects and refines code.}, and RAS puts the Coder condition of~\Cref{box:role_coder} in its place for the target context and the Tester condition of~\Cref{box:role_tester} for the non-target one.
This region is the sole difference between the two contexts of a pair.
\end{ctxregion}

\begin{ctxregion}{\ding{183}\; Turn 0 \textnormal{— student trajectory}}{pslate}{student and teacher}
\small
\elide{the problem as the Coder template of~\Cref{app:student_code} posed it, and what both roles submitted at turn 0 together with the disagreement the environment reported}
\end{ctxregion}

\begin{ctxloop}{Repeated for $t'=1,\ldots,t$, a verdict on the preceding turn ahead of the turn it guides}

\begin{ctxregion}{\ding{184}\; Verdict on turn $t'-1$ \textnormal{— attached by PAC}}{pgold}{\bfseries teacher only}
\begin{prompttext}
\pfield{Verified attribution (available during training only):}
Verification result for turn \ph{t'-1}:
- \ph{observation}
- \ph{conclusion}
\end{prompttext}
\vspace{-2pt}
\small
Produced by the verification of~\Cref{app:pac_attribution} and read by the teacher as it scores turn $t'$.
\end{ctxregion}

\begin{ctxregion}{\ding{185}\; Turn $t'$ \textnormal{— student trajectory}}{pslate}{student and teacher}
\small
\elide{for $t'<t$, what both roles submitted at that turn and the disagreement reported; for $t'=t$, the decision the role faces and the response format, after which the response being scored begins}
\end{ctxregion}

\end{ctxloop}

\end{tcolorbox}

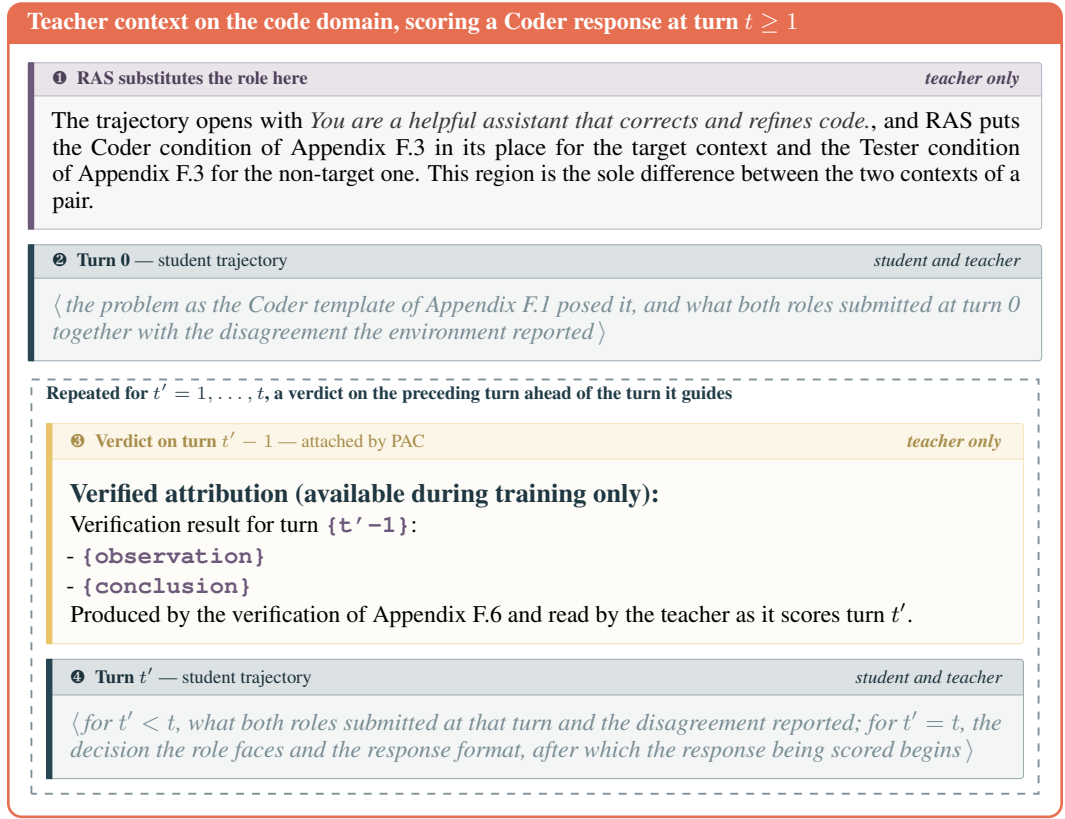
\captionof{figure}{\textbf{Composition of a teacher context on the code domain.} The display reads as a timeline. Turn 0 stands alone because nothing has yet been verified when it begins, and from then on each turn is preceded by the verdict on the turn before it, which is the pair the dashed block repeats. Regions~\ding{183} and~\ding{185} are the student trajectory carried over unchanged and are elided here, being the Coder template of~\Cref{app:student_code} as a rollout filled it in; ordinary on-policy distillation would hand the teacher those alone. Region~\ding{182} is where RAS replaces the line the trajectory opens with, which is what it varies between the two evaluations of one response, and region~\ding{184} is a verdict the students never receive at any turn. Scoring a Tester response exchanges the two role conditions and draws the trajectory as the Tester saw it.}
\label{fig:compose_code}
\vspace{2 mm}

A context is assembled afresh at every turn and the timeline lengthens by a verdict and a turn at a time, so scoring a turn-0 response calls for no verdict at all, scoring a turn-1 response has the teacher read turn 0 and the verdict on it before turn 1, and scoring a response at turn $t$ has it read a verdict on each of the $t$ earlier turns.
A student reads none of those verdicts at any point.
Nothing has been verified when turn 0 begins, which is the sense in which PAC does not act there.

\begin{tcolorbox}[enhanced, breakable, colback=white, colframe=pteal,
  boxrule=1pt, arc=1.6mm, left=4pt, right=4pt, top=4pt, bottom=4pt,
  fonttitle=\bfseries\small, colbacktitle=pteal, coltitle=white,
  title={Teacher context on the mathematics domain, scoring a Reasoner response at turn $t\ge1$}]

\begin{ctxregion}{\ding{182}\; RAS substitutes the role here}{pplum}{\bfseries teacher only}
\small
The trajectory opens with \popen{You are a helpful assistant that refines mathematical solutions through reasoning.}, and RAS puts the Reasoner condition of~\Cref{box:role_reasoner} in its place for the target context and the Tool-User condition of~\Cref{box:role_tooluser} for the non-target one.
\end{ctxregion}

\begin{ctxregion}{\ding{183}\; Turn 0 \textnormal{— student trajectory}}{pslate}{student and teacher}
\small
\elide{the problem as the Reasoner template of~\Cref{app:student_math} posed it, and what both roles submitted at turn 0 together with the disagreement the environment reported}
\end{ctxregion}

\begin{ctxloop}{Repeated for $t'=1,\ldots,t$, a verdict on the preceding turn ahead of the turn it guides}

\begin{ctxregion}{\ding{184}\; Verdict on turn $t'-1$ \textnormal{— attached by PAC}}{pgold}{\bfseries teacher only}
\begin{prompttext}
\pfield{Verified attribution (available during training only):}
Verification result for turn \ph{t'-1}:
- \ph{observation}
- \ph{conclusion}
\end{prompttext}
\vspace{-2pt}
\small
The verdict rests on the reference answer, yet it names which of the two candidate values agrees with it and never states the reference answer itself.
\end{ctxregion}

\begin{ctxregion}{\ding{185}\; Turn $t'$ \textnormal{— student trajectory}}{pslate}{student and teacher}
\small
\elide{for $t'<t$, what both roles submitted at that turn and the disagreement reported; for $t'=t$, the decision the role faces and the response format, after which the response being scored begins}
\end{ctxregion}

\end{ctxloop}

\end{tcolorbox}

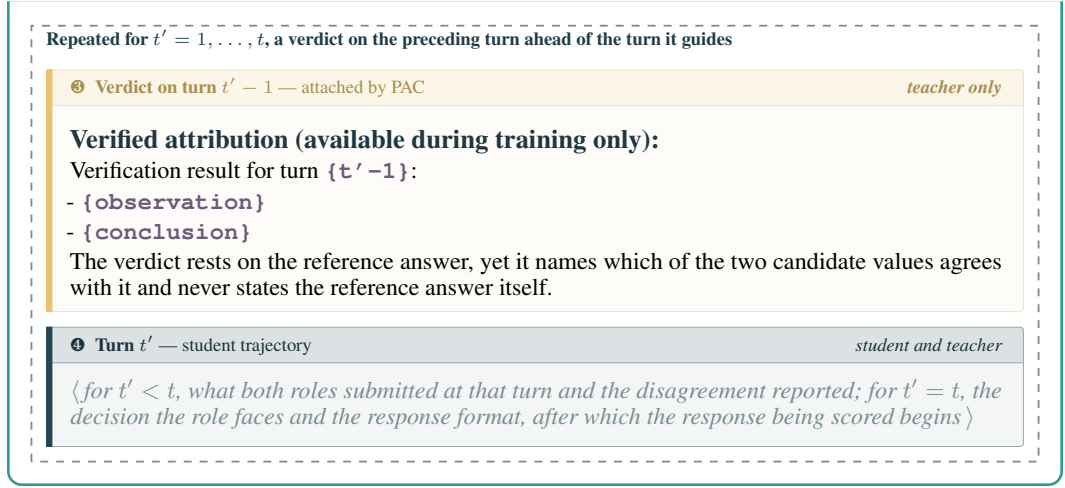
\captionof{figure}{\textbf{Composition of a teacher context on the mathematics domain.} The construction matches the code domain region for region, with the derived answer and the printed result standing in for the program output and the declared expected output. Scoring a Tool-User response exchanges the two role conditions of region~\ding{182} and draws the trajectory as the Tool-User saw it.}
\label{fig:compose_math}
\vspace{2 mm}

\subsection{The Two Teacher Contexts of RAS}
\label{app:ras_contexts}

Scoring one response calls for two of these contexts, each force decoded once, and we write the pair out separately below.
What the two share is abbreviated as \ph{student trajectory context}, standing for regions~\ding{183} to~\ding{185} of~\Cref{fig:compose_code,fig:compose_math}, the verdicts included, none of which differs between them.

\begin{promptbox}{Code domain, target role context for a Coder response}{pcoral}{white}
\label{box:ras_code_target}
\begin{prompttext}
[the Coder role condition, \Cref{box:role_coder}]

\ph{student trajectory context}
\end{prompttext}
\prule
\textbf{Output.} The teacher score $\ell_k^{(i)}$ of~\Cref{eq:teacher_score} at every token position.
\end{promptbox}

\begin{promptbox}{Code domain, non-target role context for the same Coder response}{psand}{pinkdark}
\label{box:ras_code_nontarget}
\begin{prompttext}
[the Tester role condition, \Cref{box:role_tester}]

\ph{student trajectory context}
\end{prompttext}
\prule
\textbf{Output.} The teacher score $\ell_k^{(j)}$ at the same positions.
\end{promptbox}

\begin{promptbox}{Mathematics domain, target role context for a Reasoner response}{pteal}{white}
\label{box:ras_math_target}
\begin{prompttext}
[the Reasoner role condition, \Cref{box:role_reasoner}]

\ph{student trajectory context}
\end{prompttext}
\prule
\textbf{Output.} The teacher score $\ell_k^{(i)}$ at every token position.
\end{promptbox}

\begin{promptbox}{Mathematics domain, non-target role context for the same Reasoner response}{pdeepteal}{white}
\label{box:ras_math_nontarget}
\begin{prompttext}
[the Tool-User role condition, \Cref{box:role_tooluser}]

\ph{student trajectory context}
\end{prompttext}
\prule
\textbf{Output.} The teacher score $\ell_k^{(j)}$ at the same positions.
\end{promptbox}

The teacher force decodes the same response under both contexts of a pair, and the difference between the two scores is the role advantage of~\Cref{eq:role_adv}.
Because the contexts agree word for word outside the role condition, that difference expresses the preference of the teacher under the two identities and nothing else.
The target context is also the one that yields the OPD advantage of~\Cref{eq:mas_opd_adv}, so its forward pass is shared and one response costs one additional teacher forward pass.
Scoring a Tester or a Tool-User response exchanges the two role conditions of the pair and draws the student segments from that role's own template.

\subsection{Attribution in PAC}
\label{app:pac_attribution}

\paragraph{What an attribution supplies.}
A conflict leaves two contradictory candidate values in the student context, the output the program actually produced against the output the test case declared, or the answer reasoning arrived at against the result the program printed.
The students see both and the workflow asks them which to trust while giving them nothing to decide with.
An attribution supplies that missing criterion by pointing at one of the two values rather than by introducing a third, and three properties follow from doing it this way.
It labels a candidate the agents produced instead of stating a value of its own, so no reference solution and no reference answer ever appears in it.
The teacher therefore learns nothing beyond a label on outputs already in front of it, and when both candidates are wrong it is told only that, which leaves it without a correct value as well.
The decision to revise or to hold turns on the two Boolean checks described below, so what reaches the teacher is a criterion that has been checked against the problem rather than a judgement it has to form on its own.

\paragraph{The verification that produces it.}
The attribution $a^t$ of~\Cref{eq:attribution} follows from a pair of checks, each reading the output of one role against a training-time reference and never the output of the other role, and that independence is what lets a failing check name the role responsible rather than merely report a disagreement.
On the code domain \lit{program\_correct} holds when the student program passes the reference tests of the problem, and \lit{test\_correct} holds when the expected output the student declared is the one the reference supports for the values it declared, so the first check reads the program alone and the second the test case alone.
Both checks concern what a role worked out rather than how it laid the result out: whether a test case is well formed is a separate property, scored by the validity term of~\Cref{app:reward_code} and deliberately not one of the two checks here, since a disagreement owed to layout alone is one in which neither role reasoned wrongly.
On the mathematics domain \lit{reasoning\_correct} holds when the answer stated after \lit{\#\#\#\#} is equivalent to the reference answer and \lit{program\_correct} holds when the printed result is, the equivalence being decided by numerical tolerance and by symbolic simplification in turn.

\paragraph{The template.}
All four outcomes share the two-line form of~\Cref{box:pac_template}, an observation restating the two candidate values followed by a conclusion naming the side to trust.
Every field is drawn from the outputs of the agents, and the first failing sample is used on the code domain when several exist.

\begin{promptbox}{Attribution template, shared by the four outcomes}{pgold}{pinkdark}
\label{box:pac_template}
\begin{prompttext}
\pfield{Verification result for the previous turn:}
- \ph{observation}
- \ph{conclusion}
\end{prompttext}
\prule
\seg{pinkdark}{Observation, code domain.}
\begin{promptflow}
On test input \ph{test\_input}, the program printed \ph{program\_output}, while the declared expected output was \ph{declared\_expected\_output}.
\end{promptflow}
\prule
\seg{pinkdark}{Observation, mathematics domain.}
\begin{promptflow}
The derivation reported \ph{reasoning\_answer} and the program printed \ph{program\_answer}.
\end{promptflow}
\end{promptbox}

\paragraph{The four conclusions on the code domain.}
\begin{itemize}[leftmargin=*]

    \item \textbf{\texttt{PROGRAM\_INCONSISTENT}, the program fails \no{} and the test case holds \yes.}
    The test case is consistent with the reference, the program is not, and the previous inconsistency therefore originates from the program.

    \item \textbf{\texttt{TEST\_INCONSISTENT}, the program holds \yes{} and the test case fails \no.}
    The program is consistent with the reference, the declared expected output is not, and the previous inconsistency therefore originates from the test case.

    \item \textbf{\texttt{BOTH\_INCONSISTENT}, neither holds \no\no.}
    Neither the program nor the test case is consistent with the reference.

    \item \textbf{\texttt{BOTH\_CONSISTENT}, both hold \yes\yes.}
    Neither check finds fault with the artefact it reads, so no side is named and the teacher is told only that the disagreement survives both checks, a case the formatting of the input or the output commonly produces.

\end{itemize}

In the last case neither artefact is contradicted by the reference on its own, so there is no ground for making either role give way and the conclusion directs both roles at the interface between their outputs rather than at either output.
That is where such a conflict usually sits, as when the test input is laid out differently from the format the problem states and the program reads it wrongly, and putting it there is what spares the teacher from reconciling a conflict between two sides it has just been told are both correct.

\paragraph{The four conclusions on the mathematics domain.}
\begin{itemize}[leftmargin=*]

    \item \textbf{\texttt{DERIVATION\_INCONSISTENT}, the derivation fails \no{} and the program holds \yes.}
    The printed result is consistent with the reference answer, the derived answer is not, and the previous inconsistency therefore originates from the derivation.

    \item \textbf{\texttt{COMPUTATION\_INCONSISTENT}, the derivation holds \yes{} and the program fails \no.}
    The derived answer is consistent with the reference answer, the printed result is not, and the previous inconsistency therefore originates from the program.

    \item \textbf{\texttt{BOTH\_INCONSISTENT}, neither holds \no\no.}
    Neither is consistent with the reference answer.

    \item \textbf{\texttt{BOTH\_CONSISTENT}, both hold \yes\yes.}
    Both the derivation and the program are consistent with the reference answer, and the previous inconsistency therefore originates from a difference in answer formatting rather than from either of them.

\end{itemize}

\paragraph{Where it enters and when.}
An attribution occupies region~\ding{184} of~\Cref{fig:compose_code,fig:compose_math}, ahead of the turn it guides, and it is the same text in both evaluations of a pair.
A turn is judged only once it is complete, as~\Cref{eq:priv_state} states, which keeps the teacher from seeing information determined by the very response it is scoring and leaves the first turn with no verdict at all.
No student context carries the segment at any turn, and the verifier along with the attribution is removed once training ends, after which the roles decide whether to revise or to hold from their own observations alone.

\paragraph{An instance on the code domain.}
\Cref{box:pac_instance} takes a problem asking for the number of integers in a closed interval, where an off-by-one error is available to both roles, and contrasts what the students hold at the second turn against what the teacher additionally receives.
The value \lit{5} appears in the attribution because the Tester declared it, while the closed form \lit{b - a + 1} and the golden reference solution stay undisclosed.

\begin{promptbox}{A conflict as the students and the teacher each see it}{pslate}{white}
\label{box:pac_instance}
\small
\seg{pinkdark}{Problem.} Read two integers $a$ and $b$, one per line, and print the number of integers in the closed interval $[a,b]$. At turn 0 the Coder computed \lit{b - a} and the Tester declared the expected output \lit{5} for the input \lit{3} followed by \lit{7}.
\prule
\seg{pinkdark}{Held by both students.} The declared expected output \lit{5} and the program output \lit{4} on the test input, and the fact that the two disagree. Which of the two values the specification supports is not among them, and each role is nonetheless asked to decide.
\prule
\seg{pgold!55!black}{Received by the teacher in addition.} That \lit{5} is the consistent one \yes{} and \lit{4} is not \no, stated as follows.

\begin{promptflow}
Verification result for the previous turn:

- On test input "3\textbackslash n7", the program printed 4, while the declared expected output was 5.

- The test case is consistent with the reference, the program is not, and the previous inconsistency therefore originates from the program.
\end{promptflow}
\end{promptbox}

\paragraph{An instance on the mathematics domain.}
\Cref{box:pac_instance_math} contrasts the same two viewpoints where the derivation and the program disagree on the final answer.
The value \lit{36} appears in the attribution because the Tool-User printed it, and the reference answer is never stated in its own right.

\begin{promptbox}{A conflict on the mathematics domain as each side sees it}{pslate}{white}
\label{box:pac_instance_math}
\small
\seg{pinkdark}{Held by both students.} The derived answer \lit{48}, the printed result \lit{36} and the fact that the two disagree, with nothing to say which of them the reference answer supports.
\prule
\seg{pgold!55!black}{Received by the teacher in addition.} That \lit{36} is the consistent one \yes{} and \lit{48} is not \no, stated as follows.

\begin{promptflow}
Verification result for the previous turn:

- The derivation reported 48 and the program printed 36.

- The printed result is consistent with the reference answer, the derived answer is not, and the previous inconsistency therefore originates from the derivation.
\end{promptflow}
\end{promptbox}

Both displays say what each side holds rather than reproduce a trajectory, and the problems and the numbers in them stand in for a logged interaction.

\paragraph{What the asymmetry produces.}
Conditioned on the attribution the teacher writes token targets that revise the side at fault and hold the side that is sound, whereas without it the teacher reads the same interaction as the students and is equally unable to tell which side to trust, leaving its targets to express local quality alone and to carry no direction of concession.
The students hold the two candidate values without the label, so the only way open to them for lowering the distillation loss is learning to tell from the problem statement and from their own derivation which side is at fault.
What they internalize is therefore the ability to recognize their own errors rather than a correct value they were shown, since which value is correct is never disclosed to them.

\section{Case Study}
\label{app:casestudy}

\raggedbottom

\newcommand{\pblank}{\vspace{0.35\baselineskip}}

\newcommand{\casehue}{pcoral}

\newcommand{\tagword}[1]{%
  \colorbox{\casehue}{\color{white}\scriptsize\sffamily\bfseries\upshape\hspace{0.5pt}#1\hspace{0.5pt}}%
}

\newcommand{\cnote}[1]{%
  \par\vspace{3.5pt}%
  \begingroup
  \leftskip=14pt \rightskip=14pt
  \footnotesize\itshape\color{pslate!88}%
  \noindent\hspace*{-14pt}\tagword{ANALYSIS}\hspace{5pt}\ignorespaces #1%
  \par\endgroup\vspace{2.5pt}%
}

\definecolor{rolepos}{HTML}{2E6FB7}
\definecolor{roleneg}{HTML}{C97A1E}
\newcommand{\tkbox}[2]{{\setlength{\fboxsep}{1.15pt}\colorbox{#1}{\strut #2}}\allowbreak}
\newcommand{\pE}[1]{\tkbox{rolepos!56}{#1}}    %
\newcommand{\pD}[1]{\tkbox{rolepos!45}{#1}}    %
\newcommand{\pC}[1]{\tkbox{rolepos!35}{#1}}    %
\newcommand{\pB}[1]{\tkbox{rolepos!25}{#1}}    %
\newcommand{\pA}[1]{\tkbox{rolepos!15}{#1}}    %
\newcommand{\tkO}[1]{\tkbox{black!12}{#1}}     %
\newcommand{\nA}[1]{\tkbox{roleneg!15}{#1}}    %
\newcommand{\nB}[1]{\tkbox{roleneg!26}{#1}}    %
\newcommand{\nC}[1]{\tkbox{roleneg!38}{#1}}    %
\newcommand{\nD}[1]{\tkbox{roleneg!50}{#1}}    %
\newcommand{\nE}[1]{\tkbox{roleneg!64}{#1}}    %
\newcommand{\tkchip}[1]{#1{\rule{6pt}{0pt}\rule{0pt}{5.5pt}}}

\newcommand{\raskey}[2]{%
  \par\vspace{2pt}\noindent{\scriptsize
  towards the #1\;%
  \tkchip{\nE}\tkchip{\nD}\tkchip{\nC}\tkchip{\nB}\tkchip{\nA}\tkchip{\tkO}\tkchip{\pA}\tkchip{\pB}\tkchip{\pC}\tkchip{\pD}\tkchip{\pE}%
  \;towards the #2%
  \hfill the middle level is a role advantage near zero%
  \par}\vspace{3pt}}

\newtcolorbox{caseregion}[3]{
  enhanced, breakable,
  colback=#2!5, colframe=#2!45, boxrule=0.4pt, arc=1pt,
  borderline west={2.4pt}{0pt}{#2},
  left=6pt, right=5pt, top=3pt, bottom=3pt,
  fonttitle=\bfseries\scriptsize, coltitle=#2!72!black, colbacktitle=#2!16,
  title={#1\hfill\normalfont\scriptsize\itshape #3},
}

\newcommand{\casetitle}[2]{%
  \par\vspace{4pt}%
  {\color{#2}\rule{\linewidth}{1pt}}\par\vspace{-2pt}%
  \noindent{\small\bfseries\color{#2!80!black}#1}%
  \par\vspace{2pt}%
}

\newcommand{\casefoot}[1]{%
  \par\vspace{1pt}%
  {\color{#1}\rule{\linewidth}{1pt}}%
  \par\vspace{2pt}%
}

Two kinds of case are collected here, one per module.
The first follows an interaction through a conflict and shows what each side holds while it is being resolved, which is where PAC acts.
The second colours a response by the role advantage of~\Cref{eq:role_adv} and shows where the signal RAS adds actually falls, which is a property of single tokens rather than of a turn.
Each is given on the code domain and on the mathematics domain.

\subsection{An Interaction Through a Conflict}
\label{app:case_interaction}

\Cref{app:teacher_context} draws a teacher context as a timeline and elides the regions that a rollout fills in.
This subsection supplies them, with the regions in the same order and the same colours and each marked with who holds it.
The students hold the slate regions and nothing else, and the two teacher-only regions are the substitution RAS makes and the verdict PAC attaches.

\subsubsection{Code Domain}
\label{app:case_code}

\renewcommand{\casehue}{pcoral}

\paragraph{What this case shows.}
A mismatch is ordinarily read as one side being wrong, and this interaction is one where neither is.
The program implements the formula the specification states, the expected output is the value that specification gives, and the two disagree because the test input was laid out in the notation of the worked examples while the program reads one value per line, which is a discrepancy neither role can locate from the record they share.
The verification reaches that conclusion because its two checks read one artefact each and neither reads the other, so both are free to hold at once and the teacher is told where the conflict actually lies rather than which side to blame.
The turn that follows is the one worth reading: the Coder keeps its program and says why, the Tester treats the layout rather than its own reasoning as the fault, and the pair converges on a case the task actually poses.

\casetitle{A conflict on the code domain, scoring the Coder response of turn 1}{pcoral}

\begin{caseregion}{\ding{182}\; RAS substitutes the role here}{pplum}{\bfseries teacher only}
\small
The target context opens with the Coder condition of~\Cref{box:role_coder} and the non-target one with the Tester condition of~\Cref{box:role_tester}, in place of the line the trajectory opens with.
Everything below is identical in the two.
\end{caseregion}

\begin{caseregion}{\ding{183}\; Turn 0 \textnormal{— student trajectory}}{pslate}{student and teacher}
\small
\seg{pslate}{Problem.} Held by both roles, in the \pfield{Problem:} field of the templates of~\Cref{app:student_code}.

\begin{prompttext}
A deck is an \pcode{n x n} grid of cells and each cell holds at most one container of weight \pcode{w}. The total weight of the containers loaded may not exceed \pcode{maxWeight}. Report how many containers can be loaded.

\elide{the input and output format, the constraints on the three values and the worked examples}
\end{prompttext}

\vspace{2pt}
\seg{pcoral}{Coder.}

\begin{prompttext}
To solve this problem, we want to determine how many containers, each weighing \pcode{w}, fit on an \pcode{n x n} grid such that their total weight does not exceed \pcode{maxWeight}.

\pfield{Step-by-step Plan:}
1. Calculate the total available space, which is the number of cells on the grid, $n \times n$.
2. Determine the maximum number of containers the weight constraint allows, $\lfloor \mathtt{maxWeight}/w \rfloor$.
3. Take the minimum of the two, since both limits apply.

Checking the two limits on a small case, a deck of side 3 carrying containers of weight 4 under a limit of 20 should print $\min(9,5)=5$.

\pfield{Code:} \cbq \pcode{python}
\pcode{\# Read input from standard input}
\pcode{n = int(input())}
\pcode{w = int(input())}
\pcode{maxWeight = int(input())}
\pblank
\pcode{\# Total number of cells on the grid}
\pcode{total\_cells = n * n}
\pblank
\pcode{\# Maximum number of containers limited by weight}
\pcode{container\_by\_weight = maxWeight // w}
\pblank
\pcode{\# Result is the minimum of the two constraints}
\pcode{result = min(total\_cells, container\_by\_weight)}
\pblank
\pcode{\# Print the final answer}
\pcode{print(result)}
\cbq
\end{prompttext}

\vspace{2pt}
\seg{psand!70!black}{Tester.}

\begin{prompttext}
An \pcode{n x n} grid gives $n^2$ positions and each holds at most one container of weight \pcode{w}, so the number loaded satisfies $k \le \lfloor \mathtt{maxWeight}/w \rfloor$ and $k \le n^2$, and the answer is the minimum of the two.

\elide{the choice of values, working through $n=5$, $w=6$ and $\mathtt{maxWeight}=150$, for which both limits equal 25}

\pfield{Test Input:} \cbq
\pcode{n = 5, w = 6, maxWeight = 150}
\cbq
\pfield{Test Output:} \cbq
\pcode{25}
\cbq
\end{prompttext}

\vspace{2pt}
\seg{pslate}{Environment.} Ran the program on that input and reported the disagreement.

\begin{prompttext}
\pcode{error: invalid literal for int() with base 10:}
\pcode{'n = 5, w = 6, maxWeight = 150'}
\end{prompttext}

\cnote{Both artefacts are in fact sound. The arithmetic $\min(25,25)=25$ is right, the program is the formula the specification states, and the trace names neither side. A reader at this point holds what the two roles hold, and the reading that suggests itself, that one of them must be wrong, is the wrong one.}
\end{caseregion}

\begin{caseregion}{\ding{184}\; Verdict on turn 0 \textnormal{— attached by PAC}}{pgold}{\bfseries teacher only}
\begin{promptflow}
\pfield{Verified attribution (available during training only):}
Verification result for the previous turn:

- On test input \lit{n = 5, w = 6, maxWeight = 150}, the program printed \lit{error: invalid literal for int() with base 10}, while the declared expected output was \lit{25}.

- Neither check finds fault with the artefact it reads, so no side is named and the teacher is told only that the disagreement survives both checks, a case the formatting of the input or the output commonly produces.
\end{promptflow}
\cnote{The two checks of~\Cref{app:pac_attribution} read one artefact each and neither reads the other, so they can both hold at once: the program passes the official tests, and the reference gives \lit{25} for the values the Tester declared. What differs from the task's prescription is the layout of those values rather than the values themselves, which is what the program's \lit{ValueError} reports, and the checks concern what each role worked out rather than how it was laid out, so the fault lies in neither artefact and the conclusion is that subsection's \texttt{BOTH\_CONSISTENT} line unaltered. The next turn bears this out: the Coder keeps its program, and the Tester, rather than correcting the value it had derived, records that both artefacts were right and rewrites the test in the prescribed layout. Every value named above was produced by an agent, one printed by the program and one declared by the Tester. No third value is introduced and no corrected input is supplied.}
\end{caseregion}

\begin{caseregion}{\ding{185}\; Turn 1 \textnormal{— student trajectory}}{pslate}{student and teacher}
\small
\seg{pcoral}{Coder.}

\begin{prompttext}
The program reads the three values one per line with \pcode{input()}, as the task requires, and returns $\min(n^2, \lfloor \mathtt{maxWeight}/w \rfloor)$, which is what the specification asks for. The trace reports a parsing failure on the first line it read rather than a wrong count, so the program already satisfies the specification and I keep it.

\pfield{Code:} \cbq \pcode{python}
\pcode{\# Read input from standard input}
\pcode{n = int(input())}
\pcode{w = int(input())}
\pcode{maxWeight = int(input())}
\pblank
\pcode{\# Calculate total available cells (containers) on the deck}
\pcode{total\_cells = n * n}
\pblank
\pcode{\# Calculate maximum possible containers based on weight constraint}
\pcode{max\_containers\_by\_weight = maxWeight // w}
\pblank
\pcode{\# The smaller of the two calculations is the answer}
\pcode{result = min(total\_cells, max\_containers\_by\_weight)}
\pblank
\pcode{\# Output the result}
\pcode{print(result)}
\cbq
\end{prompttext}

\vspace{2pt}
\seg{psand!70!black}{Tester.}

\begin{prompttext}
\pfield{Problem Analysis and Correction}
The count is $\min(n^2, \lfloor \mathtt{maxWeight}/w \rfloor)$, which the previous test case computed correctly as 25.

\pfield{Previous Test Case Mistake}
The expected output was right and so is the program: it was the input that was written as \pcode{n = 5, w = 6, maxWeight = 150}, in the notation the statement uses in its examples, whereas the task states that the three values are read one per line. The disagreement is one of format and the format of the test input is mine to correct.

\elide{the choice of replacement values, taking $n=10$, $w=10$ and $\mathtt{maxWeight}=87$ so that the weight limit rather than the deck is what binds}

\pfield{Test Input:} \cbq
\pcode{10}
\pcode{10}
\pcode{87}
\cbq
\pfield{Test Output:} \cbq
\pcode{8}
\cbq
\end{prompttext}

\vspace{2pt}
\seg{pslate}{Environment.} The program printed \lit{8} and the interaction ended in agreement.

\cnote{Neither role gave up what it had got right. The Coder kept its program and gave its reason, and the Tester, instead of revising the value it had derived as though that had been the error, recorded that both artefacts were sound and rewrote the test in the layout the task prescribes. No student receives a verdict at any turn, so both decisions were reached from the records above. Had either treated the mismatch as evidence that its own artefact was wrong, a correct program or a correct expected output would have been given up.}
\end{caseregion}

\casefoot{pcoral}
\captionof{figure}{\textbf{A conflict on the code domain and the turn that resolves it.} The regions, their order and their colours are those of~\Cref{fig:compose_code}, with the trajectory supplied rather than elided. The students hold region~\ding{183} and region~\ding{185}; region~\ding{182} is the substitution that distinguishes the two contexts of a pair and region~\ding{184} is the verdict, and no student receives either of them. The verdict restates the two values the agents produced and names where the conflict lies, here in the format of the test input rather than in either artefact. Passages tagged \tagword{ANALYSIS} are our commentary on the case, not trajectory text, and form no part of any context; region~\ding{184} in particular holds the two attribution lines and nothing besides. Long derivations are elided where marked.}
\label{fig:case_code}
\vspace{2 mm}

\subsubsection{Mathematics Domain}
\label{app:case_math}

\renewcommand{\casehue}{pteal}

\paragraph{What this case shows.}
Three turns are shown here because the verification reaches opposite conclusions within them.
The derivation and the program disagree at the first turn and the derivation is the one at fault, they disagree again at the second turn after the program has been rewritten and the program is the one at fault, and the roles themselves are the same throughout.
What changes between the two verdicts is only which artefact the reference answer supports, which is what it means for the two checks of~\Cref{app:pac_attribution} to read one artefact each: a verdict is not a ranking of the roles and cannot be anticipated from knowing which role is speaking.
Both roles move at turn 1 and only one of them should have, since the Reasoner abandons a case analysis that was genuinely incomplete while the Tool-User abandons an enumeration that was already correct, and the second verdict is what puts the run back on the value it had reached at the outset.

\casetitle{A conflict on the mathematics domain, scoring the Reasoner response of turn 2}{pteal}

\begin{caseregion}{\ding{182}\; RAS substitutes the role here}{pplum}{\bfseries teacher only}
\small
The target context opens with the Reasoner condition of~\Cref{box:role_reasoner} and the non-target one with the Tool-User condition of~\Cref{box:role_tooluser}, in place of the line the trajectory opens with.
Everything below is identical in the two.
\end{caseregion}

\begin{caseregion}{\ding{183}\; Turn 0 \textnormal{— student trajectory}}{pslate}{student and teacher}
\small
\seg{pslate}{Problem.} Held by both roles, in the \pfield{Problem:} field of the templates of~\Cref{app:student_math}.

\begin{prompttext}
Find the number of triples of nonnegative integers $(a,b,c)$ satisfying $a + b + c = 300$ and $a^2b + a^2c + b^2a + b^2c + c^2a + c^2b = 6{,}000{,}000$.
\end{prompttext}

\vspace{2pt}
\seg{pteal}{Reasoner.}

\begin{prompttext}
\pfield{Reasoning Steps:}
Since $a + b + c = 300$, each pair sums to $300$ minus the third coordinate, so
$$a^2b + a^2c + b^2a + b^2c + c^2a + c^2b = ab(300-c) + bc(300-a) + ca(300-b),$$
which expands to $300(ab+bc+ca) - 3abc$. Dividing the constraint by three leaves
$$100(ab+bc+ca) - abc = 2{,}000{,}000.$$

\elide{the cases examined from here, namely $c=0$, which forces $ab = 20{,}000$ and yields the triple $(200,100,0)$, and $a=b=c=100$}

A short loop over the pairs $(a,b)$, with $c$ fixed by the sum, would confirm the count.

\pfield{\#\#\#\#} 4
\end{prompttext}

\vspace{2pt}
\seg{pdeepteal}{Tool-User.}

\begin{prompttext}
The two constraints leave one free pair: choosing $a$ and $b$ fixes $c = 300 - a - b$, so every triple is reached by iterating over the pairs and evaluating the second constraint directly.

\pfield{Code:} \cbq \pcode{python}
\pcode{count = 0}
\pcode{target\_sum\_abc = 300}
\pcode{target\_expression\_value = 6\_000\_000}
\pblank
\pcode{for a in range(target\_sum\_abc + 1):}
\pcode{\ \ \ \ for b in range(target\_sum\_abc - a + 1):}
\pcode{\ \ \ \ \ \ \ \ c = target\_sum\_abc - a - b}
\pcode{\ \ \ \ \ \ \ \ \# Compute LHS of the second equation}
\pcode{\ \ \ \ \ \ \ \ lhs = (a*a*b + a*a*c + b*b*a +}
\pcode{\ \ \ \ \ \ \ \ \ \ \ \ \ \ \ b*b*c + c*c*a + c*c*b)}
\pcode{\ \ \ \ \ \ \ \ if lhs == target\_expression\_value:}
\pcode{\ \ \ \ \ \ \ \ \ \ \ \ count += 1}
\pblank
\pcode{print(count)}
\cbq
\end{prompttext}

\vspace{2pt}
\seg{pslate}{Environment.} Compared the two answers and reported the disagreement.

\begin{prompttext}
\pcode{derived answer: 4 \ \ \ \ printed result: 601}
\end{prompttext}

\cnote{The record names neither side. A closed-form argument that reduces the constraint to a single equation and an enumeration that visits every admissible pair are each the kind of artefact the other is normally checked against, and comparing them settles nothing, because they disagree and that is all the comparison says.}
\end{caseregion}

\begin{caseregion}{\ding{184}\; Verdict on turn 0 \textnormal{— attached by PAC}}{pgold}{\bfseries teacher only}
\begin{promptflow}
\pfield{Verified attribution (available during training only):}
Verification result for the previous turn:

- The derivation reported \lit{4} and the program printed \lit{601}.

- The printed result is consistent with the reference answer, the derived answer is not, and the previous inconsistency therefore originates from the derivation.
\end{promptflow}
\cnote{The conclusion is the \texttt{DERIVATION\_INCONSISTENT} line of~\Cref{app:pac_attribution} unaltered. Both values it names were produced by the agents, one derived and one printed, and the reference answer is not stated in its own right even though it decides the verdict.}
\end{caseregion}

\begin{caseregion}{\ding{185}\; Turn 1 \textnormal{— student trajectory}}{pslate}{student and teacher}
\small
\seg{pteal}{Reasoner.}

\begin{prompttext}
\pfield{Reasoning Steps:}
The identity is sound and the faulty step is the case analysis that followed it. Setting $c=0$ and setting $a=b=c$ reach two triples and stop, and they miss that every triple with a coordinate equal to $100$ satisfies the constraint: taking $a=100$ gives $ab+bc+ca = 100(b+c) + bc = 20{,}000 + bc$, so
$$100(ab+bc+ca) - abc = 100(20{,}000 + bc) - 100bc = 2{,}000{,}000$$
for every $b+c=200$, independently of $bc$.

\elide{the count of the triples having at least one coordinate equal to $100$, by inclusion and exclusion over the three coordinates}

\pfield{\#\#\#\#} 601
\end{prompttext}

\vspace{2pt}
\seg{pdeepteal}{Tool-User.}

\begin{prompttext}
Scanning every pair is wasteful, and a count two orders of magnitude above the derived one suggests the scan is admitting triples it should not. Restricting to $a \le b \le c$ and recovering the rest by permutation should cover the same ground at a fraction of the cost, and the triples the derivation exhibits can be inserted directly.

\pfield{Code:} \cbq \pcode{python}
\pcode{from itertools import permutations}
\pblank
\pcode{total = 300}
\pblank
\pcode{def check(a, b, c):}
\pcode{\ \ \ \ return (a + b + c == total and}
\pcode{\ \ \ \ \ \ \ \ \ \ \ \ a**2*b + a**2*c + b**2*a +}
\pcode{\ \ \ \ \ \ \ \ \ \ \ \ b**2*c + c**2*a + c**2*b == 6\_000\_000)}
\pblank
\pcode{seen = set()}
\elide{the triples the derivation exhibited, inserted into \pcode{seen} before the search begins}
\pblank
\pcode{for a in range(total // 3 + 1):}
\pcode{\ \ \ \ for b in range(max(0, total // 2 - a), total - a):}
\pcode{\ \ \ \ \ \ \ \ c = total - a - b}
\pcode{\ \ \ \ \ \ \ \ if a <= b <= c and check(a, b, c):}
\pcode{\ \ \ \ \ \ \ \ \ \ \ \ seen.update(permutations([a, b, c]))}
\pblank
\pcode{print(len(seen))}
\cbq
\end{prompttext}

\vspace{2pt}
\seg{pslate}{Environment.} Compared the two answers and reported the disagreement.

\begin{prompttext}
\pcode{derived answer: 601 \ \ \ \ printed result: 307}
\end{prompttext}

\cnote{Both roles moved, and the one that had been right moved too. The Reasoner repaired the step the verdict pointed at, while the Tool-User read the same disagreement as evidence against its own enumeration and narrowed a scan that had been complete. A second conflict follows, on the same problem and between the same roles, and it is now the computed side that departs from the reference.}
\end{caseregion}

\begin{caseregion}{\ding{186}\; Verdict on turn 1 \textnormal{— attached by PAC}}{pgold}{\bfseries teacher only}
\begin{promptflow}
\pfield{Verified attribution (available during training only):}
Verification result for the previous turn:

- The derivation reported \lit{601} and the program printed \lit{307}.

- The derived answer is consistent with the reference answer, the printed result is not, and the previous inconsistency therefore originates from the program.
\end{promptflow}
\cnote{This is the \texttt{COMPUTATION\_INCONSISTENT} line, and it points at the role the verdict above exonerated. The roles are the same as they were a turn earlier and so are their templates, so nothing about who is speaking accounts for the change and only the agreement with the reference does. A procedure that ranked the two roles, or that treated the computed value as the more reliable of the two by default, could not have produced these two verdicts in this order.}
\end{caseregion}

\begin{caseregion}{\ding{187}\; Turn 2 \textnormal{— student trajectory}}{pslate}{student and teacher}
\small
\seg{pdeepteal}{Tool-User.}

\begin{prompttext}
The previous program restricted the scan to $b \ge 150 - a$, and that bound removes triples the ordering constraint keeps: at $a=0$ the scan starts at $b=150$, so $(0,100,200)$ is never examined. The defect is the bound rather than the enumeration, and dropping it costs one pass over the pairs and skips nothing. The seven triples written in by hand go as well, since a scan that skips nothing has no need of them.

\pfield{Code:} \cbq \pcode{python}
\pcode{total\_sum = 300}
\pcode{expression\_target = 6\_000\_000}
\pblank
\pcode{count = 0}
\pcode{for a in range(total\_sum + 1):}
\pcode{\ \ \ \ for b in range(total\_sum - a + 1):}
\pcode{\ \ \ \ \ \ \ \ c = total\_sum - a - b}
\pcode{\ \ \ \ \ \ \ \ lhs = (a * a * b + a * a * c + b * b * a +}
\pcode{\ \ \ \ \ \ \ \ \ \ \ \ \ \ \ b * b * c + c * c * a + c * c * b)}
\pcode{\ \ \ \ \ \ \ \ if lhs == expression\_target:}
\pcode{\ \ \ \ \ \ \ \ \ \ \ \ count += 1}
\pblank
\pcode{print(count)}
\cbq
\end{prompttext}

\vspace{2pt}
\seg{pteal}{Reasoner.}

\begin{prompttext}
\pfield{Reasoning Steps:}
The count rests on the identity and on the family with a coordinate equal to $100$, and that family is exhibited rather than searched for, so the derivation stands on its own and the printed $307$ does not bear on it. The previous program reported $601$ before it was rewritten and $307$ after, and a count that falls when a scan is narrowed is a count that was narrowed past what it should cover.

\pfield{\#\#\#\#} 601
\end{prompttext}

\vspace{2pt}
\seg{pslate}{Environment.} Both answers were \lit{601} and the interaction ended in agreement.

\cnote{The run passes through both verdicts and ends on the value the reference supports. No student receives either of them, so the repair at turn 1 and the one at turn 2 were reached from the records above alone, and what the pair converged on is not the answer that was stated more confidently but the one each role could still justify after being made to look again.}
\end{caseregion}

\casefoot{pteal}

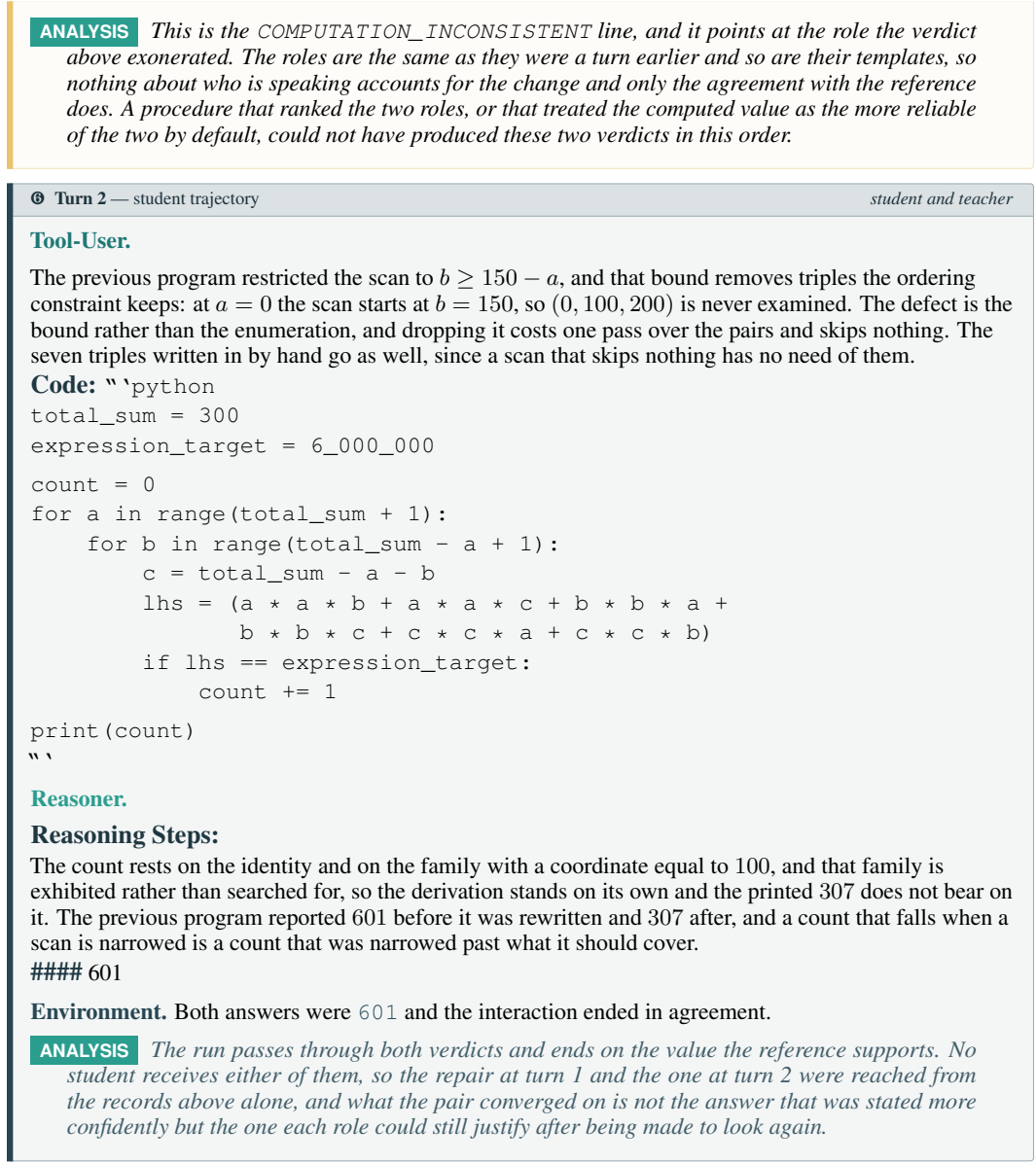
\captionof{figure}{\textbf{A conflict on the mathematics domain, resolved in two steps.} The regions and their colours are those of~\Cref{fig:compose_math}, with the trajectory supplied rather than elided; regions~\ding{184} and~\ding{185} together with regions~\ding{186} and~\ding{187} are the two instances of the verdict and turn pair that the dashed block of that figure repeats. The students hold regions~\ding{183},~\ding{185} and~\ding{187}, and receive neither the substitution of region~\ding{182} nor either verdict. The two verdicts name different roles, the derivation at turn 0 and the program at turn 1, and each states which of the two candidate values the reference answer supports without stating that answer. Passages tagged \tagword{ANALYSIS} are our commentary on the case, not trajectory text, and form no part of any context. Long derivations are elided where marked.}
\label{fig:case_math}
\vspace{2 mm}

\subsection{Where the Role Advantage Falls}
\label{app:case_ras}

The role advantage of~\Cref{eq:role_adv} is defined on a single token, as the difference between the two teacher scores that~\Cref{app:ras_contexts} obtains for it, so it is shown here on the tokens rather than summarised over a response.
What is shaded is the response alone.
The teacher force decodes the response against a context and scores its tokens, so a score exists at those positions and nowhere else, and the problem statement together with the role condition is what the response is read against rather than something the signal is defined on.
The role condition in particular is the one passage that differs between the two contexts of a pair, and what the shading reports is the effect of that difference on the response rather than the difference itself.
Each display takes the turn-0 response of the run above it, divides it into blocks of a few tokens running from the first word to the last, and shades every block by the level its role advantage sits at, the three regimes of~\Cref{sec:ras} appearing as levels above, at and below the middle of the scale.
The signal is defined at every position, so every block carries a level and the middle one is a value near zero rather than the absence of a value; only an elision is left plain, standing as it does for tokens that are not on the page.

\subsubsection{Code Domain}
\label{app:case_ras_code}

\renewcommand{\casehue}{pcoral}

\paragraph{What this case shows.}
The response is the one the Coder produced at turn 0 of~\Cref{fig:case_code}, scored once under the Coder condition of~\Cref{box:role_coder} and once under the Tester condition of~\Cref{box:role_tester}.
The response sorts itself into the three regimes of~\Cref{sec:ras} without anything in the training objective naming them.
The work the Coder condition centres on sits above the middle, which is the plan as it turns into a decision and the statements that carry those decisions out; the one stretch in which the Coder fixes on a case of its own and states the output the specification implies for it sits below the middle, that being what the second paragraph of the Coder condition sets outside the role; and what neither condition claims stays at the middle, which covers the restatement of the task, the field labels and the fence, and the comments that describe in prose what the statements beside them do.
The last of these is the one worth pausing on, since a comment and the statement it annotates are adjacent and describe the same thing while sitting a band apart, and no rule written over the surface of a response would separate them.
None of the three is uniform: the level jumps by two or three steps between neighbouring blocks as readily as by one, and a handful of blocks sit on the far side of the middle from everything around them.

\casetitle{The role advantage on the Coder response of turn 0}{pcoral}
\raskey{Tester}{Coder}

\begin{caseregion}{\ding{182}\; Response as submitted \textnormal{— shaded by role advantage}}{pslate}{response: student and teacher}
\small
\begin{prompttext}
\tkO{To solve this}\nA{ problem, we}\tkO{ want to}\pA{ determine how}\tkO{ many containers,}\nA{ each weighing}\pA{ \pcode{w}, fit}\tkO{ on an}\nB{ \pcode{n x n} grid}\tkO{ such that}\pA{ their total}\pA{ weight does}\tkO{ not exceed}\nA{ \pcode{maxWeight}.}

\tkO{\pfield{Step-by-step Plan:}}
\nA{1. Calculate}\pA{ the total}\pC{ available space,}\tkO{ which is}\pC{ the number}\pB{ of cells}\nA{ on the}\pB{ grid, $n \times n$.}
\pA{2. Determine}\pB{ the maximum}\pD{ number of}\pA{ containers the}\pD{ weight constraint}\pC{ allows, $\lfloor \mathtt{maxWeight}/w \rfloor$.}
\pB{3. Take}\pD{ the minimum}\pE{ of the}\pC{ two, since}\pE{ both limits}\pA{ apply.}

\nA{Checking the}\nB{ two limits}\pA{ on a}\nC{ small case,}\nE{ a deck}\nD{ of side}\nE{ 3 carrying}\nB{ containers of}\nE{ weight 4}\nC{ under a}\nA{ limit of}\pB{ 20 should}\nD{ print $\min(9,5)$}\nE{ $=5$.}

\tkO{\pfield{Code:} \cbq \pcode{python}}
\tkO{\pcode{\# Read input}}\nA{\pcode{ from standard}}\tkO{\pcode{ input}}
\pB{\pcode{n = int(}}\pA{\pcode{input())}}
\pA{\pcode{w = int(}}\nA{\pcode{input())}}
\pC{\pcode{maxWeight = }}\pA{\pcode{int(input())}}
\pblank
\tkO{\pcode{\# Total number}}\pA{\pcode{ of cells}}\tkO{\pcode{ on the grid}}
\pB{\pcode{total\_cells}}\pD{\pcode{ = n}}\pB{\pcode{ * n}}
\pblank
\nA{\pcode{\# Maximum number}}\tkO{\pcode{ of containers}}\pA{\pcode{ limited by}}\tkO{\pcode{ weight}}
\pC{\pcode{container\_by\_weight}}\pE{\pcode{ = maxWeight}}\pD{\pcode{ // w}}
\pblank
\tkO{\pcode{\# Result is}}\pB{\pcode{ the minimum}}\pC{\pcode{ of the two}}\tkO{\pcode{ constraints}}
\pD{\pcode{result = min(}}\pE{\pcode{total\_cells,}}\pC{\pcode{ container\_}}\pE{\pcode{by\_weight)}}
\pblank
\tkO{\pcode{\# Print the}}\nA{\pcode{ final answer}}
\pC{\pcode{print(}}\pA{\pcode{result)}}
\tkO{\cbq}
\end{prompttext}

\cnote{The middle band is a role advantage near zero rather than a gap, and three kinds of text fall in it. The restatement of the task wanders a step either side without settling, since both conditions require a role that has understood the problem. The field labels and the fence sit there as well, which they must, because a signal that singled out the markers a template happens to require would be sorting formatting rather than responsibility. So do the comments, and they sit a band or two below the statements they annotate even though the two say the same thing, one in prose and one in Python.}

\cnote{The level climbs as the plan turns into a decision, and it is the average over a passage that climbs rather than every block within it. The three steps of the plan sit higher in turn, and the two assignments that choose what to compute are the highest blocks in the response, which is what the Coder condition means by solution design and program implementation. The climb tracks the decisions and not the code: the input reads, which any role writing a program at all would produce, sit barely above the middle, and the comments between the statements return to it. Individual blocks go the other way throughout, \pcode{on the} in the first step and \pcode{input())} in the second read among them, each a block whose tokens neither role can claim. The objective of~\Cref{eq:ras} is built for a signal of exactly this shape, weighting every token by the role advantage rather than thresholding it, so a block that departs from the passage around it contributes its own small amount and no passage turns on any one of them.}

\cnote{The level falls through the passage in which the Coder settles on values of its own, reaching its lowest at the values themselves and at the expected output derived from them. The Coder condition puts constructing test inputs and independently deriving their expected outputs on the Tester's side, so the teacher reading those blocks as the Tester favours them more than the teacher reading them as the Coder does. The fall begins a block before the values appear, is interrupted twice, once at \pcode{on a} near its start and once at \pcode{20 should} near its end, and has not returned to the middle when the code fence arrives. RAS reduces the strength of the update across it and removes nothing, which is why the same passage still stands in~\Cref{fig:case_code}.}

\cnote{The levels are not a measure of how much a block matters, and the passage that falls furthest is the place to see it: that derivation is correct, it is useful, and the ordinary distillation term treats it as such. It falls here only because the two identities disagree about whose work it is, and it would rise by as much were the Tester the one to write it, since exchanging the two conditions of~\Cref{app:ras_contexts} exchanges the sign of every block. What survives the exchange is the arithmetic of~\Cref{eq:role_adv}, in which the student term cancels and the two contexts differ in their opening line alone.}
\end{caseregion}

\casefoot{pcoral}
\captionof{figure}{\textbf{The role advantage on a Coder response.} The response is that of turn 0 of~\Cref{fig:case_code}, unchanged, and it is the whole of what is shaded, the two contexts it is scored against carrying no score of their own. It is divided into blocks of a few tokens and each block is shaded by the level its role advantage sits at, obtained by scoring the response under the Coder condition and under the Tester condition of~\Cref{app:role_conditions} and taking the difference. The blocks are of one size and cut across the phrasing rather than following it, and the levels run in the eleven steps of the key, which record the sign and the strength of the difference and are what the objective of~\Cref{eq:ras} consumes. Every block carries a level, the middle one being a role advantage near zero rather than a gap in the signal, and it is where the formatting the template requires sits along with the restatement of the task. What separates is the average over a passage rather than the level of any one block. The stretches that decide what the program computes stand above the middle, the stretch in which the Coder settles on values of its own and states the output the specification implies for them stands below it, and the restatement of the task, the formatting and the comments stand at it, so the three regimes of~\Cref{sec:ras} appear without the objective naming any of them. Within each the level jumps by two or three steps as readily as by one and a few blocks sit on the far side of the middle from everything around them. Passages tagged \tagword{ANALYSIS} are our commentary and form no part of any context.}
\label{fig:case_ras_code}
\vspace{2 mm}

\subsubsection{Mathematics Domain}
\label{app:case_ras_math}

\renewcommand{\casehue}{pteal}

\paragraph{What this case shows.}
The response is the one the Reasoner produced at turn 0 of~\Cref{fig:case_math}, scored under the Reasoner condition of~\Cref{box:role_reasoner} and under the Tool-User condition of~\Cref{box:role_tooluser}.
The three regimes appear as they do on the code domain, and the reason for giving a second display is that here they cannot be explained away as a separation of notations.
Both roles of this workflow write mathematics about the same object and this response never leaves prose and algebra, so what the level tracks is which role the work belongs to and not whether the tokens are English or Python.
The algebra that rewrites the constraint stands above the middle, the sentence that prescribes an enumeration stands below it, and the final answer sits at the middle directly beneath the derivation that produced it, being the one thing both templates of~\Cref{app:student_math} ask for.

\casetitle{The role advantage on the Reasoner response of turn 0}{pteal}
\raskey{Tool-User}{Reasoner}

\begin{caseregion}{\ding{182}\; Response as submitted \textnormal{— shaded by role advantage}}{pslate}{response: student and teacher}
\small
\begin{prompttext}
\tkO{\pfield{Reasoning Steps:}}
\tkO{Since $a + b + c$}\nA{ $= 300$, each}\tkO{ pair sums to}\pA{ 300 minus the}\pB{ third coordinate, so}\tkO{ the constraint reads}
\pC{$ab(300-c)$}\pE{ $+\, bc(300-a)$}\pD{ $+\, ca(300-b)$}\pB{, which expands}\tkO{ to}\pD{ $300(ab+bc+ca)$}\pE{ $-\,3abc$}\pB{, and dividing}\nA{ by three}\pC{ leaves $100(ab+bc+ca)$}\pE{ $-\,abc = 2{,}000{,}000$}\tkO{.}

\elide{the cases examined from here, namely $c=0$, which forces $ab = 20{,}000$ and yields the triple $(200,100,0)$, and $a=b=c=100$}

\nA{A short loop}\nC{ over the pairs}\nE{ $(a,b)$, with $c$}\nD{ fixed by the}\nE{ sum, would}\pA{ confirm the}\nD{ count.}

\tkO{\pfield{\#\#\#\#}}\tkO{ 4}
\end{prompttext}

\cnote{The answer sits at the middle, one line below the algebra that reached it. Both templates of~\Cref{app:student_math} end by asking for a final answer and both conditions favour a role that states one, so the value carries no role difference while the route to it carries a great deal. This is the mathematics counterpart of a comment and its statement in~\Cref{fig:case_ras_code}: a result and the derivation that produced it are as close as two passages get, and anything sorting the response by what it is about would keep them together. The restatement of the constraint at the opening is at the middle for the same reason the code response begins there, both conditions requiring a role that has read the problem.}

\cnote{The level is highest across the algebra, and this is the part of the display that a difference of notations cannot account for. Nothing here is a program: the response is prose and algebra throughout, the Tool-User writes mathematics too, and the passages that separate are the ones that rewrite the constraint into a single symmetric identity, which is what the Reasoner condition means by analytic derivation and mathematical justification. The climb is uneven, dipping at \pcode{$= 300$, each} where the problem is being quoted back and at \pcode{by three}, and the elided passage carries no level at all, standing as it does for text that is not on the page.}

\cnote{The closing sentence prescribes a program and the level falls across it, although the sentence is brief, correct and useful. The Tool-User condition makes writing the computational program its responsibility and tells the Reasoner not to take it over, so a teacher reading those blocks as the Tool-User favours them more than one reading them as the Reasoner does. It is also the one place in the turn where the two roles overlap, the enumeration described here being the one the Tool-User submits in~\Cref{fig:case_math}, so an update that reinforced it on both sides would be an update spent teaching two agents the same thing. The fall is interrupted at \pcode{confirm the}, a block whose tokens belong to neither role.}
\end{caseregion}

\casefoot{pteal}
\captionof{figure}{\textbf{The role advantage on a Reasoner response.} The response is that of turn 0 of~\Cref{fig:case_math}, unchanged apart from the derivations being set inline so that the blocks can cross them, and it is the whole of what is shaded. It is divided as in~\Cref{fig:case_ras_code} and shaded on the same scale, here obtained by scoring the response under the Reasoner condition and under the Tool-User condition of~\Cref{app:role_conditions} and taking the difference. The algebra that rewrites the constraint stands above the middle, the sentence prescribing an enumeration stands below it, and the restatement of the constraint and the final answer stand at it, the answer being the one thing both templates of~\Cref{app:student_math} ask of either role. Neither role writes a program in this response, so the separation is not one of notation. Within each stretch the level jumps by two or three steps as readily as by one, and the elision carries no level, standing for tokens that are not shown. Passages tagged \tagword{ANALYSIS} are our commentary and form no part of any context.}
\label{fig:case_ras_math}
\vspace{2 mm}

\end{document}